\documentclass[10pt,journal,compsoc]{IEEEtran}
\ifCLASSOPTIONcompsoc
\usepackage[nocompress]{cite}
\usepackage{amsmath,amsfonts}
\usepackage{algorithmicx}

\usepackage{algpseudocode}
\usepackage{array}
\usepackage{amsmath}
\usepackage{amsmath, bm}
\usepackage{threeparttable}
\usepackage{enumitem}
\usepackage[caption=false,font=normalsize,labelfont=sf,textfont=sf]{subfig}
\usepackage{booktabs}
\usepackage{bbding}
\usepackage{dsfont}
\usepackage{color} 
\usepackage[ruled,linesnumbered]{algorithm2e}
\usepackage[pagebackref=true,breaklinks=true,colorlinks=true,bookmarks=false,linkcolor=red,anchorcolor=black,citecolor=green, urlcolor=black]{hyperref}

\usepackage[table]{xcolor}
\usepackage{textcomp}
\usepackage{pifont}
\usepackage{stfloats}
\usepackage{enumerate}
\usepackage{url}
\usepackage{verbatim}
\usepackage{graphicx}
\usepackage{cite}
\usepackage{bbm}
\usepackage{multirow}
\else
\usepackage{cite}
\fi
\ifCLASSINFOpdf
\else
\fi
\begin{document}
%
% paper title
% Titles are generally capitalized except for words such as a, an, and, as,
% at, but, by, for, in, nor, of, on, or, the, to and up, which are usually
% not capitalized unless they are the first or last word of the title.
% Linebreaks \\ can be used within to get better formatting as desired.
% Do not put math or special symbols in the title.
% \title{Prompt-and-Transfer: Prompt-Driven for Few-shot Segmentation}
\title{RingMo-Aerial: An Aerial Remote Sensing Foundation Model With Affine Transformation Contrastive Learning}
%
%
% author names and IEEE memberships
% note positions of commas and nonbreaking spaces ( ~ ) LaTeX will not break
% a structure at a ~ so this keeps an author's name from being broken across
% two lines.
% use \thanks{} to gain access to the first footnote area
% a separate \thanks must be used for each paragraph as LaTeX2e's \thanks
% was not built to handle multiple paragraphs
%
%
%\IEEEcompsocitemizethanks is a special \thanks that produces the bulleted
% lists the Computer Society journals use for "first footnote" author
% affiliations. Use \IEEEcompsocthanksitem which works much like \item
% for each affiliation group. When not in compsoc mode,
% \IEEEcompsocitemizethanks becomes like \thanks and
% \IEEEcompsocthanksitem becomes a line break with idention. This
% facilitates dual compilation, although admittedly the differences in the
% desired content of \author between the different types of papers makes a
% one-size-fits-all approach a daunting prospect. For instance, compsoc 
% journal papers have the author affiliations above the "Manuscript
% received ..."  text while in non-compsoc journals this is reversed. Sigh.

\author{
Wenhui~Diao$^{\dag}$, Haichen~Yu$^{\dag}$, Kaiyue~Kang$^{\dag}$, Tong~Ling$^{\dag}$,
Di~Liu, Yingchao~Feng, Hanbo~Bi, Libo~Ren, Xuexue~Li, Yongqiang~Mao, Xian~Sun

% Hanbo~Bi,~%\IEEEmembership{Member,~IEEE,}
% Yingchao~Feng,~\IEEEmembership{Member,~IEEE,}
% Wenhui~Diao,~\IEEEmembership{Member,~IEEE,}
% Peijin~Wang,~\IEEEmembership{Member,~IEEE,}\\ 
% Yongqiang~Mao,~\IEEEmembership{Graduate Student Member,~IEEE,} 
% Kun~Fu,~\IEEEmembership{Member,~IEEE,}\\
% Hongqi~Wang,~\IEEEmembership{Member,~IEEE,}
% and~Xian~Sun,~\IEEEmembership{Senior Member,~IEEE}% <-this % stops a space
\IEEEcompsocitemizethanks{
\IEEEcompsocthanksitem W. Diao, H. Yu, K. Kang, and T. Ling labeled with ${\dag}$ contribute equally to this work.
\IEEEcompsocthanksitem This work was supported by the National Nature Science Foundation of China under Grant 62331027. (Corresponding authors: Xian Sun.)
\IEEEcompsocthanksitem W. Diao, H. Yu, K. Kang, T. Ling, H. Bi, L. Ren, X. Li, and X. Sun are with the Aerospace Information Research Institute, Chinese Academy of Sciences, Beijing 100190, China, also with the School of Electronic, Electrical and Communication Engineering, University of Chinese Academy of Sciences, Beijing 100190, China, also with the University of Chinese Academy of Sciences, Beijing 100190, China, and also with the Key Laboratory of Target Cognition and Application Technology (TCAT), Aerospace Information Research Institute, Chinese Academy of Sciences, Beijing 100190, China (e-mail: diaowh@aircas.ac.cn, yuhaichen18@mails.ucas.ac.cn, kangkaiyue23@mails.ucas.ac.cn, lingtong23@mails.ucas.ac.cn, sunxian@aircas.ac.cn).
\IEEEcompsocthanksitem Y. Mao is with the Department of Electronic Engineering, Tsinghua University, Beijing 100084, China.
\IEEEcompsocthanksitem D. Liu and Y. Feng are with the Aerospace Information Research Institute, Chinese Academy of Sciences, Beijing 100190, China, and also with the Key Laboratory of Target Cognition and Application Technology (TCAT), Aerospace Information Research Institute, Chinese Academy of Sciences, Beijing 100190, China.

}% <-this % stops an unwanted space
% \thanks{Manuscript received April 19, 2005; revised August 26, 2015.}
}

% note the % following the last \IEEEmembership and also \thanks - 
% these prevent an unwanted space from occurring between the last author name
% and the end of the author line. i.e., if you had this:
% 
% \author{....lastname \thanks{...} \thanks{...} }
%                     ^------------^------------^----Do not want these spaces!
%
% a space would be appended to the last name and could cause every name on that
% line to be shifted left slightly. This is one of those "LaTeX things". For
% instance, "\textbf{A} \textbf{B}" will typeset as "A B" not "AB". To get
% "AB" then you have to do: "\textbf{A}\textbf{B}"
% \thanks is no different in this regard, so shield the last } of each \thanks
% that ends a line with a % and do not let a space in before the next \thanks.
% Spaces after \IEEEmembership other than the last one are OK (and needed) as
% you are supposed to have spaces between the names. For what it is worth,
% this is a minor point as most people would not even notice if the said evil
% space somehow managed to creep in.

% The paper headers
\markboth{Manuscripts submitted to IEEE TPAMI}%
{Shell \MakeLowercase{\textit{et al.}}: Bare Demo of IEEEtran.cls for Computer Society Journals}
% The only time the second header will appear is for the odd numbered pages
% after the title page when using the twoside option.
% 
% *** Note that you probably will NOT want to include the author's ***
% *** name in the headers of peer review papers.                   ***
% You can use \ifCLASSOPTIONpeerreview for conditional compilation here if
% you desire.

% The publisher's ID mark at the bottom of the page is less important with
% Computer Society journal papers as those publications place the marks
% outside of the main text columns and, therefore, unlike regular IEEE
% journals, the available text space is not reduced by their presence.
% If you want to put a publisher's ID mark on the page you can do it like
% this:
%\IEEEpubid{0000--0000/00\$00.00~\copyright~2015 IEEE}
% or like this to get the Computer Society new two part style.
%\IEEEpubid{\makebox[\columnwidth]{\hfill 0000--0000/00/\$00.00~\copyright~2015 IEEE}%
%\hspace{\columnsep}\makebox[\columnwidth]{Published by the IEEE Computer Society\hfill}}
% Remember, if you use this you must call \IEEEpubidadjcol in the second
% column for its text to clear the IEEEpubid mark (Computer Society jorunal
% papers don't need this extra clearance.)

% use for special paper notices
%\IEEEspecialpapernotice{(Invited Paper)}

% for Computer Society papers, we must declare the abstract and index terms
% PRIOR to the title within the \IEEEtitleabstractindextext IEEEtran
% command as these need to go into the title area created by \maketitle.
% As a general rule, do not put math, special symbols or citations
% in the abstract or keywords.
\IEEEtitleabstractindextext{%
\begin{abstract}
Aerial Remote Sensing (ARS) vision tasks \textcolor{black}{present} significant challenges due to the unique \textcolor{black}{viewing angle characteristics}. Existing research has primarily focused on algorithms for specific tasks, which have limited applicability in a broad range of ARS vision applications. This paper proposes RingMo-Aerial, aiming to fill the gap in foundation model research in the field of ARS vision. \textcolor{black}{A Frequency-Enhanced Multi-Head Self-Attention (FE-MSA) mechanism is introduced to strengthen the model’s capacity for small-object representation. Complementarily, an affine transformation-based contrastive learning method improves its adaptability to the tilted viewing angles inherent in ARS tasks.} Furthermore, the ARS-Adapter, an efficient parameter fine-tuning method, is proposed to improve the model's adaptability and \textcolor{black}{performance} in various ARS vision tasks. Experimental results demonstrate that RingMo-Aerial achieves SOTA performance on multiple downstream tasks. This indicates the practicality and efficacy of RingMo-Aerial in enhancing the performance of ARS vision tasks.

\end{abstract}

% Note that keywords are not normally used for peerreview papers.
}

% make the title area
\maketitle

% To allow for easy dual compilation without having to reenter the
% abstract/keywords data, the \IEEEtitleabstractindextext text will
% not be used in maketitle, but will appear (i.e., to be "transported")
% here as \IEEEdisplaynontitleabstractindextext when the compsoc 
% or transmag modes are not selected <OR> if conference mode is selected 
% - because all conference papers position the abstract like regular
% papers do.
\IEEEdisplaynontitleabstractindextext
% \IEEEdisplaynontitleabstractindextext has no effect when using
% compsoc or transmag under a non-conference mode.

% For peer review papers, you can put extra information on the cover
% page as needed:
% \ifCLASSOPTIONpeerreview
% \begin{center} \bfseries EDICS Category: 3-BBND \end{center}
% \fi
%
% For peerreview papers, this IEEEtran command inserts a page break and
% creates the second title. It will be ignored for other modes.
\IEEEpeerreviewmaketitle

\begin{figure*}
\centering
\includegraphics[width=0.7\linewidth]{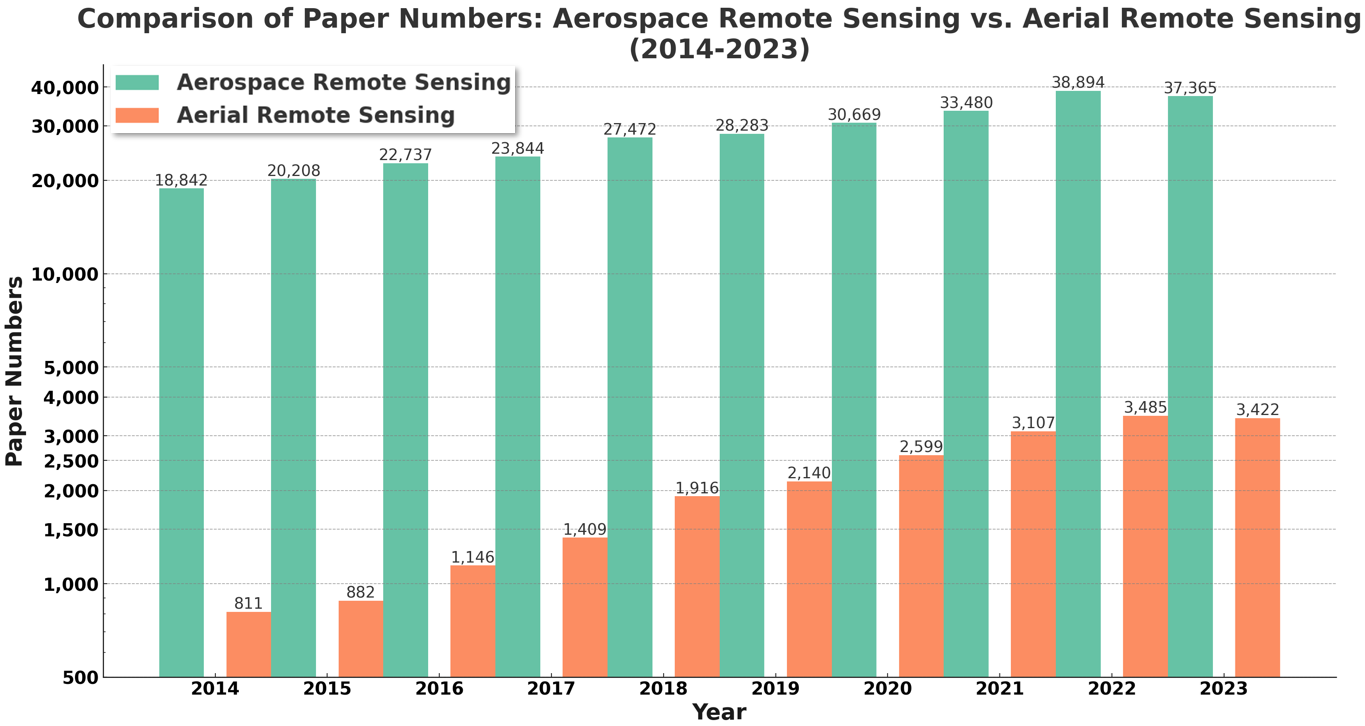}
\caption{Number of Aerospace Remote Sensing and Aerial Remote Sensing papers in Web of Science in the last decade. The green bars represent aerospace remote sensing, while the orange bars represent aerial remote sensing. Although research on ARS has grown, the number of papers remains significantly lower than that of aerospace remote sensing. }
\label{fig:paper_num}
\end{figure*}

\begin{figure*}
\centering
\includegraphics[width=0.95\linewidth]{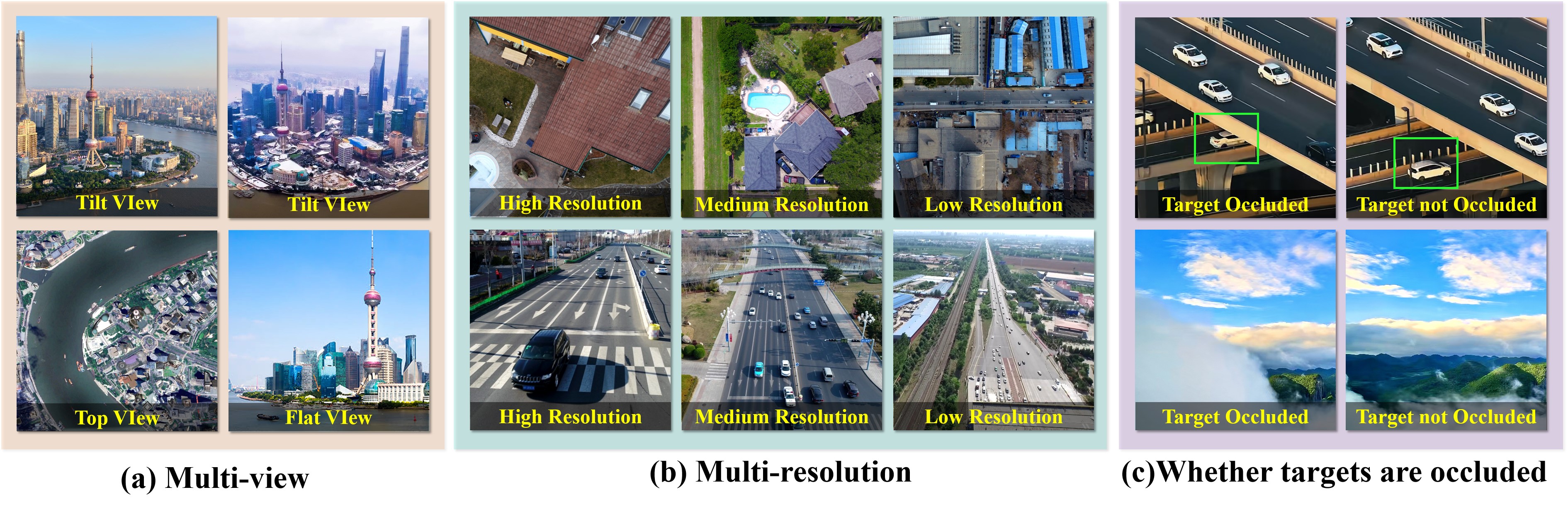}
\caption{Illustration of multi-view, multi-resolution, and occlusion of ARS images: (a) Multi-view imagery shows images captured from different angles, including tilt, top, and flat views; (b) Multi-resolution imagery demonstrates how varying low, medium, and high resolutions affect the level of detail; (c) Target occlusion compares cases where the target is partially occluded versus not occluded. These characteristics highlight the complexity of ARS imagery in different scenarios.}
\label{fig:vis-intro-multi}
\end{figure*}

\IEEEraisesectionheading{\section{Introduction}\label{sec:introduction}}
\IEEEPARstart{A}{erial} Remote Sensing (ARS) is an essential method of data acquisition, distinct from aerospace remote sensing (ApRS) in its flexibility, as it is typically free from constraints such as revisit cycles, satellite sensor incidence angles, and timing limitations. ARS sensors, usually mounted on Unmanned Aerial Vehicles (UAVs), airplanes, or balloons, offer advantages over satellite-based remote sensing by quickly reaching target areas and capturing real-time, multi-perspective observations at various resolutions. This capability enables ARS to provide detailed information in applications ranging from emergency response to environmental monitoring and national defense. Thus, enhancing intelligent information extraction from ARS images holds significant academic and practical value.

Despite ARS's distinct advantages and applications, research efforts in intelligent remote sensing interpretation are still largely focused on ApRS imagery. As shown in Fig.\ref{fig:paper_num}, although interest in ARS has grown rapidly in recent years, studies in its intelligent interpretation remain relatively limited, with publications in this area comprising less than 10\% of ApRS-related work.

% \begin{figure}
% \centering
% \includegraphics[width=1.0\linewidth]{images/output.jpg}
% \caption{RingMo-Aerial V.S Previous SOTA}
% \label{fig:sota-compare}
% \end{figure}
Unlike ApRS, ARS imagery is characterized by unique physical properties, including multi-view perspectives, varied resolutions, and occlusion effects, due to its flexible observational angles and sensor characteristics, as shown in Fig.\ref{fig:vis-intro-multi}. In Fig.\ref{fig:vis-intro-multi}(a), various perspectives are depicted, including oblique views (i and ii), a vertical view (iii), and a horizontal view (iv). Unlike ApRS, which predominantly captures vertical views, ARS imagery often incorporates oblique angles, providing more dimensional perspectives of targets. However, this also results in distant objects appearing smaller and more densely arranged, increasing the complexity of image interpretation. Fig.\ref{fig:vis-intro-multi}(b) shows images at low, medium, and high resolutions (i, ii, iii, respectively), reflecting the diverse resolutions produced by ARS sensors at varying altitudes and speeds. Such differences in resolution affect target clarity and detail across images, placing higher demands on model adaptability. Additionally, significant occlusion is common in aerial imagery, as demonstrated in Fig.\ref{fig:vis-intro-multi}(c), where some targets are partially or fully obscured by other objects, further complicating image interpretation.

These challenges highlight the need for specialized intelligent models tailored to ARS applications. While recent foundational models in computer vision and RS have shown potential \textcolor{black}{capabilities}, they are often inadequate for ARS due to its unique requirements. These models benefit from pre-training on vast amounts of data and large-scale parameters, enabling them to adapt more rapidly and effectively to various changes.

\begin{figure*}
\centering
\includegraphics[width=0.95\linewidth]{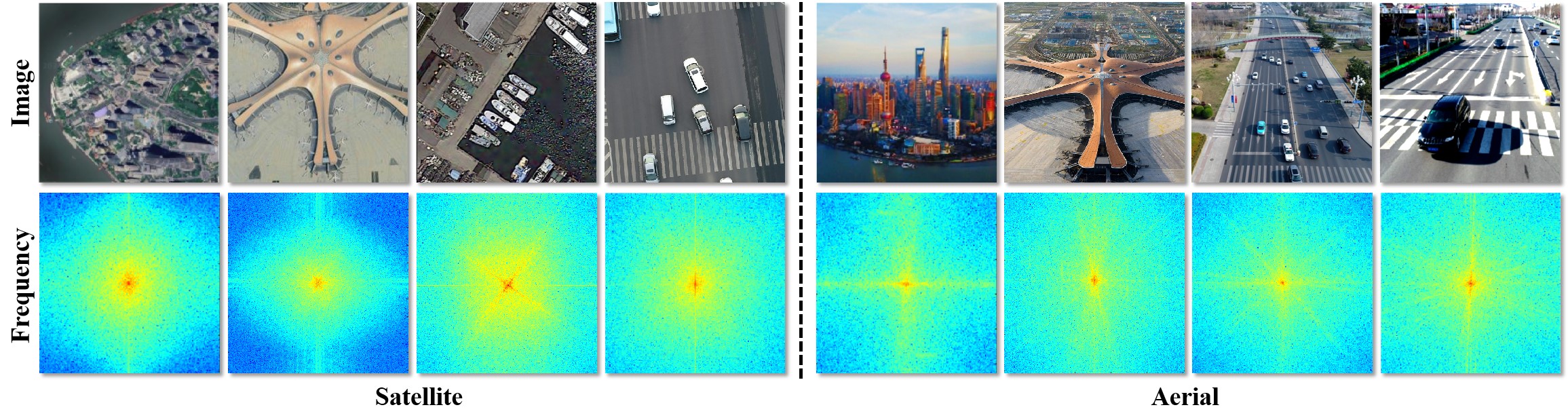}
\caption{Schematic Comparison of ApRS and ARS Image Features. Satellite images contain more low-frequency information with continuous features, while aerial images have more detailed variations and relatively less low-frequency information.}
\label{fig:fft}
\end{figure*}

However, current foundational model research primarily focuses on ApRS imagery, as evidenced by studies like RingMo \cite{sun2022ringmo} and Skysense \cite{guo2024skysense}. This underscores the need for a dedicated pre-trained foundational model specifically designed for ARS. Adapting existing foundational models for ARS involves addressing several critical challenges:

\noindent \textbf{(1) Challenges in Pre-training}: Pre-training generally relies on large-scale datasets for self-supervised learning, primarily using Masked Image Modeling (MIM) and Contrastive Learning (CL). MIM, employed by models like MAE \cite{he2022masked} and SimMIM \cite{xie2022simmim}, masks parts of images for reconstruction to enhance feature extraction, while CL, seen in models like MOCO \cite{he2020momentum}, leverages positive and negative sample pairs to learn relational features. Both methods have merits but also limitations; MIM is widely used across visual tasks, as RingMo\cite{sun2022ringmo}. However, ARS images vary significantly in features due to differences in perspective and resolution. As shown in Fig.\ref{fig:fft}, the vertical view displays structured, continuous features with broad areas of color consistency, containing abundant low-frequency information. In contrast, the oblique view, capturing building sides and heights, offers dynamic perspectives with high detail variability and lower low-frequency content. Current methods, which often focus on reconstructing local image information, struggle to capture these unique variations in ARS imagery.

% There is a clear distinction between the features of images captured at oblique and vertical angles. The current self-supervised learning methods rely on reconstructing local information in images and struggle to capture these variations effectively.

% \noindent \textbf{(2) Challenges in Fine-tuning}: Traditional model fine-tuning faces challenges, particularly in ARS, due to the extensive time and effort required. Since foundational models are pre-trained on a wide range of tasks, global fine-tuning for specific downstream applications may lead to overfitting, especially in ARS. Unlike ApRS, which typically involves a single perspective and resolution, ARS models risk overfitting to specific perspectives and resolutions, limiting their generalizability across different tasks and scenarios.

\noindent \textbf{(2) Challenges in Fine-tuning}: The traditional approach of global fine-tuning of models faces significant challenges, mainly due to the fact that fine-tuning may lead to overfitting, especially in ARS. Unlike single viewpoints and resolutions that are common in spatial remote sensing, ARS models are more likely to be overfitted with specific viewpoints and resolutions for specific downstream tasks. This overfitting reduces the ability of the model to generalize to different viewpoints and resolutions in various downstream tasks.

\begin{figure}
\centering
\includegraphics[width=0.95\linewidth]{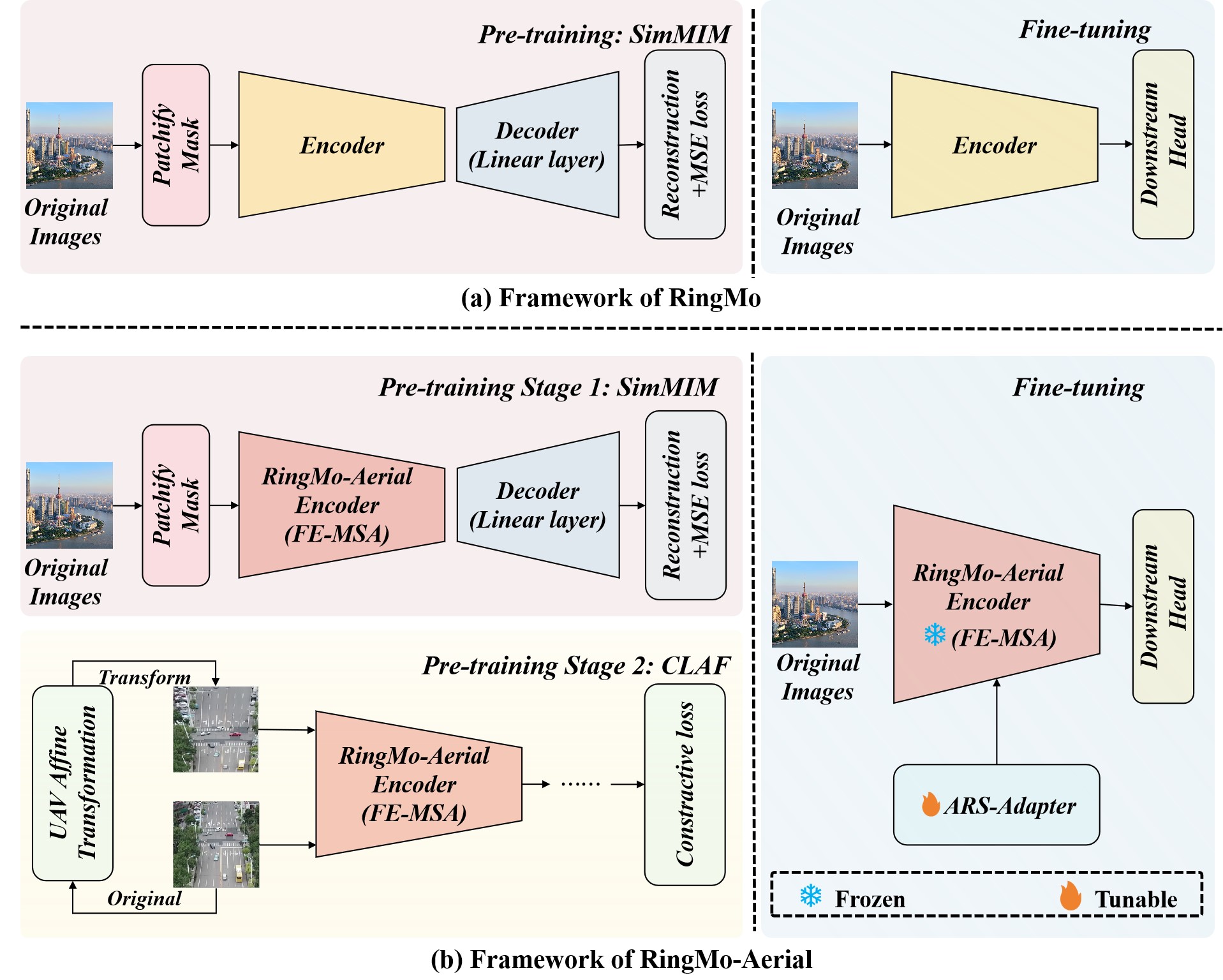}
\caption{Comparison of model frameworks between RingMo and RingMo-Aerial. (a) RingMo uses SimMIM for pre-training to perform image reconstruction. (b) RingMo-Aerial builds upon this by introducing affine transformation and contrastive learning, enhancing its ability to process multi-view images. In the fine-tuning stage, the ARS adapter is introduced to adjust a small number of parameters, further enhancing the model's generalization ability for downstream tasks.}
\label{fig:instructure-comp}
\end{figure}

To address the challenges above, we introduces RingMo-Aerial, the first foundational large model explicitly designed for ARS tasks. Building on the existing ApRS foundational model RingMo, RingMo-Aerial addresses the unique characteristics of ARS tasks through several novel approaches, as illustrated in Fig.\ref{fig:instructure-comp}.

First, we propose the Frequency-Enhanced Multi-Head Self-Attention (FE-MSA) module, designed to address the multi-scale and occlusion challenges inherent to oblique perspectives in ARS imagery. Unlike traditional multi-head attention mechanisms, FE-MSA includes an image patch expansion layer to capture dense, small objects as tokens for the Transformer model, leveraging Swin Transformer's window and shifted window attention. This enhances the model’s ability to detect densely packed and occluded objects.

%Second, to solve the problem of detecting and recognizing small distant targets in oblique perspectives, we propose a Contrastive Learning Affine Framework (CLAF) that scales and enlarges distant targets through affine transformations to construct positive samples for the contrastive learning pre-training framework. By inputting the transformed and original images as sample pairs into the model, this method guides the model to learn the unique visual characteristics of aerial remote sensing images, improving its ability to recognize small targets at a distance.

Second, to improve detection and recognition of distant small targets in oblique views, we propose the Contrastive Learning Affine Framework (CLAF). CLAF uses affine transformations to scale and enlarge distant targets, constructing positive samples to guide the model in learning ARS-specific visual features.

In addition, during the fine-tuning phase, we propose an efficient fine-tuning method for basic ARS foundational models called ARS-Adapter. Unlike traditional comprehensive fine-tuning methods, ARS-Adapter achieves efficient model fine-tuning by adjusting fewer than 5\%  of the parameters, significantly improving the training efficiency of downstream tasks while reducing the risk of overfitting. \textcolor{black}{Traditional efficient fine-tuning methods, such as Adapter and AdaptFormer, are insufficient in addressing the domain gap between self-supervised pretraining and downstream tasks. This domain gap arises from the inconsistency in feature extraction preferences between pretraining and task-specific objectives. ARS-Adapter mitigates this issue to a certain extent.} ARS-Adapter applies to all layers of the model: in the attention layers, it employs a traditional Adapter\cite{houlsby2019parameter} structure that operates parallel to the attention layers; in the Multi-Layer Perceptron (MLP) layers, it utilizes a bottleneck structure that includes MLP downsampling, deep convolutional feature extraction, and MLP upsampling. This approach optimizes the performance of the ARS foundational model, making it more adaptable and effective across various ARS vision tasks.

In summary, the contributions of this paper can be summarised as follows:

\begin{itemize}

\item \textbf{Addressing ARS-specific Characteristics}: We introduce RingMo-Aerial, the first foundational model for ARS, supporting multi-perspective, multi-resolution, and multi-modality tasks across various applications.

\item \textbf{FE-MSA Module}: By incorporating a patch expansion layer and integrating the window attention mechanism, FE-MSA effectively enhances the detection capability of small, dense targets in ARS images.

\item \textbf{CLAF Module}: CLAF uses affine transformations to capture ARS-specific visual features, particularly in oblique perspectives, improving the model's recognition of distant objects.

\item \textbf{ARS-Adapter}: The ARS-Adapter provides a fine-tuning approach tailored to ARS models, enhancing efficiency and reducing overfitting risks.

\end{itemize}

To evaluate the effectiveness of the model, we performed six downstream tasks using \textcolor{black}{12} ARS datasets and 5 ApRS datasets with different viewpoints, resolutions, and modes. The test results show that RingMo-Aerial achieves SOTA performance on all ARS datasets and reaches the leading level on ApRS datasets, which proves its strong generalization ability for intelligent interpretation of aerial and ApRS images. Compared with the single-task model, RingMo-Aerial has apparent advantages in handling multi-tasks in different scenarios.

\section{Related Work}
\subsection{Aerial remote sensing(ARS)}

ARS offers unique characteristics and applications across various domains, supporting tasks such as object detection, object tracking, semantic segmentation, and 3D-reconstruction. The most frequently utilized modalities are optical imaging \cite{deng2023towards}, infrared imaging \cite{zhu2023transformer}, and synthetic aperture radar (SAR) \cite{low2024multi}.

Compared to ApRS, ARS enables higher image resolution and flexible spatiotemporal coverage. However, it also presents distinct characteristics. Aerial imagery is typically captured at oblique angles with varying distances, leading to differences in target sizes and an increased likelihood of occlusion. For example, \cite{li2020yolo, li2023ogmn, wang2020robust, ye2023real} addressed target occlusion by modeling occluded objects at the structural level, significantly enhancing detection and segmentation performance. Studies such as \cite{liu2020small, tan2021yolov4_drone, huang2022ufpmp, yang2022querydet} have improved small target resolution by optimizing for oblique perspectives through multi-scale transformations.

Another critical difference lies in the frequency content of ARS imagery. Unlike ApRS images, which primarily capture low-frequency information, aerial images contain a mix of high-frequency and low-frequency signals, enabling detailed image interpretation. Recent work \cite{weng2024enhancing, li2024frequency} introduced frequency-domain enhancement modules to balance these signals, thereby improving performance across various ARS tasks.

\subsection{Visual foundation model}

Visual foundation models have gained attention as general-purpose solutions for computer vision tasks. Early models, based on convolutional neural networks (CNNs) like VGG \cite{simonyan2014very} and ResNet \cite{he2016deep}, used convolutional layers for feature extraction and were typically pre-trained on classification tasks using ImageNet \cite{deng2009imagenet}. With the advent of Transformers in natural language processing, visual Transformer models like ViT \cite{dosovitskiy2020image}, and Swin Transformer \cite{liu2021swin} have gained prominence, utilizing masked image modeling (MIM) or contrastive learning (CL) for self-supervised pre-training. In the field of remote sensing (RS), models such as RingMo \cite{sun2022ringmo} and SkySense \cite{guo2024skysense} have adopted the visual Transformer architecture and self-supervised pre-training strategies. \textcolor{black}{At this stage, methods integrating convolution into ViT architectures also began to emerge. Representative works such as CvT \cite{wu2021cvt}, Conformer \cite{peng2021conformer}, and CoaT \cite{xu2021co} enhance local feature modeling by embedding convolution into token projections, using parallel convolution-attention branches, or inserting convolution in early encoder stages. However, most of these approaches rely on stacked 3×3 convolutions, which suffer from a mismatch between the theoretical and effective receptive fields, limiting their ability to capture broader contexts or object-level structures \cite{luo2016understanding}.} Besides, several novel architectures, such as ConvNeXt \cite{liu2022convnet} and Mamba\cite{gu2023mamba}, have been proposed. However, these newer models have yet to achieve widespread adoption due to technical immaturity.

A major challenge for these models lies in the pre-training phase, where they learn broad visual representations from large datasets. This section provides an overview of three self-supervised training methods—contrastive learning, masked image modeling, and their combination—each addressing both natural and RS images.

\subsubsection{Contrastive Learning}

Contrastive learning is a self-supervised learning approach that emphasizes learning the standard features among similar instances while distinguishing differences between dissimilar cases. Primary contrastive learning methods include the MoCo \cite{he2020momentum, chen2020improved, chen2021empirical} series. MoCov2 \cite{chen2020improved}, an optimization of MoCo \cite{he2020momentum} inspired by SimCLR \cite{chen2020simple}, introduces enhancements such as the MLP projection head and increased data augmentation.  MoCov3 \cite{chen2021empirical} integrates MoCov2 and SimSiam \cite{chen2021exploring}, but training with ViT \cite{dosovitskiy2020image} as the backbone often leads to instability and reduced model accuracy. Stability and performance are improved by freezing the patch projection layer of ViT during training. DINO \cite{caron2021emerging}, another self-supervised learning method, employs knowledge distillation techniques to train ViT for semantic segmentation and classification feature learning without labels, leveraging momentum encoders and multi-crop training for enhanced model performance.

% In the field of remote sensing, comparative learning also has many applications. GLCNet\cite{li2022global} presents a self-supervised contrastive learning method for semantic segmentation of high-resolution remote sensing images. It employs global style contrastive learning modules and local matching contrastive learning modules to respectively learn global representations of images and representations of local regions, thereby enhancing pixel-level recognition capabilities in semantic segmentation tasks.SeCo\cite{manas2021seasonal} introduces a self-supervised pretraining method that leverages temporal information in remote sensing images to generate positive sample pairs, learning feature representations invariant and diverse to seasonal variations. The approach employs a design with multiple embedding subspaces to optimize features sensitive and insensitive to seasonal changes, enhancing the transferability of remote sensing images across various downstream tasks. In conclusion, contrastive learning serves as an effective unsupervised pre-training method with significant implications for large-scale pre-training of remote sensing image models.

In RS, contrastive learning also has many applications. GLCNet \cite{li2022global} presents a self-supervised contrastive learning method for semantic segmentation of high-resolution RS images. SeCo \cite{manas2021seasonal} introduces a self-supervised pre-training method that leverages temporal information in RS images to generate positive sample pairs, learning feature representations invariant and diverse to seasonal variations. 

% \begin{figure}
% \centering
% \includegraphics[width=1.0\linewidth]{images/token.jpg}
% \caption{Similar targets with significant size differences in the same picture. In the Vision Transformer model, objects of different
% sizes occupy different numbers of tokens. The image is from the VisDrone dataset.}
% \label{fig:img1}
% \end{figure}

% \begin{figure}
% \centering
% \includegraphics[width=0.9\linewidth]{images/problem-qingxie.jpg}
% \caption{Different shooting angles lead to variations in the final pixel sizes of similar targets in the image. When two identically sized cars are captured by a drone at different inclined angles, their differing vertical and horizontal distances result in inconsistent sizes at the pixel scale.}
% \label{fig:img2}
% \end{figure}

% \begin{figure}
% \centering
% \includegraphics[width=0.95\linewidth]{images/alpha-h.jpg}
% \caption{The same target will be represented differently in the image if its distance from the lens and imaging angle are different. As the horizontal distance increases, the number of pixels occupied by the height content increases and then decreases, and the height content occupies the most pixels at the horizontal distance $l = \sqrt{H(H - h)}$.}
% \label{fig:img2}
% \end{figure}

\subsubsection{Masked Image Modeling}

MIM is an emerging self-supervised pre-training method. It involves masking a portion of an input image and then reconstructing it through a pretext task. Unlike traditional supervised pre-training, which relies on labeled data, MIM uses self-supervised learning to learn meaningful representations from images. MAE \cite{he2022masked} pioneered the MIM by masking and reconstructing images using Vision Transformer (ViT). SimMIM \cite{xie2022simmim} differs by only reconstructing masked areas and incorporates the Swin Transformer \cite{liu2021swin}. Combining MAE and SimMIM, MixMAE \cite{liu2023mixmae} uses visible tokens from different images for reconstruction. GreenMIM \cite{huang2022green} optimizes MAE by introducing a layered architecture to lower computational costs. UM-MAE \cite{li2022uniform} applies uniform masking to address complexities in ViTs, while CAE \cite{chen2024context} adds a regressor for distinct feature encoding separate from reconstruction.

In RS, RingMo \cite{sun2022ringmo} pioneered the foundation model framework based on MIM. Unlike random masking techniques used in natural scenes, RingMo developed the PIMask strategy tailored to the complex backgrounds and small targets characteristic of RS data, focusing on preserving information about small targets. By integrating ViT and Swin Transformer, RingMo enhanced the accuracy of various downstream RS tasks, laying a foundation for developing large models in RS scenes.

\subsubsection{The combination of Contrastive Learning $\&$ Masked Image Modeling}
% for example: CMAE
Combining CL and MIM has shown significant advantages in current research, particularly in enhancing feature extraction performance. For instance, CMAE \cite{huang2023contrastive} utilizes an asymmetric encoder-decoder structure along with a momentum encoder. This enables learning visual representations from masked images and enhances feature discriminability through CL. Additionally, MimCo \cite{zhou2022mimco} introduces patch-level and image-level contrastive reconstruction losses in MIM, effectively improving the model's transferability. 

In the field of RS, TMAC \cite{cao2023transformer} and SegMind \cite{li2023segmind} demonstrate that combining MIM and CL significantly improves the utilization of unlabeled data. It also enhances the interaction between different image regions, boosting feature recognition capabilities.

Despite the success of models like ViT and RingMo in general and satellite remote sensing tasks, their performance in ARS remains limited. Currently, there is no dedicated foundation model for ARS. To address this gap, we propose RingMo-Aerial, a model designed specifically for ARS tasks, and demonstrate its effectiveness in typical applications.

\section{Problem Description}

\begin{figure}
\centering
\includegraphics[width=1.0\linewidth]{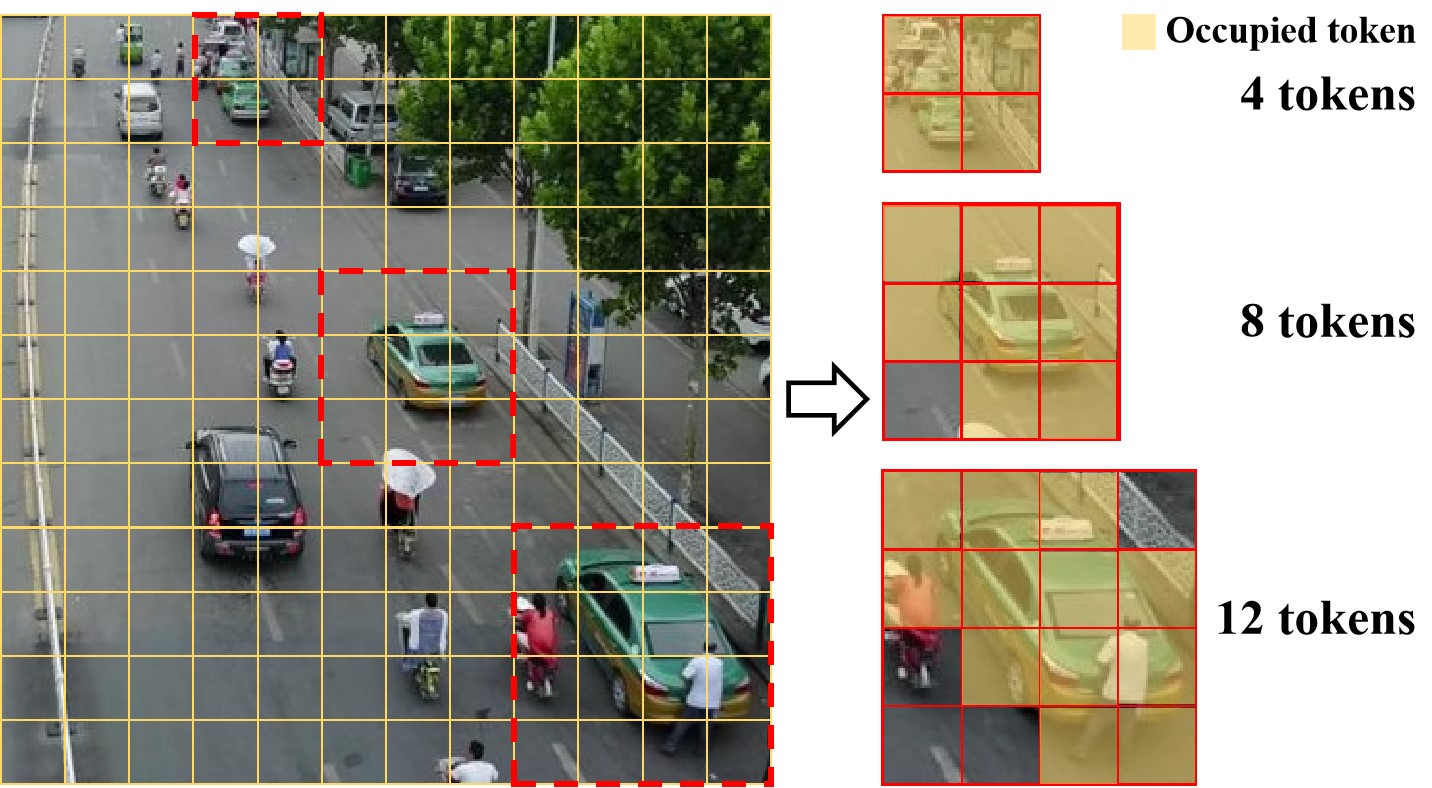}
\caption{Similar targets with significant size differences in the same picture. In the Vision Transformer model, objects of different
sizes occupy different numbers of tokens. The image is from the VisDrone dataset. \textcolor{black}{The yellow box represents the tokens occupied by the target}.}
\label{fig:img1}
\end{figure}

The unique imaging properties of ARS photography stem from the inherent properties of aerial perspective. Specifically, this perspective manifests itself in size differences within the same image due to the tilted angle of the shot. As shown in Fig.\ref{fig:img1}, similar objects at different distances from the ARS are rendered at different sizes in the image. \textcolor{black}{This phenomenon results in distant objects occupying fewer tokens in the Vision Transformer’s input sequence, while closer objects occupy more.}

\begin{figure}
\centering
\includegraphics[width=0.8\linewidth]{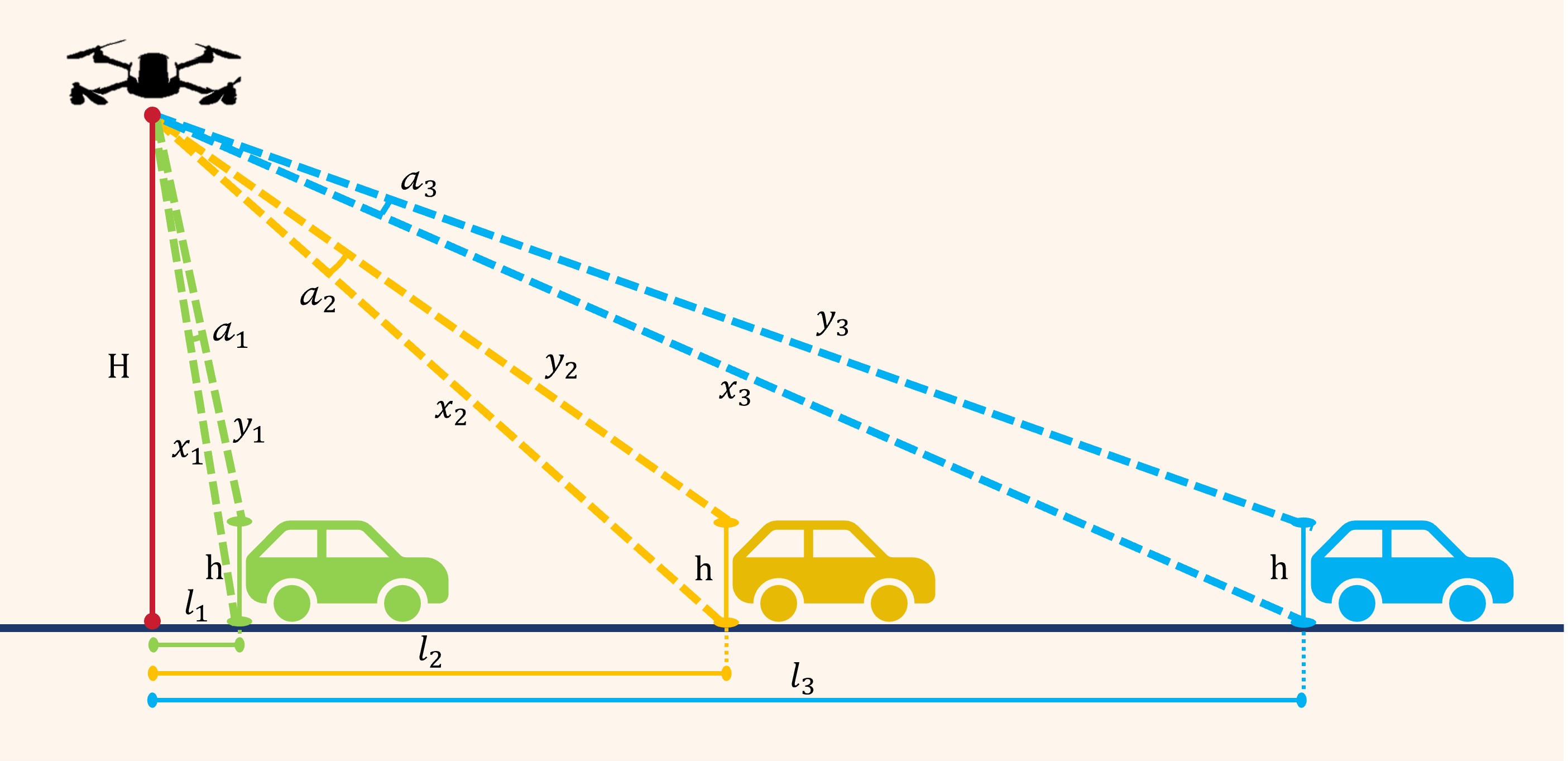}
\caption{The same target will be represented differently in the image if its distance from the lens and imaging angle are different. As the horizontal distance increases, the number of pixels occupied by the height content increases and then decreases, and the height content occupies the most pixels at the horizontal distance $l = \sqrt{H(H - h)}$.}
\label{fig:img2}
\end{figure}

As shown in Fig.\ref{fig:img2}, when the camera is positioned at a height \( H \) and the target object has a height \( h \), 
%\textcolor{red}{the angular size \( \alpha \) of the target's height in the camera's field of view varies as a function of the horizontal distance \( l \) between the target and the camera.}
\textcolor{black}{the angular size \( \theta \) of the target's height in the camera's field of view varies as a function of the horizontal distance \( l \) between the target and the camera.}At different horizontal distances, the imaging size corresponding to the same target height differs. Specifically, when the target is directly beneath the camera, its height is not captured in the image. As the horizontal distance increases, the number of pixels occupied by the target's height initially increases and then decreases. By considering the distances from the target's lowest and highest points to the camera, given by \( x = \sqrt{H^2 + l^2} \) and \( y = \sqrt{(H - h)^2 + l^2} \) respectively, we can derive the relationship between the viewing angle \( \alpha \) and the horizontal distance \( l \) between the target and the camera:

% Using the cosine rule:
% \textcolor{red}{
% \begin{equation}\label{eq:1}
%   x^2 + y^2 - 2xy \cos \alpha = h^2
% \end{equation}
% }

\textcolor{black}{
\begin{equation}\label{eq:1}
   x^2 + y^2 - 2xy \cos \theta = h^2
\end{equation}
}

Substituting the values of \(x\) and \(y\), The final cosine formula becomes:
% \textcolor{red}{
% \begin{equation}\label{eq:2}
%   \cos \alpha = \frac{H^2 - Hh + l^2}{\sqrt{H^2 + l^2} \cdot \sqrt{(H - h)^2 + l^2}}
% \end{equation}
% }

\textcolor{black}{
\begin{equation}\label{eq:2}
   \cos \theta = \frac{H^2 - Hh + l^2}{\sqrt{H^2 + l^2} \cdot \sqrt{(H - h)^2 + l^2}}
\end{equation}
}

By setting \(a = H\) and \(b = H - h\), we have the target function is:
% \textcolor{red}{
% \begin{equation}\label{eq:3}
%   f(l) = \cos \alpha = \frac{ab + l^2}{\sqrt{a^2 + l^2} \cdot \sqrt{b^2 + l^2}}
% \end{equation}
% }

\textcolor{black}{
\begin{equation}\label{eq:3}
   f(l) = \cos \theta = \frac{ab + l^2}{\sqrt{a^2 + l^2} \cdot \sqrt{b^2 + l^2}}
\end{equation}
}

Taking the derivative of \(f(l)\):
\begin{equation}\label{eq:4}
   f'(l) = \frac{2l(a^2 + l^2)(b^2 + l^2) - l(ab + l^2)(a^2 + b^2 + 2l^2)}{(a^2 + l^2)^{3/2}(b^2 + l^2)^{3/2}}
\end{equation}

Solving \(f'(l) = 0\) results in the equation:
\begin{equation}\label{eq:5}
   2(a^2 + l^2)(b^2 + l^2) = (ab + l^2)(a^2 + b^2 + 2l^2)
\end{equation}

Simplifying further:
\begin{equation}\label{eq:6}
   2a^2b^2 + a^2l^2 + b^2l^2 = a^3b + ab^3 + 2ab l^2
\end{equation}

Combining terms:
\begin{equation}\label{eq:7}
   l^2(a - b)^2 = ab(a - b)^2
\end{equation}

Since \(a = H\) and \(b = H - h\), and \(a > b\), it follows that \(a - b = h > 0\).

When \(f'(l) = 0\), we obtain:
\begin{equation}\label{eq:8}
   l^2 = ab
\end{equation}

Therefore:
\begin{equation}\label{eq:9}
   l = \sqrt{ab} = \sqrt{H(H - h)}
\end{equation}

% \textcolor{red}{
% This completes the derivation of the maximum value $\alpha$ and the corresponding minimum value \(\cos\alpha\). From Eq. (2), when $l = 0$, $\alpha$ is equal to 0. From Eqs.(4) and (9), $\alpha$ increases with $l$ is less than $\sqrt{H(H - h)}$, $\alpha$ is maximum when $l = \sqrt{H(H - h)}$, and decreases with $l$ when $l$ is greater than $\sqrt{H(H - h)}$, and decreases with $l$ in the case of $l \rightarrow \infty$  when $\alpha$ is again equal to 0. Assuming that the optical axis of the camera is always at the center of the target viewpoint $\alpha$, the pixel change of the target height information in the picture is the same as the change of the corresponding viewpoint $\alpha$ of the target. This indicates that if the same target is at different distances from the lens and imaging angles, it will have different representations in the image.
% }

\textcolor{black}{
This completes the derivation of the maximum value $\theta$ and the corresponding minimum value \(\cos\theta\). From Eq. (2), when $l = 0$, $\theta$ is equal to 0. From Eqs.(4) and (9), $\theta$ increases with $l$ is less than $\sqrt{H(H - h)}$, $\theta$ is maximum when $l = \sqrt{H(H - h)}$, and decreases with $l$ when $l$ is greater than $\sqrt{H(H - h)}$, and decreases with $l$ in the case of $l \rightarrow \infty$  when $\theta$ is again equal to 0. Assuming that the optical axis of the camera is always at the center of the target viewpoint $\theta$, the pixel change of the target height information in the picture is the same as the change of the corresponding viewpoint $\theta$ of the target. This indicates that if the same target is at different distances from the lens and imaging angles, it will have different representations in the image.
}

% According to the imaging principle, the pixel length and width of the object are inversely proportional to the distance between the object and the camera. The ratio of the distance between Object 1 and Object 2 and the UAV's camera in the Figure can be given by the sine function as \ref{eq:1}:

% \begin{equation}\label{eq:1}
% \setlength{\abovecaptionskip}{1pt}
% \setlength{\belowcaptionskip}{1pt}
% \frac{d_1}{d_2} = \frac{\sin{\theta_2}}{\sin{\theta_1}}
% \end{equation}

% Then the length ratio of Object 1 and Object 2 is as \ref{eq:2}:

% \begin{equation}\label{eq:2}
% \setlength{\abovecaptionskip}{1pt}
% \setlength{\belowcaptionskip}{1pt}
% \frac{l_1}{l_2} = \frac{d_2}{d_1} = \frac{\sin{\theta_1}}{\sin{\theta_2}}
% \end{equation}

% Then the ratio of pixels occupied by Object 1 and Object 2 is as \ref{eq:3}:

% \begin{equation}\label{eq:3}
% \setlength{\abovecaptionskip}{1pt}
% \setlength{\belowcaptionskip}{1pt}
% \frac{p_1}{p_2} = \frac{l_1^2}{l_2^2} = \frac{\sin^2{\theta_1}}{\sin^2{\theta_2}}
% \end{equation}

\begin{figure*}
\centering
\includegraphics[width=0.85\linewidth]{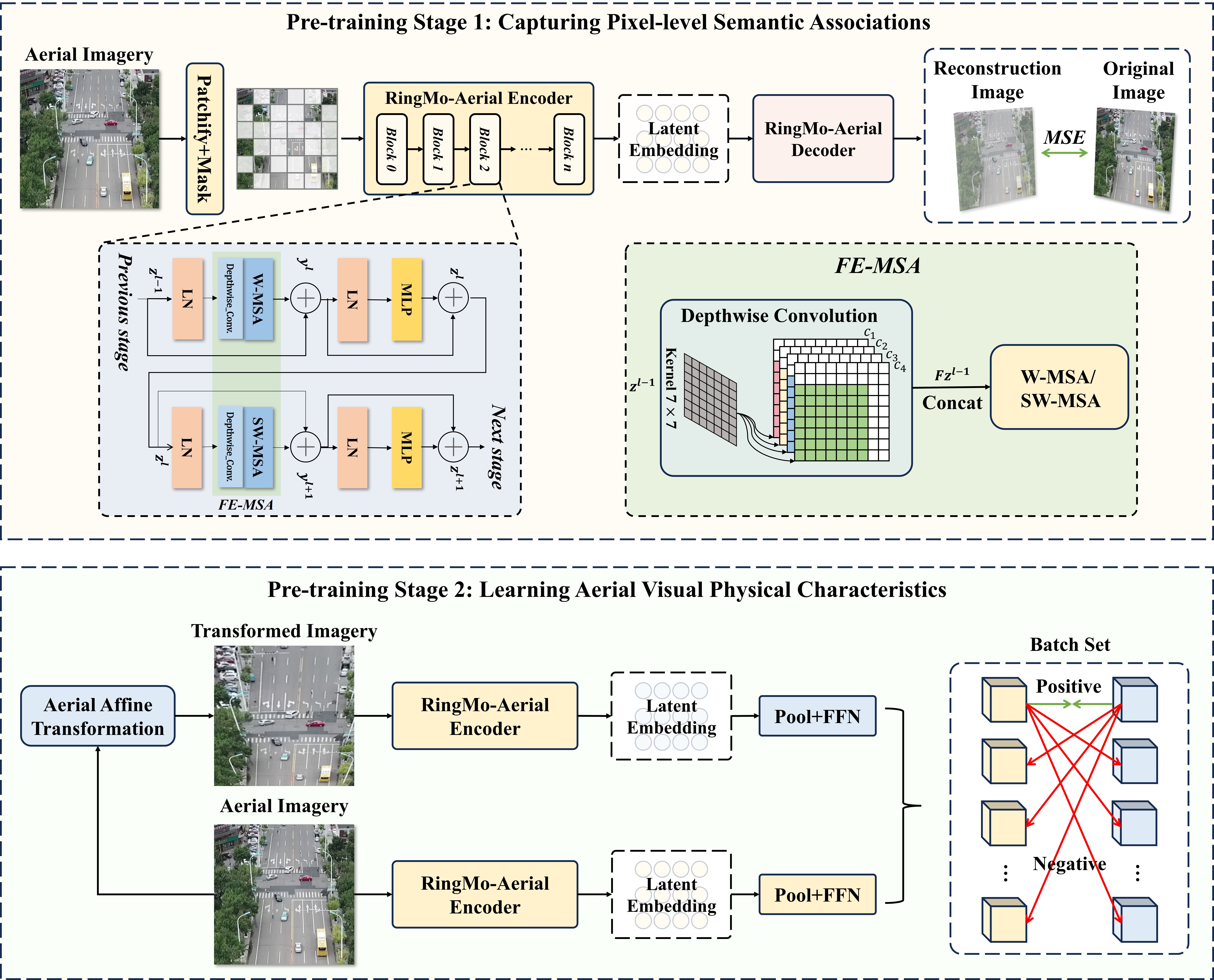}
\caption{RingMo-Aerial's two-stage pre-training principles Capturing Pixel-level Semantic Associations and Learning Aerial Visual Physical Characteristics, and the architecture of the FE-MSA module.}
\label{fig:structre}
\end{figure*}

In the visual transformer model, the number of pixels occupied by an object is related to the number of tokens it occupies, as Fig.\ref{fig:img1}. Therefore, it is helpful to adopt a method to balance the number of tokens occupied by similar objects of different sizes in an image.

In addition to the oblique angles, ARS images typically contain many small objects, which can be very dense. CNNs have fixed convolution kernel sizes and limited receptive fields, which makes it challenging to capture large-scale information and detailed information in ARS images simultaneously. This can result in missed detections and false detections of objects. For instance, when Sensors capture congested urban streets, pedestrians and vehicles may randomly occlude other objects, making it challenging for detection algorithms to locate and classify them accurately. ViT employs the same downsampling rate and directly extracts features from image tokens, which fails to account for the spatial context of objects.

Furthermore, ARS images are typically high resolution, containing many intricate details and complex background information. This places greater demands on pre-trained models' feature extraction and processing capabilities. Traditional pre-trained models are typically trained on large-scale natural scene datasets, which cannot adequately represent the characteristics of ARS images.

Consequently, a practical pre-training framework and method, specifically designed to address the distinctive visual characteristics of ARS images, is essential to achieve satisfactory performance in a range of downstream tasks.

\section{Methodology}

% The proposed foundation model, called RingMo-Aerial, is developed based on Swin Transformer \cite{liu2021swin}. 

% \textbf{introduction to Swin Transformer}

% \textbf{introduction to downstream tasks}

\subsection{Method Overview}

The proposed RingMo-Aerial serves as a foundational framework with broad applicability. Its implementation involves three main stages: model design, pre-training, and fine-tuning for downstream tasks. In Section \ref{sec:arch}, we address the challenge of detecting small objects in ARS imagery caused by oblique viewing angles during the model design phase. To enhance small object detection, we introduce the Frequency-Enhanced Multi-Head Self-Attention (FE-MSA) layer by incorporating a patch expansion layer, shown in Fig.\ref{fig:structre}. In Section \ref{sec:pre}, we combine the commonly used masked image modeling (MIM) pre-training with additional contrastive learning (CL) methods based on affine transformations during the pre-training stage. This hybrid pre-training approach significantly improves both training speed and the performance of the converged model on downstream tasks, shown in Fig.\ref{fig:structre}. Furthermore, in Section \ref{sec:ada}, we introduce a parallel Adapter fine-tuning method during the fine-tuning phase. The proposed ARS-Adapter module includes downsampling, channel-wise convolution, and upsampling components, enhancing its adaptability and learning capacity for specific tasks.

\subsection{Model Architecture}
\label{sec:arch}

RingMo-Aerial's design is based on the RingMo. The architecture of the Swin Transformer consists of four stages, each comprising either a linear encoder or Patch Merging, along with an even number of Swin Transformer Blocks. Each block includes Window Multi-Head Self-Attention (W-MSA) and Shifted Window Multi-Head Self-Attention (SW-MSA). These attention mechanisms effectively compute attention maps within the windows, extracting spatial feature information between pixels while reducing the computational complexity of the model.

Although the classical Swin Transformer introduces spatial interactions through two types of window attention mechanisms, emphasizing high-level semantic information within feature vectors, challenges arise in ARS imagery. The varying shooting distances, angles, and scale differences between distant and nearby objects present significant difficulties. Relying solely on semantic information may not be sufficient to achieve optimal performance in downstream tasks.

To address the challenges above, we propose the overall architecture of RingMo-Aerial. As shown in Fig.\ref{fig:structre} Pre-training Stage 1, considering the unique characteristics of the ARS perspective and based on the window attention mechanisms (W-MSA and SW-MSA), a new attention mechanism called Frequency-Enhanced Multi-Head Self-Attention (FE-MSA) is introduced to enhance the model's capability in handling diverse aerial scenarios. \textcolor{black}{It is an extension to standard Multi-Head Self-Attention (MSA), defined as:}

\textcolor{black}{
\begin{align}
\mathrm{FE\text{-}MSA}(X) &= \mathrm{MSA}(Y_{\mathrm{conv}}) \nonumber \\ 
&= \mathrm{MSA}\bigl(X + \mathrm{DWConv}_{7\times7}(X)\bigr).
\end{align}
}

\textcolor{black}{The input $X$ is first processed through a residual depth-wise $7\times7$ convolution branch, which expands the local receptive field and compensates for high-frequency components. The enhanced feature map $Y_{\mathrm{conv}}$ is then fed into the MSA layer to enable global information interaction.}

% As shown in Figure \ref{fig:structre}(a), the overall architecture of RingMo-Aerial is based on the Swin Transformer. The classic Swin Transformer introduces spatial interactions through window attention mechanisms. However, it primarily emphasizes high-level semantic information within feature vectors. In drone aerial images, varying shooting distances, angles, and scale differences between distant and nearby objects pose challenges. Relying solely on semantic information may not consistently yield optimal performance in downstream tasks.

% To address this limitation, we introduce the FE-MSA module. This module captures edge features and texture information within feature vectors, enhancing the model's ability to handle various aerial scenes.

\begin{figure}
\centering
\includegraphics[width=0.90\linewidth]{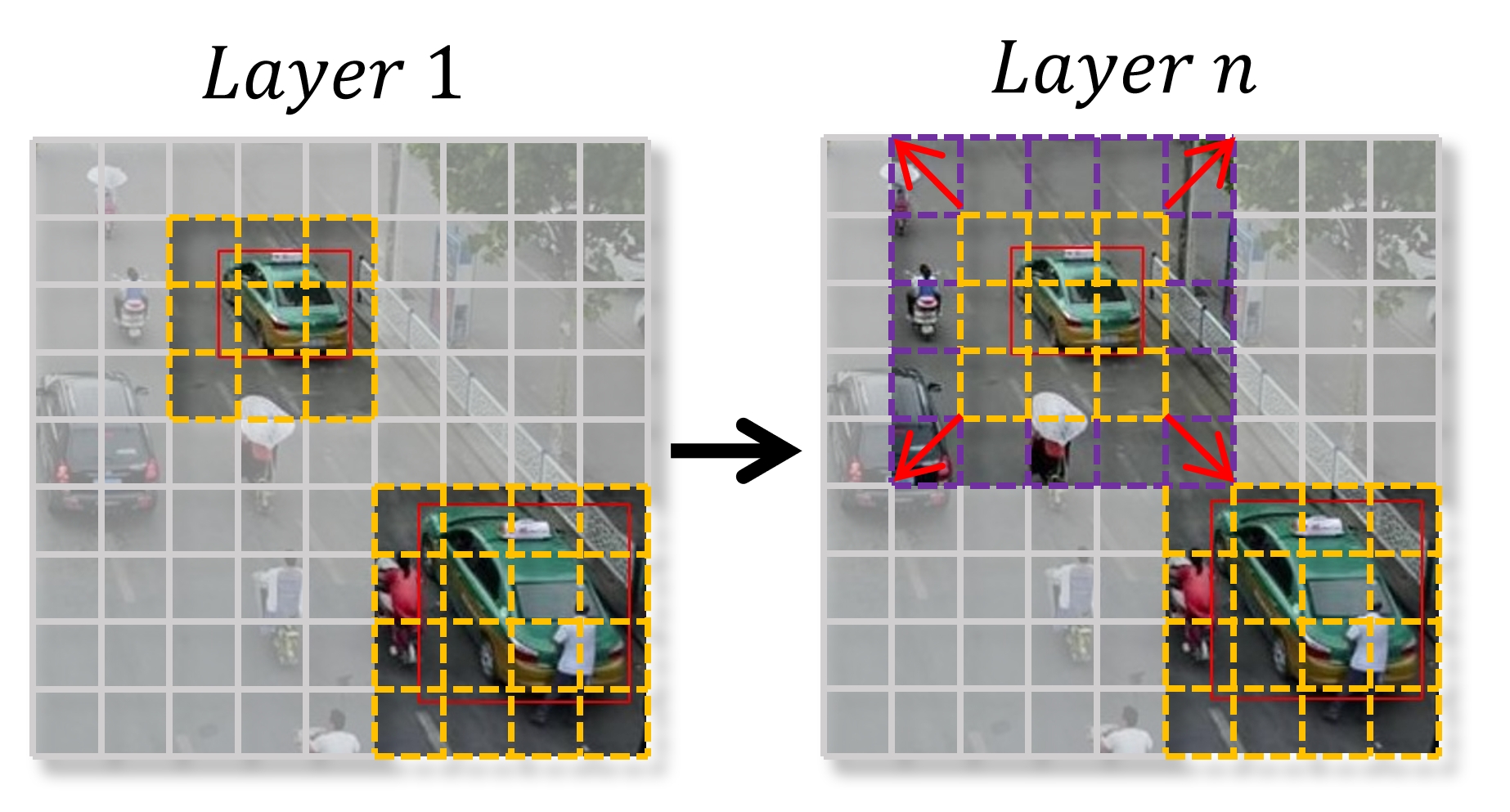}
\caption{The effect of FE-MSA. Existing window attention mechanisms cannot balance the feature vector information of objects of different sizes in the image. At the same time, FE-MSA enhances the model's receptive field for small objects (purple box).}
\label{fig:ganshouye}
\end{figure}

\textbf{Frequency-Enhanced Multi-Head Self-Attention (FE-MSA) module.}
To better understand the FE-MSA mechanism, it is essential first to recognize the limitations of the existing window attention mechanisms. These mechanisms often struggle to balance the semantic information in feature vectors of objects of different sizes. For instance, as illustrated in Fig.\ref{fig:ganshouye}, larger objects may occupy $4\times4$ tokens, while smaller distant objects may occupy only $3\times3$ tokens. The FE-MSA mechanism introduces frequency domain processing to the existing window attention, indirectly expanding the receptive field for smaller objects.

\begin{figure*}
\centering
\includegraphics[width=0.90\linewidth]{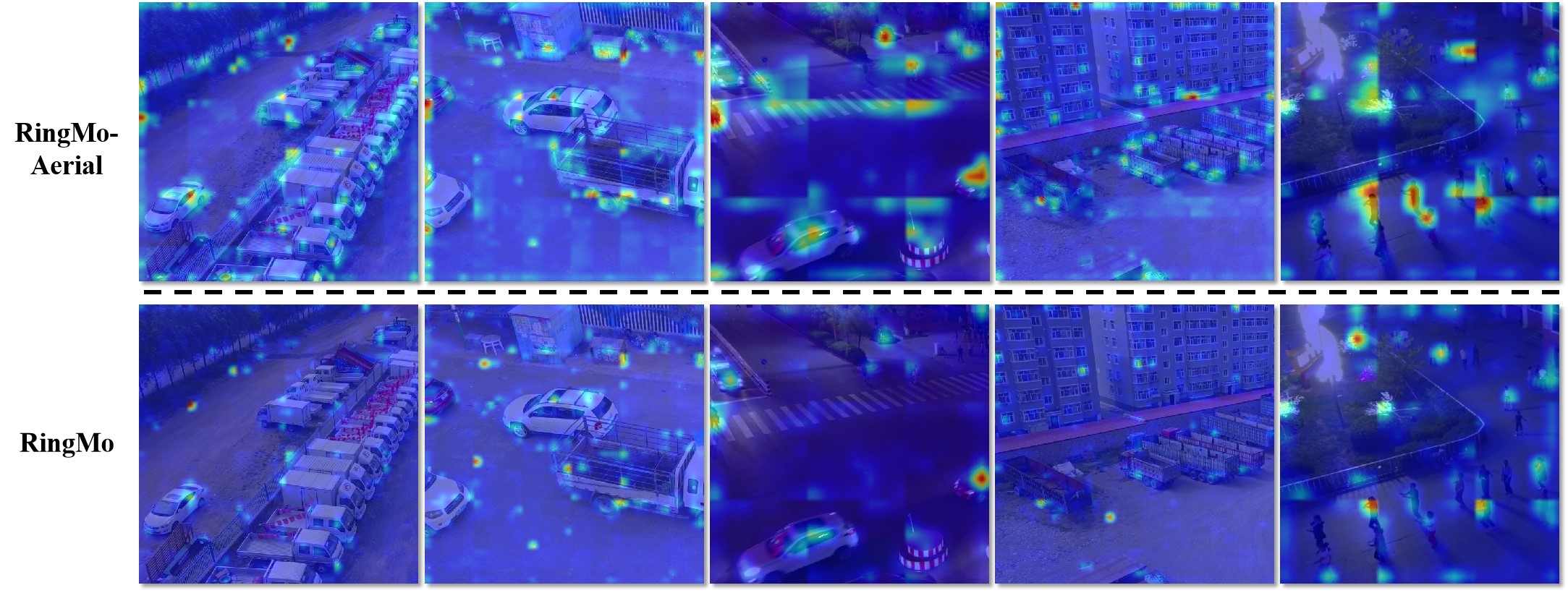}
\caption{Comparative analysis of feature heatmaps extracted from multi-view aerial imagery using RingMo-Aerial and RingMo. The feature heatmaps extracted by RingMo-Aerial are located in the upper half, while those from the RingMo model are in the lower half of the figure. In these images, the brighter the region, the more the model is focusing on that particular feature, indicating that these areas have higher feature importance for the model.}
\label{fig:relitu}
\end{figure*}

\textcolor{black}{The FE-MSA mechanism enables RingMo-Aerial to effectively capture edge and texture features of different targets in ARS images, improving the model's performance in complex aerial scenes. Fig.\ref{fig:relitu} presents a comparative analysis of the feature extraction capabilities of RingMo-Aerial and RingMo models in processing multiview ARS images. The heatmaps generated by RingMo-Aerial show more focused attention on critical areas, such as vehicles and buildings, as indicated by the bright regions. In contrast, the RingMo in the lower half has a more diffuse focus, with less concentration on specific features. This comparison demonstrates the superior ability of RingMo-Aerial to handle complex aerial scenes, enabling better feature extraction and object recognition in multiview ARS images. These results visually confirm the improvements introduced by RingMo-Aerial, particularly in addressing challenges such as varying scales and perspectives in aerial imagery.}

\textcolor{black}{In ARS tasks such as object detection and semantic segmentation, small targets are far more prevalent and exhibit more extreme scale variation than in natural or satellite imagery. Standard Vision Transformer MSA or Swin Transformer Window Attention limits each token’s initial receptive field to a single patch, severely hindering edge and detail capture for small targets. Introducing an additional convolutional branch serves to (1) spatially expand the receptive field and (2) bolster the model’s ability to focus attention on fine-scale structures. From a frequency-domain perspective, small targets manifest stronger high-frequency energy, yet attention mechanisms tend to attenuate such components. By inserting a residual $7\times7$ depth-wise separable convolution prior to attention, FE-MSA enhances band-pass filtering in the high-frequency range, improving detail modeling for small targets.}

Fig.\ref{fig:structre}(b) illustrates the RingMo-Aerial Block structure, including two window attention mechanisms optimized by the FE-MSA mechanism. Given \( z^{(l-1)} \in \mathbb{R}^{\frac{W}{4} \times \frac{H}{4} \times C} \), where \( l \) is the id of each RingMo-Aerial Block. 
% \textcolor{red}{Taking the first stage as an example, the input feature vector \( z^l \) adopts the same slicing parameters as the standard Swin-Transformer, resulting in four-channel slices \( C = \{c_1, c_2, c_3, c_4\} \). After linear normalization, it enters the FE-MSA module. For each patch with a dimension of \( C \), a \( 7 \times 7 \) convolution kernel is used for depthwise convolution to obtain the feature map \( Fz^{(l-1)} = \text{Concat}\{Fc_1, Fc_2, Fc_3, Fc_4\} \).}
\textcolor{black}{Taking the first stage as an example, the input feature vector \( z^{(l-1)} \) adopts the same partitioning strategy as the standard Swin Transformer. After linear normalization, the feature is passed into the FE-MSA module. In this module, all channels are independently processed using a shared \( 7 \times 7 \) depthwise convolution kernel, and the resulting outputs are concatenated along the channel dimension.} Then, W-MSA or SW-MSA is used to compute the attention map. After the FE-MSA, small objects' edge and texture features are further enhanced, improving the model's ability to perceive local and global features. The computation process for each token patch through the RingMo-Aerial block is as follows:

\begin{equation}
z^l = \text{FE-MSA}(z^{l-1})
\end{equation}

\begin{equation}
\text{FE-MSA}(z^{l-1}) = \text{SoftMax}\left(\frac{qk^T}{\sqrt{d}} + B\right)v
\end{equation}

\textcolor{black}{
In the $l$-th RingMo-Aerial Block, the input feature map $z^{l-1}\in\mathbb{R}^{\frac{W}{4}\times\frac{H}{4}\times C}$ is first enhanced via a depthwise convolution in the frequency domain, after which multi-head self-attention is performed within each window: queries $q$, keys $k$, and values $v$ are obtained by linear projections of the enhanced features, their dot-products are scaled by $\sqrt{d}$, augmented with a relative position bias $B$, and normalized by SoftMax, then subsequently weighted-summed with $v$. 
% This design preserves the Swin Transformer’s efficient spatial interactions while, through frequency-domain processing, reinforcing the model’s sensitivity to edge and texture details of small-scale objects, thereby improving feature extraction in complex aerial imagery.
}

\subsection{Pre-training}
\label{sec:pre}

% In the context of UAV imagery, open-source datasets are scarce and challenging to acquire. To address this limitation, we construct a large-scale pre-training dataset by leveraging the unique characteristics of aerial images captured from UAVs. Specifically, we introduce an affine transformation-based data augmentation technique. This method involves introducing random affine transformation coefficients to adjust the tilt angle of UAV images by stretching the upper and lower edges, resulting in a new image that differs from the original data.

% \begin{figure}
% \centering
% \includegraphics[width=1.0\linewidth]{images/cl_demo.jpg}
% \caption{Contrastive Learning Methods Simple Schematic.}
% \label{fig:cl_demo}
% \end{figure}

% \textbf{introduction to affine transformation-based data augmentation technique, need an image}

% Our comprehensive pre-training dataset covers a wide range of aerial scenes, including high-altitude, low-altitude, and near-ground perspectives. It spans diverse geographical regions such as urban, rural, and wilderness areas. The dataset includes various targets, such as roads, vehicles, and pedestrians, and captures different seasons and periods. Notably, it exhibits multi-source, multi-temporal, and multi-instance characteristics. Before training, we crop the entire dataset into 640x640 or 448x448 resolution slices, all captured using optical three-channel imaging. In total, the sliced dataset comprises approximately 5.4 million images.

\begin{figure}
\centering
\includegraphics[width=1.0\linewidth]{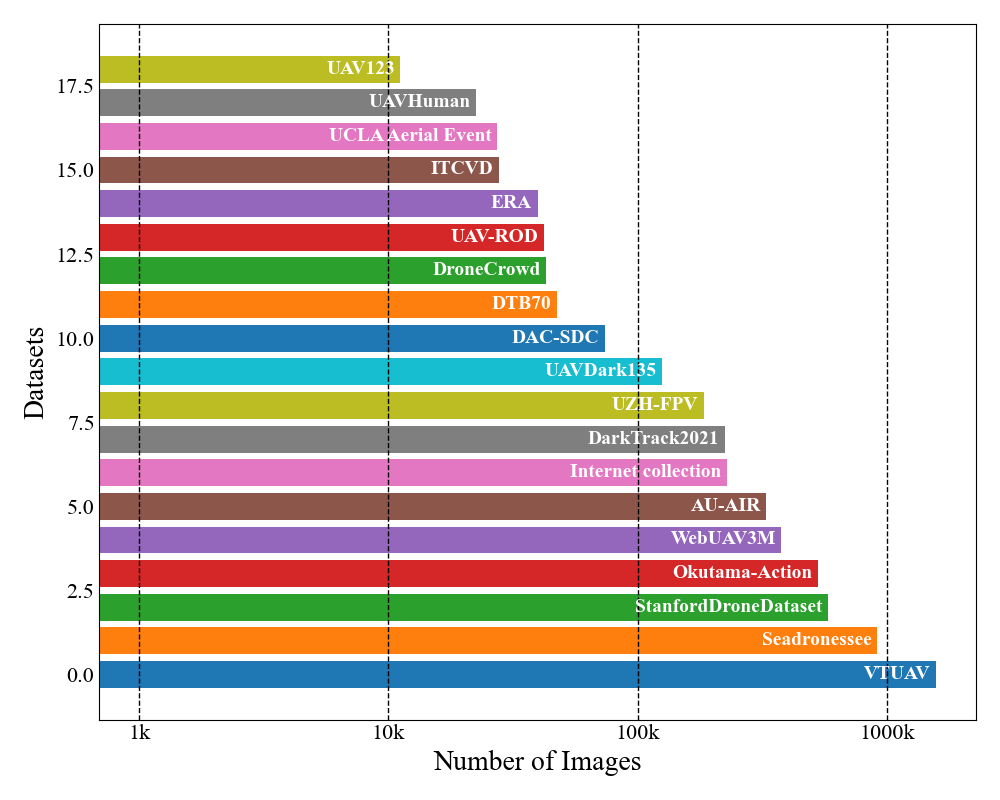}
\caption{The figure shows the number of slices corresponding to the slicing process performed on multiple collected open source airborne remote sensing datasets, which were used as the dataset for pre-training of the base model.}
\label{fig:dataset_show}
\end{figure}

For pre-training, we construct a large-scale pre-training dataset. Our comprehensive pre-training dataset covers a wide range of aerial scenes, including high-altitude, low-altitude, and near-ground perspectives. It spans diverse geographical regions such as urban, rural, and wilderness areas. The dataset includes various targets, such as roads, vehicles, and pedestrians, and captures different seasons and periods. Notably, it exhibits multi-source, multi-temporal, and multi-instance characteristics. Before training, we crop the entire dataset into $640 \times 640$ or $448 \times 448$ resolution slices, all captured using optical three-channel imaging. The sliced dataset comprises approximately 5.4 million images (Fig.\ref{fig:dataset_show}).

During the pre-training phase, we combine Masked Image Modeling (MIM) and Contrastive Learning (CL) techniques. For MIM training, we use the standard SimMIM pre-training framework. The primary goal of this approach is to enhance the model's ability to capture pixel-level semantic associations, thus initially equipping the model with feature extraction capabilities. However, as analyzed previously, solely using the MIM pre-training method for ARS foundational model training is insufficient for the model to learn the intrinsic object characteristics specific to ARS vision. Therefore, building upon the MIM pre-training, we propose a contrastive learning framework based on affine transformations. This involves generating positive and negative image pairs through affine transformations, which serve as the basis for the contrastive learning pre-training of the model.

The process of affine transformation of an image is completed through perspective transformation. Perspective transformation first defines the source point and the destination point, which are defined as:

\begin{equation}
\label{formula:7}
\text{src\_points} = \begin{bmatrix}
x_0 & y_0 \\
x_1 & y_1 \\
x_2 & y_2 \\
x_3 & y_3
\end{bmatrix}, \quad
\text{dst\_points} = \begin{bmatrix}
u_0 & v_0 \\
u_1 & v_1 \\
u_2 & v_2 \\
u_3 & v_3
\end{bmatrix}
\end{equation}

The transformation relationship between the source point and the destination point used in the proposed method is:

\begin{equation}
\label{formula:8}
\text{dst\_points} = \begin{bmatrix}
u_0 & v_0 \\
u_1 & v_1 \\
u_2 & v_2 \\
u_3 & v_3
\end{bmatrix}
 = 
\begin{bmatrix}
x_0 & y_0 \\
x_1 & y_1 \\
\alpha x_2 + (1 - \alpha) x_3 & y_2 \\
\alpha x_3 + (1 - \alpha) x_2 & y_3
\end{bmatrix}
\end{equation}

Among them, $\alpha$ is the transformation coefficient, and the value range is randomly between 0.05 and 0.35. On this basis, the perspective transformation matrix can be calculated. The perspective transformation matrix is defined as a $3 \times 3$ matrix in the form of:

\begin{equation}
M = \begin{bmatrix}
m_{11} & m_{12} & m_{13} \\
m_{21} & m_{22} & m_{23} \\
m_{31} & m_{32} & 1
\end{bmatrix}
\end{equation}

The solution process of this transformation matrix is as follows:

\begin{equation}
\begin{cases}
u_i = \frac{m_{11} x_i + m_{12} y_i + m_{13}}{m_{31} x_i + m_{32} y_i + 1} \\
v_i = \frac{m_{21} x_i + m_{22} y_i + m_{23}}{m_{31} x_i + m_{32} y_i + 1}
\end{cases} ~ \text{for} ~ i = 0, 1, 2, 3
\end{equation}

Based on formulas (\ref{formula:7}) and (\ref{formula:8}), the above transformation matrix can be solved by the least square method. After obtaining the transformation matrix, the point $(x, y)$ on the source image can be mapped to the point $(u, v)$ on the destination image:

\begin{equation}
\begin{bmatrix}
u \\
v \\
1
\end{bmatrix}
= M \cdot
\begin{bmatrix}
x \\
y \\
1
\end{bmatrix}
\end{equation}

\begin{figure}
\centering
\includegraphics[width=1.0\linewidth]{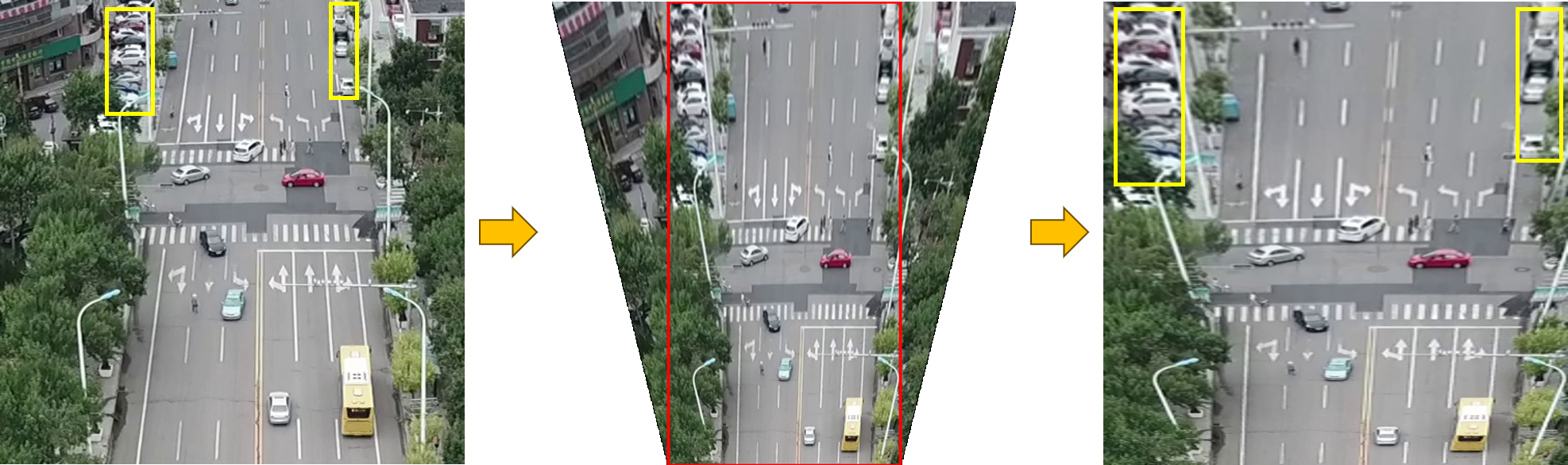}
\caption{Positive samples for contrastive learning are obtained after the image's affine transformation, cropping, and resizing.}
\label{fig:cl}
\end{figure}

\textcolor{black}{
When generating positive sample pairs via affine transformation, we first apply a random coefficient $\alpha \in [0.05, 0.35]$ to compress the bottom width of the image to a specified proportion of its original size while preserving the top width, thereby creating a pixel-level “magnification” effect for distant small objects. The transformed image is then center-cropped to a rectangular region and rescaled to match the model’s input dimensions, effectively removing irrelevant background and further enhancing the visibility of small objects. By sampling $\alpha$ from a diversified range, we achieve multi-scale coverage—from subtle magnification to global micro-distortion—which enables the model to learn semantically consistent embeddings across varying perspectives and spatial resolutions, thereby improving its ability to perceive distant small targets in aerial imagery. For negative samples, we select all full, transformed images from other instances within the same mini-batch.}

\textcolor{black}{
Specifically, as illustrated in Figure \ref{fig:cl}, positive samples are generated by first applying an affine transformation that warps the image into an isosceles trapezoid while keeping the top width unchanged. This operation compresses the lower part of the image and stretches the upper part. The transformed image is then cropped to retain only the central rectangular region, which is subsequently resized to form the final positive sample. Although this procedure discards certain areas, it effectively amplifies the upper region of the image—i.e., distant small targets in aerial scenes. Incorporating such positive samples into contrastive pre-training helps the model to better capture scale-invariant features and geometric invariance, thereby improving its ability to detect small objects under complex aerial viewpoints.}

\textcolor{black}{
We obtain the source image \( I_i \) and its transformed counterpart \( I_i' \) via the above affine process. The pair \( (I_i, I_i') \) constitutes a positive sample, while the pairs \( (I_i, I_j') \) where \( i \ne j \), constitute negative samples. By optimizing a contrastive loss, the model is encouraged to learn view- and scale-invariant visual representations, thus enhancing its ability to extract stable physical characteristics from aerial scenes under diverse imaging conditions.
}

\textcolor{black}{Specifically, the contrastive loss used in our method is defined as follows:
}

\textcolor{black}{
\begin{equation}
\mathcal{L}_{\text{contrast}} = -\log \frac{\exp(\text{sim}(\mathbf{h}_i, \mathbf{h}_i')/\tau)}{\sum\limits_{j=1,j \ne i}^{N} \exp(\text{sim}(\mathbf{h}_i, \mathbf{h}_j')/\tau)}
\end{equation}
}

\textcolor{black}{where $\mathbf{h}_i$ and $\mathbf{h}_i'$ are the normalized representations of a positive pair, $\text{sim}(\mathbf{h}_i, \mathbf{h}_j) = \frac{\mathbf{h}_i^\top \mathbf{h}_j}{\|\mathbf{h}_i\| \, \|\mathbf{h}_j\|}$ is the cosine similarity between two vectors, $\tau$ is a temperature hyper parameter, and $N $is the batch size.}

\subsection{Downstream Fine-tuning}
\label{sec:ada}

\begin{figure}
\centering
\includegraphics[width=0.85\linewidth]{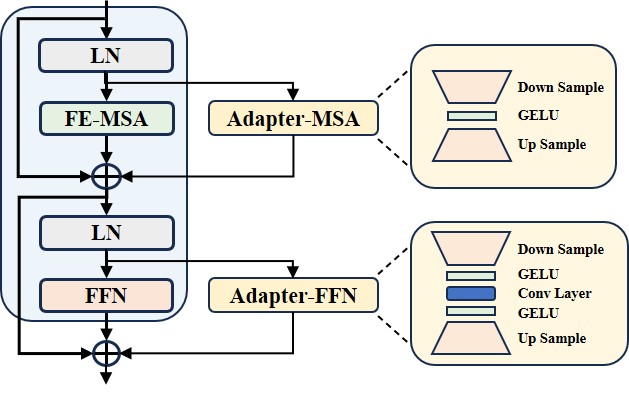}
\caption{ARS-Adapter is an enhancement module designed to improve transformer models by adding efficient, low-cost adaptation layers. It integrates down-sampling, activation (e.g., GELU), and up-sampling, optionally including convolution, to enrich feature representations while reducing computational overhead.}
\label{fig:adapter}
\end{figure}

In the methodological section of the paper, we expand upon the fine-tuning strategies employed for downstream tasks. Beyond the conventional approach of fine-tuning the entire model, this study introduces an adapter-based fine-tuning method tailored for an aviation foundation model. The proposed ARS-Adapter comprises a standard Adapter module coupled with an additional depthwise convolution module, shown as Fig.\ref{fig:adapter}. The latter is specifically designed to enhance the model's performance in tasks such as detection and tracking. Like AdaptFormer \cite{chen2022adaptformer}, the introduced ARS-Adapter module operates in parallel with the model. The Adapter module featuring the depthwise convolution is paralleled with the MLP layer, while the Adapter module without this convolution module is paralleled with the Attention layer. Therefore, the forward propagation is:

\begin{subequations}
\begin{align}
&\text{Adapter-MSA} = \text{Up}(\text{GELU}(\text{Down}(x))) \\
&\text{Adapter-FFN} = \text{Up}(\text{GELU}(\text{Conv}(\text{GELU}(\text{Down}(x)))))
\end{align}
\end{subequations}

% This dual-adapter architecture aims to provide a flexible and efficient means of transferring knowledge from the aviation foundation model to various target tasks, thereby improving the model's adaptability and performance in specialized applications.

Specifically, the downsampling layer performs 16 times downsampling in the embedding dimension, and the corresponding upsampling layer performs 16 times upsampling. In parallel with FFN, the adapter introduces an additional convolution layer, which uses 7×7 depth-separated convolution to reduce fine-tuning parameters while introducing more spatial information.

During training, the parameters of the entire pre-trained model will be frozen, and only the parameters of the ARS-Adapter mentioned above module will be trained and updated. This approach allows the pre-trained model to achieve the same effect as full fine-tuning by only fine-tuning less than 0.5$\%$ of the parameters. \textcolor{black}{Parameter Analysis are detailed in Appendix A.} The specific experimental results are detailed in Sec.\ref{sec:comp.adapter}.

\section{Experiments}

\subsection{Pre-training Implementation}
\label{sec:Pre-training Implementation}
In the pre-training phase of our model, approximately 5.4 million image patches, as mentioned earlier, are utilized. Before pre-training, we initialize the model parameters using Kaiming initialization. Subsequently, we perform 100 epochs of masked image modeling (MIM) pre-training. During MIM pre-training, the input image size is 224, and random cropping and scaling are applied to the input images. The mask size is set to 32 pixels. The batch size is 2048, with a base learning rate of 2e-4, a cosine annealing learning rate schedule, and a warm-up period of 2 epochs. \textcolor{black}{We adopt a masking ratio of 0.6 as the default configuration. This value is selected based on preliminary experiments, balancing semantic abstraction and structural preservation in ARS images characterized by multi-scale sparsity.} Following MIM pre-training, we continue with contrastive learning (CL) pre-training, jointly training for 36 epochs. The input image size remains 224, and similar random cropping and scaling are applied. The batch size is adjusted to 1024, with a base learning rate of 2e-4, a cosine annealing learning rate schedule, and a warm-up period of 1 epoch.

\begin{figure}
\centering
\includegraphics[width=1.0\linewidth]{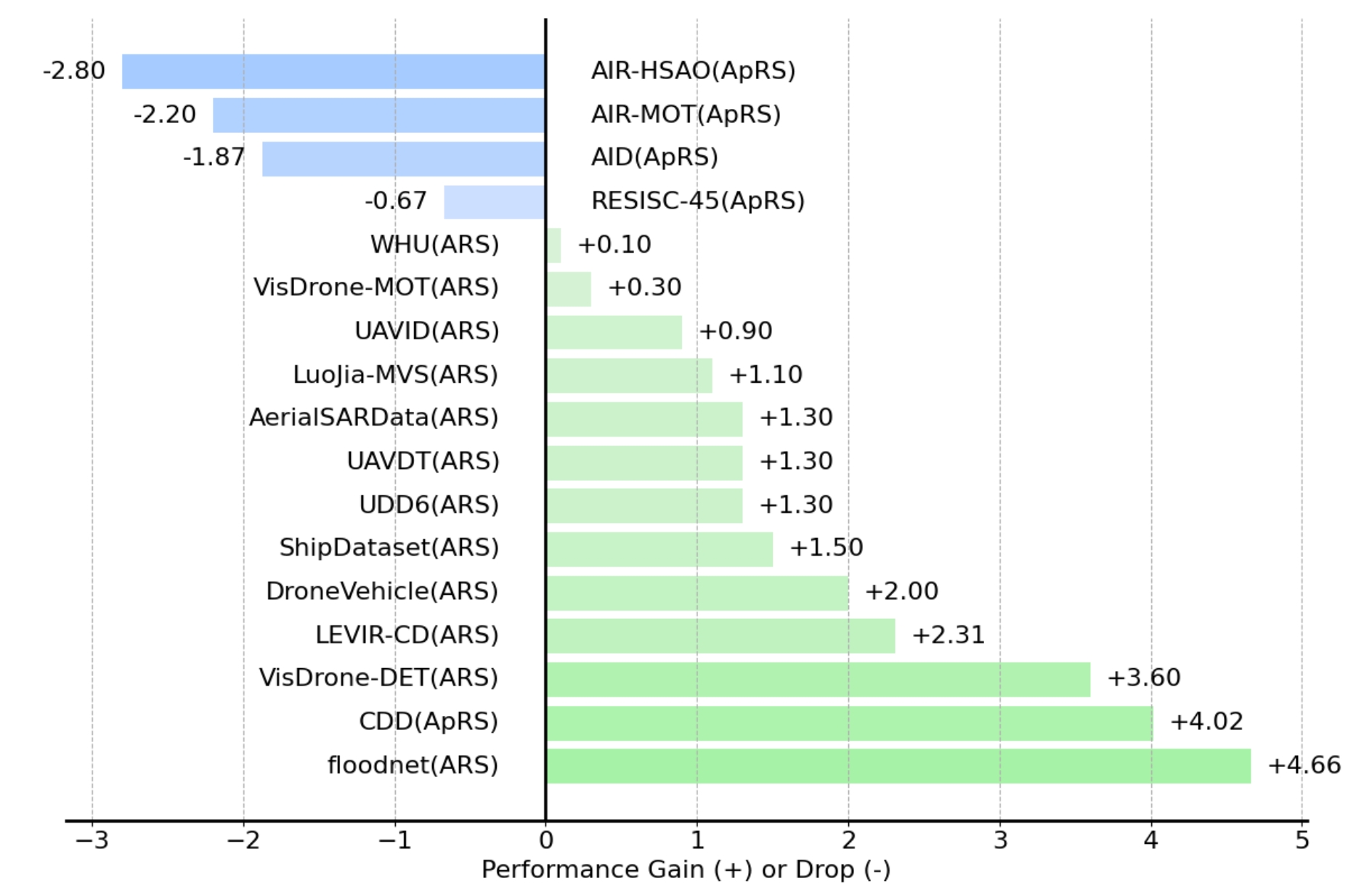}
\caption{\textcolor{black}{Performance improvement of RingMo-Aerial over previous SOTA methods across various remote sensing datasets. The horizontal axis denotes the percentage change in performance.}}
\label{fig:compare_with_sota}
\end{figure}

\subsection{Performance on Single/Multi-modal Downstream Tasks}

We validated the performance of RingMo-Aerial across six downstream tasks, covering both single-modal and multi-modal tasks, and involving a variety of data modalities and tasks commonly encountered in remote sensing scenarios. \textcolor{black}{Fig.~\ref{fig:compare_with_sota} presents a comprehensive comparison between RingMo-Aerial and previous state-of-the-art (SOTA) methods on 17 datasets. It clearly demonstrates that RingMo-Aerial achieves superior performance, particularly in ARS imagery characterized by diverse viewing angles and object scales. Notably, RingMo-Aerial yields significant gains in semantic segmentation (e.g., +4.66 mIoU on FloodNet), object detection (e.g., +3.6 mAP on VisDrone-DET), and change detection tasks (e.g., +4.02 F1 on CDD), highlighting its strong generalization and representation capabilities across complex airborne environments.}

Detailed descriptions of the datasets, experimental setups, \textcolor{black}{table of comparison with state-of-the-art methods} and additional visualizations for each task are provided in the Appendix (Semantic Segmantation C, Object Detection D, Change Detection E, Scene Classification F, 3D-Reconstruction G and Object Tracking H).

% \begin{figure}
% \centering
% \includegraphics[width=0.90\linewidth]{images/Adapter.jpg}
% \caption{ARS-Adapter is an enhancement module designed to improve transformer models by adding efficient, low-cost adaptation layers. It integrates down-sampling, activation (e.g., GELU), and up-sampling, optionally including convolution, to enrich feature representations while reducing computational overhead.}
% \label{fig:adapter}
% \end{figure}

\subsection{Ablation Experiments}

\subsubsection{Ablation Experiments on pre-training}
\begin{figure}
\centering
\includegraphics[width=0.75\linewidth]{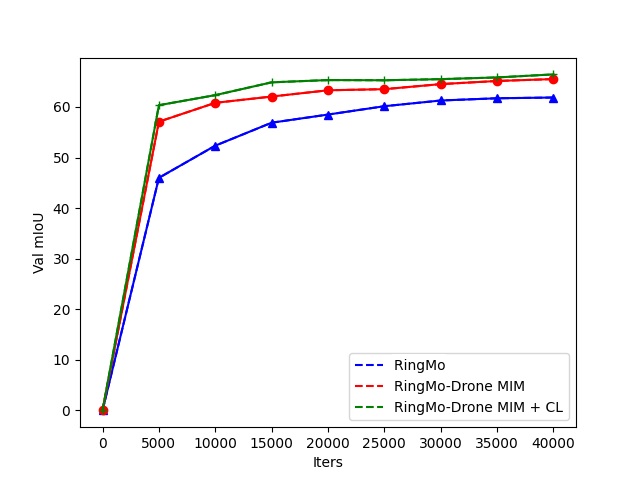}
\caption{The ablation experiment on the UAVID dataset and UperNet is used as the segmentation head. RingMo-Aerial pretrained with MIM and CL achieves the fastest convergence speed and optimal segmentation performance.}
\label{fig:abl}
\end{figure}

\begin{table}[t]
\centering
\setlength{\abovecaptionskip}{1pt}
\setlength{\belowcaptionskip}{1pt}
\caption{Ablation Experiments of pretrain methods}
\label{tab:abl}
\begin{tabular}{l|ll|ll}
\toprule
& \multicolumn{2}{c|}{UAVID} & \multicolumn{2}{c}{Visdrone-DET}                    \\

& mAcc        & mIoU        & \multicolumn{1}{c}{mAP} & \multicolumn{1}{c}{mAP$_{50}$} \\
\midrule
RingMo               & 71.6        & 61.9        & 28.4                    & 45.2                      \\
RingMo-Aerial (MIM)      & 75.7        & 65.8        & 30.5                    & 50.3                      \\
RingMo-Aerial (MIM + CL) & \textbf{76.5}        & \textbf{66.4}        & \textbf{31.1}                    & \textbf{51.8}   \\
\bottomrule 
\end{tabular}
\end{table}

\begin{table}[t]
\centering
\setlength{\abovecaptionskip}{1pt}
\setlength{\belowcaptionskip}{1pt}
\caption{Ablation Experiments with the proposed ARS-Adapter}
\label{tab:abl_ad}
\begin{tabular}{l|ll|ll}
\toprule
& \multicolumn{2}{c|}{UAVID} & \multicolumn{2}{c}{Visdrone-DET}                    \\

& mAcc        & mIoU        & \multicolumn{1}{c}{mAP} & \multicolumn{1}{c}{mAP$_{50}$} \\
\midrule
RingMo-Aerial (MIM)      & 71.6        & 62.9        & 25.3                    & 42.7                      \\
RingMo-Aerial (MIM + CL) & \textbf{74.3}        & \textbf{64.4}        & \textbf{31.0}                    & \textbf{51.4}   \\
\bottomrule 
\end{tabular}
\end{table}

We utilize UperNet model on UAVID and Cascade RCNN model on Visdrone-DET. The comparative analysis involves three foundation models: RingMo pre-trained with MIM \cite{sun2022ringmo}, RingMo-Aerial solely pre-trained with MIM, and RingMo-Aerial pre-trained with both MIM and CL. This comparative investigation aims to substantiate the efficacy and consistency of the proposed method.

The comparison is shown in Fig.\ref{fig:abl} and Tab.\ref{tab:abl}. The experimental findings reveal a progressive convergence speed and performance enhancement across the three foundation models. Notably, RingMo-Aerial trained with MIM and CL exhibits the swiftest convergence and attains the optimal performance. This observation underscores the synergistic relationship between the FE-MSA module proposed in this study and the MIM + CL pre-training strategy, affirming their mutually reinforcing nature and consistent effectiveness.

In addition, models pretrained with MIM combined with CL exhibit enhanced adaptability when used with ARS-Adapter. Consequently, ablation studies were conducted to compare ARS-Adapter between models solely pretrained with MIM and those pretrained with both MIM and CL. Comparative results on the UAVID and VisDrone-DET datasets are presented in Tab.\ref{tab:abl_ad}. The experimental findings indicate that models additionally trained with CL significantly improve performance in both detection and segmentation tasks when fine-tuned with ARS-Adapter.

\subsubsection{Performance Comparison Between ARS-Adapter and Other Fine-Tuning Methods}
\label{sec:comp.adapter}

\begin{table}[t]
\begin{threeparttable}
\centering
\setlength{\abovecaptionskip}{1pt}
\setlength{\belowcaptionskip}{1pt}
\caption{Comp. between fine-tuning methods}
\label{tab:adapter}
\begin{tabular}{llllll}
\toprule
    \multicolumn{1}{c}{\multirow{2}{*}{Methods}}        & \multicolumn{1}{c}{\multirow{2}{*}{Parameters}} & \multicolumn{2}{c}{UAVID} & \multicolumn{2}{c}{VisDrone-DET} \\
            & \multicolumn{1}{c}{}                            & mAcc        & mIoU        & mAP            & mAP$_{50}$           \\
\midrule
Froezn      & 0\%                                             & 61.2        & 49.3        & 21.5           & 36.2            \\
Full        & 100\%                                           & 76.5        & 66.4        & 31.1           & 51.8            \\
Adapter     & 0.39\%                                          & 68.9        & 57.7        & 28.9           & 49.1            \\
AdaptFormer & 0.39\%                                          & 71.3        & 62.1        & 30.3           & 50.8            \\
\midrule
ARS-Adapter \tnote{*} & 0.41\%                                          & \textbf{74.3}        & \textbf{64.4}        & \textbf{31.0}             & \textbf{51.4}         \\
\bottomrule 
\end{tabular}
\begin{tablenotes}
\footnotesize
\item[*] Although ARS-Adapter is less accurate than global fine-tuning, it significantly outperforms other fine-tuning methods when the proportion of fine-tuning parameters is almost constant.
\end{tablenotes}
\end{threeparttable}
\end{table}
% As previously discussed, this paper proposes the ARS-Adapter, a parameter-efficient fine-tuning method tailored for ARS vision. This section focuses on comparing the performance of the ARS-Adapter with other fine-tuning methods in ARS vision applications. The comparisons are conducted using the UAVID and VisDrone-DET datasets, employing UperNet and Cascade RCNN as the segmentation and detection frameworks. The evaluation aims to highlight the efficacy and advantages of the ARS-Adapter in enhancing ARS vision tasks compared to conventional fine-tuning techniques.

% The results of the comparative experiments are presented in Tab.\ref{tab:adapter}. The "Parameters" column indicates the proportion of parameters in the pre-trained model backbone that require fine-tuning. It is observed that the proportion of parameters needing adjustment with the ARS-Adapter is approximately equivalent to that required by the classical Adapter and AdaptFormer. However, the model fine-tuned using the ARS-Adapter achieves superior fine-tuning performance compared to these methods. Specifically, the performance loss relative to full fine-tuning is limited to approximately 2$\%$, demonstrating the efficacy of the ARS-Adapter in maintaining high performance with minimal parameter adjustments.

In the context of ARS, traditional model fine-tuning presents specific challenges. These include the propensity for overfitting due to the extensive pre-training on diverse tasks, which can lead to a model that is overly tailored to the nuances of the training data, thereby compromising its applicability to new, unseen data. Overfitting is particularly problematic in ARS, where models are not only expected to generalize across various tasks but also to adapt to the unique perspectives and resolutions inherent in aerial imagery. This contrasts with ApRS, which typically operates within a more constrained perspective and resolution, reducing the risk of overfitting to specific views. The ARS-Adapter, therefore, addresses these challenges by introducing a parameter-efficient fine-tuning approach that minimizes overfitting risks while maintaining high performance, as evidenced by the experimental results.

This section compares ARS-Adapter against other fine-tuning methods, using UAVID and VisDrone-DET datasets with UperNet and Cascade RCNN as the segmentation and detection frameworks, respectively. The aim is to assess the effectiveness of ARS-Adapter compared to conventional fine-tuning techniques.

The experimental results are presented in Tab.\ref{tab:adapter}. For clarity, we compare ARS-Adapter with full fine-tuning, zero fine-tuning, and two other fine-tuning methods (Adapter and AdaptFormer). Full fine-tuning is a performance upper bound, while zero fine-tuning provides a reference for minimal parameter adjustment. ARS-Adapter requires a similar proportion of parameters(0.41\%) to be adjusted as Adapter(0.39\%) and AdaptFormer(0.39\%), but achieves superior fine-tuning efficiency. Additionally, the performance loss compared to full fine-tuning(100\%) is limited to approximately 2\%, demonstrating that ARS-Adapter can achieve high performance with minimal parameter adjustments.

These results indicate that ARS-Adapter maintains high performance while significantly reducing the number of parameters requiring adjustment, demonstrating its efficiency in fine-tuning. 

\subsubsection{Ablation of Mask ratio}
\label{sec:Different ratio of Masking}
% Major Revision
\begin{table}[htbp]
\centering
\caption{Object detection performance of RingMo-Aerial under different masking ratios on the \textbf{DroneVehicle} dataset (modality: RGB+IR). The result at 0.6 is from the original configuration.}
\label{tab:masking_ratio_ablation_od}

\begin{tabular}{c|c c c c c c}
\toprule
\textbf{Mask Ratio} & \textbf{mAP} & \textbf{mAP$_{50}$} & \textbf{mAP$_{75}$} & \textbf{mAP$_s$} & \textbf{mAP$_m$} & \textbf{mAP$_l$} \\
\midrule
0.5 & 0.489 & 0.728 & 0.566 & 0.196 & 0.493 & 0.664 \\
\textbf{0.6} & \textbf{0.492} & \textbf{0.735} & \textbf{0.570} & 0.228 & \textbf{0.497} & \textbf{0.684} \\
0.7 & 0.487 & 0.726 & 0.567 & 0.195 & 0.490 & 0.668 \\
0.8 & 0.490 & 0.731 & 0.567 & \textbf{0.235} & 0.494 & 0.681 \\
\bottomrule
\end{tabular}
\end{table}

\begin{table}[htbp]
\centering
\caption{Semantic segmentation performance of RingMo-Aerial under different masking ratios on the \textbf{UAVid} dataset. The result at 0.6 is from the original configuration.}
\label{tab:masking_ratio_ablation_ss}
\begin{tabular}{ccc}
\toprule
\textbf{Mask Ratio} & \textbf{mIoU (\%)} & \textbf{mAcc (\%)} \\
\midrule
0.5 & 68.29 & 78.18 \\
\textbf{0.6} & \textbf{68.70} & \textbf{78.70} \\
0.7 & 67.72 & 77.60 \\
0.8 & 68.56 & 78.28 \\
\bottomrule
\end{tabular}
\end{table}

\textcolor{black}{To investigate the influence of masking ratio in pretraining, we conduct additional experiments under different ratios (0.5, 0.6, 0.7, 0.8). Results on both object detection (DroneVehicle, RGB+IR) and semantic segmentation (UAVid) are presented in Tab.~\ref{tab:masking_ratio_ablation_od} and Tab.~\ref{tab:masking_ratio_ablation_ss} The masking ratio of 0.6 achieves the highest mAP and mIoU, confirming its effectiveness across tasks. Interestingly, higher ratios (e.g., 0.8) slightly improve small object detection, while lower ratios (e.g., 0.5) provide marginal gains on medium-sized targets. These findings suggest that ARS imagery, due to its multi-scale structure and sparse layout, requires more carefully tuned masking strategies than natural images.}

\subsubsection{Ablation of depthwise convolution placement}
\label{sec:depthwise convolution placement}

\begin{table}[htbp]
\centering
\caption{Ablation study on the placement of depthwise convolution in the FE-MSA module. Results are reported on the \textbf{DroneVehicle} dataset with \textbf{RGB+IR} modality.}
\label{tab:dwconv_ablation}
\begin{threeparttable}
\begin{tabular}{c|cccccc}
\toprule
\textbf{Variant} & \textbf{mAP} & \textbf{mAP$_{50}$} & \textbf{mAP$_{75}$} & \textbf{mAP$_s$} & \textbf{mAP$_m$} & \textbf{mAP$_l$} \\
\midrule
Original\tnote{*} & \textbf{0.492} & \textbf{0.735} & \textbf{0.570} & \textbf{0.228}e & \textbf{0.497} & \textbf{0.684} \\
Conv1\tnote{*} & 0.485 & 0.720 & 0.561 & 0.183 & 0.485 & 0.664 \\
Conv2\tnote{*} & 0.483 & 0.719 & 0.556 & 0.193 & 0.485 & 0.670 \\
\bottomrule
\end{tabular}
\begin{tablenotes}
\footnotesize
\item[\tnote{*}] Original: Depthwise convolution is inserted after \texttt{norm1} and before the attention module (used in our final model). Conv1: Depthwise convolution is placed after the FFN block. Conv2: Depthwise convolution is placed between the multi-head attention and FFN.
\end{tablenotes}
\end{threeparttable}
\end{table}

\textcolor{black}{To analyze the effect of depthwise convolution placement in the FE-MSA module, we compare three configurations using object detection performance on the DroneVehicle dataset (RGB+IR). As shown in Tab.~\ref{tab:dwconv_ablation}, the results demonstrate that the \textbf{Original configuration achieves the best overall performance}, with 0.492 mAP and consistent advantages across all scales, especially for large objects. Although Conv2 slightly improves small object detection (mAP$_s$ = 0.193), it performs worse overall. This confirms that applying depthwise convolution early, before semantic modeling via attention and FFN, provides better spatial enhancement for ARS images. Therefore, we adopt the Original configuration in our final design.
}

% \noindent\textbf{Reviewer Comment:} \\
% \textit{Could the ablation study be augmented with an ablation of the depthwise convolution placement?}

% \vspace{1ex}
% \noindent\textbf{Response:} \\
% Thank you for the thoughtful suggestion. In response, we conduct an ablation study comparing three placements of the depthwise convolution in the FE-MSA module:
% \begin{itemize}
%     \item \textbf{Original (used in our paper)}: inserted after the first normalization and before the attention module;
%     \item \textbf{Conv1}: inserted after the FFN block;
%     \item \textbf{Conv2}: inserted between the attention module and FFN block.
% \end{itemize}

% The comparison is performed on the \textbf{DroneVehicle dataset with RGB+IR modality}, and the results are shown in Table~\ref{tab:dwconv_ablation}. The \textbf{Original} placement achieves the best mAP (0.492), mAP$_{50}$ (0.735), and mAP$_{75}$ (0.570). While Conv2 improves small object detection (mAP$_s$ = 0.193), it underperforms overall. These results validate our design choice to apply local spatial enhancement \textit{before} attention, allowing better feature modulation for ARS-specific multi-scale structures.

% This ablation is now included in the revised manuscript in \textbf{Sec.~\ref{sec:depthwise convolution placement}} and presented in \textbf{Tab.~\ref{tab:dwconv_ablation}}.

\subsubsection{Ablation of depthwise convolution kernel size}
\label{sec:Different ratio of kernel size}
% Major Revision
\begin{table}[htbp]
\centering
\caption{Object detection performance of RingMo-Aerial under different size depthwise separation convolution kernels on the \textbf{DroneVehicle} dataset (modality: RGB+IR). The results at 7 are from the original configuration.}
\label{tab:kernel_size_ablation_od}
\begin{tabular}{c|c c c c c c}
\toprule
\textbf{Kernel Size} & \textbf{mAP} & \textbf{mAP$_{50}$} & \textbf{mAP$_{75}$} & \textbf{mAP$_s$} & \textbf{mAP$_m$} & \textbf{mAP$_l$} \\
\midrule
3 & 0.485 & 0.722 & 0.564 & 0.192 & 0.489 & 0.661 \\
5 & 0.490 & 0.729 & 0.568 & 0.179 & 0.494 & 0.673 \\
\textbf{7} & \textbf{0.492} & \textbf{0.735} & \textbf{0.570} & \textbf{0.228} & \textbf{0.497} & \textbf{0.684} \\
9 & 0.490 & 0.733 & 0.566 & 0.192 & 0.493 & 0.684 \\
\bottomrule
\end{tabular}
\end{table}

\begin{table}[htbp]
\centering
\caption{Semantic segmentation performance of RingMo-Aerial under different size depthwise separation convolution kernels on the \textbf{UAVid} dataset. The result at 7 is from the original configuration.}
\label{tab:kernel_size_ablation_ss}
\begin{tabular}{ccc}
\toprule
\textbf{Kernel Size} & \textbf{mIoU (\%)} & \textbf{mAcc (\%)} \\
\midrule
3 & 67.18 & 77.77 \\
5 & 67.57 & 78.21 \\
\textbf{7} & \textbf{68.70} & \textbf{78.70} \\
9 & 67.03 & 78.15 \\
\bottomrule
\end{tabular}
\end{table}

\textcolor{black}{To investigate the impact of different sizes of depthwise separable convolution kernels during pre-training, we conducted additional experiments using kernels of sizes 3, 5, 7, and 9. The experimental results for object detection (DroneVehicle, RGB+IR) and semantic segmentation (UAVid) are presented in Table.~\ref{tab:kernel_size_ablation_od} and Table.~\ref{tab:kernel_size_ablation_ss}, respectively. The results show that a kernel size of 7 achieves the highest mAP and mIoU, demonstrating its effectiveness across different tasks. These findings suggest that kernels of sizes 3 and 5 provide insufficient receptive fields, limiting the ability to extract complete and informative features. Conversely, a kernel size of 9 results in an excessively large receptive field, introducing redundant information that interferes with the extraction of relevant features. Therefore, a kernel size of 7 strikes the optimal balance, enabling effective feature extraction and leading to superior performance.}

\section{Conclusion}
RingMo-Aerial presented in this paper represents the first large foundation model specifically tailored for Aerial Remote Sensing (ARS) vision tasks. Following an in-depth analysis of the unique aspects of ARS vision tasks, this study identified the necessity for foundation model research in this domain and proposed a series of innovative solutions based on the RingMo framework to address the specific challenges therein. Our experiments demonstrate that RingMo-Aerial achieves  SOTA performance across various datasets for ARS vision tasks. This demonstrates the utility and effectiveness of our approach in improving ARS vision tasks. In the future, RingMo-Aerial and its associated technologies are poised to provide a robust foundation model for ARS vision research. Future work will focus on improving the model's generalization capabilities, computational efficiency, and broader real-world deployment. Additionally, we will explore the extension of RingMo-Aerial to other vision tasks and scenarios, aiming to expand its application scope.

% if have a single appendix:
%\appendix[Proof of the Zonklar Equations]
% or
%\appendix  % for no appendix heading
% do not use \section anymore after \appendix, only \section*
% is possibly needed

% use appendices with more than one appendix
% then use \section to start each appendix
% you must declare a \section before using any
% \subsection or using \label (\appendices by itself
% starts a section numbered zero.)
%

% % you can choose not to have a title for an appendix
% % if you want by leaving the argument blank
% \section{}
% Appendix two text goes here.

% use section* for acknowledgment
\ifCLASSOPTIONcompsoc
% The Computer Society usually uses the plural form
\section*{Acknowledgments}
\else
% regular IEEE prefers the singular form
\section*{Acknowledgment}
\fi

This work was supported by the National Nature Science Foundation of China under Grant 62331027 and 62301538. 

\clearpage

% \bibliographystyle{ieeetr}
% \bibliography{refs}

% Can use something like this to put references on a page
% by themselves when using endfloat and the captionsoff option.
\ifCLASSOPTIONcaptionsoff
\newpage
\fi

\vspace{-3.5em}

\appendices

\section{Frequency-Enhanced Multi-Head Self-Attention}
\label{appendices: FE-MSA}
\textcolor{black}{
The following content provides a derivation analysis of the FE-MSA module from both the spatial domain and frequency domain perspectives.}

\subsection{Spatial-Domain Analysis}

\textcolor{black}{Self-attention performs “global” modeling by assigning correlation weights at the patch level. For small targets, fine-scale features remain under-integrated. Inserting a $7\times7$ convolution before attention effectively enlarges the receptive field at each position from $1\times1$ (a single point within a patch) to $7\times7$, while a residual connection preserves the original features:}
\textcolor{black}{
\begin{equation}
Y_{\mathrm{conv}} = X + \mathrm{DWConv}_{7\times7}(X).
\end{equation}
For each spatial location $i$,
\begin{equation}
y_i = x_i + \bigl[\mathrm{DWConv}_{7\times7}(X)\bigr]_i,
\end{equation}
}
\textcolor{black}{where $\bigl[\mathrm{DWConv}_{7\times7}(X)\bigr]_i$ aggregates high-frequency (edge, texture) information from a $7\times7$ neighborhood around $i$, thereby enhancing local edges and structural details of small targets. The convolutional branch thus accentuates small-target features in the local context, enabling subsequent attention to focus more precisely.}

\textcolor{black}{Consider the standard self-attention formulation for head $h$:}

% \begin{equation}
% A^{(h)} = \mathrm{Softmax}\Bigl(Q^{(h)} {K^{(h)}}^\top / \sqrt{d}\Bigr),\\
% Q^{(h)} = X W_Q^{(h)},\\
% K^{(h)} = X W_K^{(h)}.
% \end{equation}
\textcolor{black}{
% \begin{equation}
\begin{align}
&A^{(h)} = \mathrm{Softmax}\Bigl(Q^{(h)} {K^{(h)}}^\top / \sqrt{d}\Bigr), \\
&Q^{(h)} = X W_Q^{(h)}, \\
&K^{(h)} = X W_K^{(h)}.
\end{align}
% \end{equation}
}
\textcolor{black}{
After adding the convolutional branch,
\begin{align}
Q'^{(h)} &= Y W_Q^{(h)} = X W_Q^{(h)} + \mathrm{DWConv}(X) W_Q^{(h)},\\
K'^{(h)} &= Y W_K^{(h)} = X W_K^{(h)} + \mathrm{DWConv}(X) W_K^{(h)}.
\end{align}}
\textcolor{black}{For any query–key pair $(i,j)$, the modified dot-product logit expands to}
% \begin{equation}
% \ell'_{ij} = Q'_i \cdot K'_j = \underbrace{x_i W_Q W_K^\top x_j}_{\ell_{ij}} + \underbrace{x_i W_Q W_K^\top [\mathrm{Conv}(X)]_j\\
% + [\mathrm{Conv}(X)]_i W_Q W_K^\top x_j}_{\Delta_{ij}}\\ 
% + \underbrace{[\mathrm{Conv}(X)]_i W_Q W_K^\top [\mathrm{Conv}(X)]_j}_{\delta_{ij}}.
% \end{equation}
\textcolor{black}{
\begin{align}
\ell'_{ij} &= Q'_i \cdot K'_j \nonumber \\
&= \underbrace{x_i W_Q W_K^\top x_j}_{\ell_{ij}} \nonumber\\
&\quad + \underbrace{x_i W_Q W_K^\top [\mathrm{Conv}(X)]_j 
+ [\mathrm{Conv}(X)]_i W_Q W_K^\top x_j}_{\Delta_{ij}} \nonumber\\
&\quad + \underbrace{[\mathrm{Conv}(X)]_i W_Q W_K^\top [\mathrm{Conv}(X)]_j}_{\delta_{ij}}.
\end{align}
}
\textcolor{black}{Here, the original term $\ell_{ij}$ captures pure attention on the raw features, while the enhancement term $\Delta_{ij}$ and the double-enhancement term $\delta_{ij}$ arise from the convolutional branch. Since $[\mathrm{Conv}(X)]$ is amplified at high-frequency regions corresponding to small-target edges and textures, these additional terms preferentially boost logits for positions inside or related to small targets. Consequently,}
\textcolor{black}{
\begin{equation}
\ell'_{ij} > \ell_{ij}
\end{equation}
}
\textcolor{black}{
for small-target regions, and the softmax scaling further amplifies attention weights at these locations, resulting in more comprehensive aggregation of small-target information.
}

\subsection{Frequency-Domain Analysis}
\textcolor{black}{
Denote the 2D Fourier transform of $X \in \mathbb{R}^{H \times W \times C}$ as
\begin{equation}
\widehat{X}(\omega_x, \omega_y) = \sum_{m=0}^{H-1} \sum_{n=0}^{W-1} X[m,n] e^{-j(\omega_x m + \omega_y n)}.
\end{equation}
}
\textcolor{black}{Small targets concentrate energy at higher spatial frequencies; letting}

% \begin{equation}
% E(\omega) = \|\widehat{X}(\omega)\|^2,
% \quad
% E_{\mathrm{high}} = \int_{\|\omega\| \ge \omega_0} E(\omega) \, d\omega
% > E_{\mathrm{low}} = \int_{\|\omega\| < \omega_0} E(\omega) \, d\omega.
% \end{equation}
\textcolor{black}{
\begin{align}
E(\omega) &= \|\widehat{X}(\omega)\|^2, \\
E_{\mathrm{high}} &= \int_{\|\omega\| \ge \omega_0} E(\omega) \, d\omega \notag\\
&> E_{\mathrm{low}} = \int_{\|\omega\| < \omega_0} E(\omega) \, d\omega.
\end{align}
}
\textcolor{black}{
Patch-level self-attention acts as a low-pass filter, responding strongly to low frequencies while attenuating high frequencies. A rough model is}

% \begin{equation}
% H_{\mathrm{attn}}(\omega) = e^{-\alpha \|\omega\|},\; \alpha > 0,
% \quad\text{or}
% \quad
% H_{\mathrm{att}}(\omega) =
% \begin{cases}
% 1,& \|\omega\| < \omega_0,\\
% \beta,& \|\omega\| \ge \omega_0,\; 0 \le \beta < 1.
% \end{cases}
% \end{equation}
\textcolor{black}{
\begin{equation}
% \begin{split}
H_{\mathrm{attn}}(\omega) = e^{-\alpha \|\omega\|}, \quad \alpha > 0, 
\end{equation}
}

\textcolor{black}{or}

\textcolor{black}{
\begin{equation}
H_{\mathrm{att}}(\omega) =
\begin{cases}
1, & \|\omega\| < \omega_0, \\
\beta, & \|\omega\| \ge \omega_0,\quad 0 \le \beta < 1.
\end{cases}
% \end{split}
\end{equation}
}
\textcolor{black}{
In contrast, a depth-wise separable $7\times7$ convolution employs a spatial filter $h[m,n]$ whose frequency response}
\textcolor{black}{
\begin{equation}
H_{\mathrm{conv}}(\omega_x, \omega_y) = \sum_{m=-3}^{3} \sum_{n=-3}^{3} h[m,n] e^{-j(\omega_x m + \omega_y n)}
\end{equation}
}
\textcolor{black}{
acts as a band-pass filter with relatively high gain in the high-frequency range. Therefore, by preceding attention with this convolution, FE-MSA compensates for the attention filter’s high-frequency attenuation:
\begin{equation}
\bigl|H_{\mathrm{conv}}(\omega)\bigr|_{\|\omega\| \ge \omega_0} > \bigl|H_{\mathrm{attn}}(\omega)\bigr|_{\|\omega\| \ge \omega_0}.
\end{equation}
This frequency-domain enhancement better preserves small-target detail during global aggregation.}

% \begin{figure}[ht]
%   \centering
%   \includegraphics[width=0.7\textwidth]{spatial_distribution.png}
%   \caption{Spatial distribution of small targets.}
%   \label{fig:spatial}
% \end{figure}

% \begin{figure}[ht]
%   \centering
%   \includegraphics[width=0.7\textwidth]{frequency_spectrum.png}
%   \caption{Energy spectrum of small targets in the frequency domain.}
%   \label{fig:frequency}
% \end{figure}

\section{Parameter analysis of ARS-Adapter}
\label{appendices: parameter analysis}

\textcolor{black}{
Let $C$ be the number of channels (i.e., embedding dimension) of the encoder layer. The Transformer encoder layer includes a depthwise separable convolution layer with $C$ channels, and four linear layers with C channels. The Feed-Forward Network (FFN) layer includes a linear layer with input $C$ and output $4C$, and another linear layer with input $4C$ and output $C$. The total number of parameters in a single encoder layer is then:}

\textcolor{black}{
\begin{equation}\label{eq:4}
   \mathrm{P_{e}}=C\times 7 \times7+4\times C\times C+ 2 \times C\times 4C = 49C + 12C^2
\end{equation}}

\textcolor{black}{Besides, the total parameters added by the ARS-Adapter are calculated as follows:}

\textcolor{black}{For each encoder layer, the added components include: Downsampling linear layers $2\times(C\times C/64)$, Upsampling linear layers $2\times(C\times C/64)$ and Depthwise separable convolution layer $\frac{C}{64} \times 3 \times 3$. Then, the total additional parameters for each encoder layer are:}

\textcolor{black}{\begin{equation}\label{eq:5}
   \mathrm{P_{a}}=4\times(C\times C/64)+\frac{C}{64} \times 3 \times 3 = \frac{9}{64}C + \frac{1}{16} C^2
\end{equation}}

\textcolor{black}{Finally, if $C = 512$, we can calculate the proportion of added parameters compared to the total parameters of the model:}

\textcolor{black}{\begin{equation}\label{eq:6}
   \mathrm{Proportion}=\frac{\mathrm{P_{a}}}{\mathrm{P_{e}} + \mathrm{P_{a}}} \approx 0.5\%
\end{equation}}

\section{Semantic Segmentation}
\label{appendices: semantic segmentation}

\begin{table*}[t]
\centering
\setlength{\abovecaptionskip}{1pt}
\setlength{\belowcaptionskip}{1pt}
\caption{Semantic Segmentation: Comp. on UAVID}
\label{tab:uavid}
\resizebox{0.99\linewidth}{!}{
\begin{threeparttable}
\begin{tabular}{lllllllllllllllllll}
\toprule
\multicolumn{1}{c}{\multirow{2}{*}{Methods}} & \multicolumn{2}{c}{Building}                      & \multicolumn{2}{c}{Tree}                          & \multicolumn{2}{c}{Clutter}                       & \multicolumn{2}{c}{Road}                          & \multicolumn{2}{c}{Low vegetation}                & \multicolumn{2}{c}{Static car}                    & \multicolumn{2}{c}{Moving car}                    & \multicolumn{2}{c}{Human}                         & \multicolumn{1}{c}{\multirow{2}{*}{mAcc}} & \multicolumn{1}{c}{\multirow{2}{*}{mIoU}} \\
\multicolumn{1}{c}{}                         & \multicolumn{1}{c}{Acc} & \multicolumn{1}{c}{IoU} & \multicolumn{1}{c}{Acc} & \multicolumn{1}{c}{IoU} & \multicolumn{1}{c}{Acc} & \multicolumn{1}{c}{IoU} & \multicolumn{1}{c}{Acc} & \multicolumn{1}{c}{IoU} & \multicolumn{1}{c}{Acc} & \multicolumn{1}{c}{IoU} & \multicolumn{1}{c}{Acc} & \multicolumn{1}{c}{IoU} & \multicolumn{1}{c}{Acc} & \multicolumn{1}{c}{IoU} & \multicolumn{1}{c}{Acc} & \multicolumn{1}{c}{IoU} & \multicolumn{1}{c}{}                      & \multicolumn{1}{c}{}                      \\
\midrule
DeeplabV3+ \cite{chen2018encoder}                                  & 82.5                    & 80.6                    & 78.1                    & 73.9                    & 71.8                    & 45.3                    & 77.9                    & 65.2                    & 58.3                    & 45.3                    & 45.8                    & 24.2                    & 70.2                    & 53.3                    & 8.2                     & 1.8                     & 61.6                                      & 48.7                                      \\
DANet \cite{fu2019dual}                                       & 81.8                    & 81.2                    & 80.2                    & 72.3                    & 69.3                    & 46.1                    & 78.2                    & 63.6                    & 56.9                    & 46.2                    & 48.3                    & 23.8                    & 70.8                    & 54.2                    & 8.6                     & 3.0                     & 61.8                                      & 48.8                                      \\
ACNet \cite{ding2019acnet}                                       & 82.3                    & 81.6                    & 80.8                    & 72.6                    & 68.9                    & 45.3                    & 79.4                    & 64.1                    & 57.3                    & 46.5                    & 48.1                    & 23.2                    & 71.2                    & 55.0                    & 8.8                     & 3.2                     & 62.1                                      & 48.9                                      \\
OCRNet \cite{yuan2020object}                                      & 81.9                    & 81.4                    & 80.6                    & 72.4                    & 69.2                    & 46.0                    & 78.6                    & 63.7                    & 57.0                    & 46.4                    & 48.5                    & 23.7                    & 71.0                    & 54.8                    & 8.5                     & 2.9                     & 61.9                                      & 48.9                                      \\
SETR \cite{zheng2021rethinking}                                        & 82.3                    & 78.9                    & 78.3                    & 71.2                    & 66.9                    & 45.8                    & 76.2                    & 63.8                    & 60.6                    & 43.7                    & 45.6                    & 22.6                    & 70.1                    & 54.8                    & 7.8                     & 1.6                     & 61.0                                      & 47.8                                      \\
Segformer \cite{xie2021segformer}                                   & 83.6                    & 81.2                    & 80.3                    & 74.1                    & 70.3                    & 46.3                    & 76.8                    & 64.6                    & 61.0                    & 46.2                    & 45.3                    & 24.6                    & 70.3                    & 56.1                    & 12.2                    & 5.3                     & 62.5                                      & 49.8                                      \\
CSwin \cite{dong2022cswin}                                       & 86.3                    & 82.8                    & 80.6                    & 75.6                    & 68.9                    & 49.2                    & 79.2                    & 66.3                    & 61.2                    & 47.8                    & 48.3                    & 26.9                    & 70.6                    & 55.0                    & 16.5                    & 10.8                    & 64.0                                      & 51.8                                      \\

UAVFormer \cite{yi2023uavformer}                                   & 88.5                    & 81.5                    & 82.3                    & 76.2                    & 67.5                    & 48.8                    & 78.6                    & 67.1                    & 63.8                    & 48.5                    & 49.6                    & 28.8                    & 71.2                    & 62.2                    & 19.3                    & 12.5                    & 65.1                                      & 53.2                                      \\
BiSeNetV2 \cite{yu2021bisenet}                                      & -                    & 81.6                    & -                    & 76.0                    & -                    & 61.2                    & -                    & 77.1                    & -                    & 61.3                    &-                    & 38.5                    & -                    & 66.4                    & -                    & 15.4                    & -                                      & 59.7                                      \\
Segmenter \cite{strudel2021segmenter}                           & -                    & 84.4                    & -                    & 76.1                    & -                   & 64.2                    & -                    & 79.8                    & -                    &57.6                    & -                    & 34.5                    & -                    & 59.2                    & -                    & 14.2                    & -                                      & 58.7                                      \\
CoaT \cite{xu2021co}                           & -                    & 88.5                    & -                    & 79.3                    & -                   & 69.0                    & -                    & 80.0                    & -                    &62.0                    & -                    & 59.1                    & -                    & 70.0                    & -                    & 18.9                    & -                                      & 65.8                                      \\
BANet \cite{chen2020banet}                           & -                    & 85.4                    & -                    & 78.9                    & -                   & 66.6                    & -                    & 80.7                    & -                    &62.1                    & -                    & 52.8                    & -                    & 69.3                    & -                    & 21.0                    & -                                      & 64.6                                      \\
UNetFormer \cite{wang2022unetformer}                           & -                    & \textbf{87.4}                    & -                    & \textbf{80.2}                    & -                   & \textbf{68.4}                    & -                    & 81.5                    & -                    &\textbf{63.5}                    & -                    & 56.4                    & -                    & 73.6                    & -                    & 31.0                    & -                                      & 67.8                                      \\
RingMo(U)\tnote{*}\cite{sun2022ringmo}                           & 90.9                    & 85.4                    & 82.6                    & 74.3                    & 73.1                    & 56.8                    & 81.6                    & 73.7                    & 71.9                    & 58.2                    & 69.9                    & 58.3                    & 75.4                    & 67.5                    & 27.4                    & 21.0                    & 71.6                                      & 61.9                                      \\
\textcolor{black}{RingMo(M)\tnote{*}\cite{sun2022ringmo}} & 
\textcolor{black}{93.0} & \textcolor{black}{83.6} & \textcolor{black}{87.9} & 
\textcolor{black}{77.4} & \textcolor{black}{74.9} & \textcolor{black}{62.4} & 
\textcolor{black}{86.4} & \textcolor{black}{78.7} & \textcolor{black}{78.5} & 
\textcolor{black}{61.5} & \textcolor{black}{62.3} & \textcolor{black}{53.0} & 
\textcolor{black}{86.4} & \textcolor{black}{\textbf{78.7}} & \textcolor{black}{36.5} & 
\textcolor{black}{28.2} & \textcolor{black}{75.7} & \textcolor{black}{64.9} \\
\midrule
RingMo-Aerial(U)\tnote{*}                            & 92.6                    & 83.8                    & 88.5                    & 77.2                    & 78.4                    & 65.5                    & 87.5                    & 80.2                    & 73.3                    & 60.7                    & \textbf{72.5}                    & 65.3                    & 84.4                    & 71.6                    & 34.8                    & 26.9                    & 76.5                                      & 66.4                                      \\
RingMo-Aerial(A)\tnote{*}                            & 92.4                    & 83.5                    & 89.0                    & 77.7                    & 77.2                    & 63.5                    & 85.8                    & 79.4                    & 75.2                    & 60.4                    & 60.5                    & 53.4                    & 85.7                    & 72.7                    & 28.5                    & 24.7                    & 74.3                                      & 64.4                                      \\
RingMo-Aerial(M)\tnote{*}                    & \textbf{93.2}                    & 85.7                    & \textbf{89.8}                   & \textbf{79.0}                    & 79.0                    & 66.3                    & \textbf{87.8}                    & \textbf{82.0}                    & \textbf{76.8}                    & 62.0                    & \textbf{72.5}                    & \textbf{65.5}                    & \textbf{88.9}                    & \textbf{76.6}                    & \textbf{41.3}                    & \textbf{32.7}                    & 78.7                                      & \textbf{68.7}    \\
\bottomrule
\end{tabular}
\begin{tablenotes}
\footnotesize
\item[*] Swin-Base (U) uses the Swin-Base backbone plus the UperNet segmentation algorithm. RingMo-Aerial (U), (A), (M) represent the RingMo aerial backbone with UperNet segmentation algorithm, ARS-Adapter and UperNet algorithm, and Mask2Former algorithm. Same below.
\end{tablenotes}
\end{threeparttable}}
\end{table*}

\begin{table*}[t]
\centering
\setlength{\abovecaptionskip}{1pt}
\setlength{\belowcaptionskip}{1pt}
\caption{Semantic Segmentation: Comp. on UDD6}
\label{tab:udd6}
\begin{tabular}{lllllllllllllll}
\toprule
\multicolumn{1}{c}{\multirow{2}{*}{Methods}} & \multicolumn{2}{c}{Other}                         & \multicolumn{2}{c}{Facade}                        & \multicolumn{2}{c}{Road}                          & \multicolumn{2}{c}{Vegetation}                    & \multicolumn{2}{c}{Vehicle}                       & \multicolumn{2}{c}{Roof}                          & \multicolumn{1}{c}{\multirow{2}{*}{mAcc}} & \multicolumn{1}{c}{\multirow{2}{*}{mIoU}} \\
\multicolumn{1}{c}{}                         & \multicolumn{1}{c}{Acc} & \multicolumn{1}{c}{IoU} & \multicolumn{1}{c}{Acc} & \multicolumn{1}{c}{IoU} & \multicolumn{1}{c}{Acc} & \multicolumn{1}{c}{IoU} & \multicolumn{1}{c}{Acc} & \multicolumn{1}{c}{IoU} & \multicolumn{1}{c}{Acc} & \multicolumn{1}{c}{IoU} & \multicolumn{1}{c}{Acc} & \multicolumn{1}{c}{IoU} & \multicolumn{1}{c}{}                      & \multicolumn{1}{c}{}                      \\
\midrule
DeeplabV3+ \cite{chen2018encoder}                                  & 92.6                    & 81.8                    & 88.6                    & 72.3                    & 85.3                    & 71.2                    & 83.6                    & 70.8                    & 78.9                    & 62.9                    & 92.4                    & 80.2                    & 86.9                                      & 73.2                                      \\
DANet \cite{fu2019dual}                                       & 92.8                    & 82.1                    & 89.3                    & 72.6                    & 84.5                    & 71.3                    & 81.8                    & 71.6                    & 82.6                    & 63.1                    & 93.8                    & 81.3                    & 87.5                                      & 73.7                                      \\
ACNet \cite{ding2019acnet}                                       & 93.1                    & 82.8                    & 90.1                    & 73.2                    & 84.8                    & 71.5                    & 81.6                    & 71.2                    & \textbf{83.2}           & 64.2                    & 94.2                    & 81.8                    & 87.8                                      & 74.1                                      \\
OCRNet \cite{yuan2020object}                                      & 92.7                    & 81.9                    & 90.2                    & 73.4                    & 84.3                    & 71                      & 81.9                    & 71.8                    & 83                      & 63.6                    & 94                      & 81.6                    & 87.7                                      & 73.9                                      \\
SETR \cite{zheng2021rethinking}                                        & 91.8                    & 79.6                    & 88.3                    & 70.9                    & 82.6                    & 69.8                    & 81.6                    & 69.6                    & 77.8                    & 61.8                    & 90.8                    & 79.9                    & 85.5                                      & 71.9                                      \\
Segformer \cite{xie2021segformer}                                   & 93                      & 82.5                    & 88.9                    & 73.1                    & 86.2                    & 72.3                    & 89.3                    & 71.8                    & 80.1                    & 68.6                    & 91.6                    & 80.8                    & 88.1                                      & 74.9                                      \\
CSwin \cite{dong2022cswin}                                       & 94.2                    & 84.8                    & 89.8                    & 74.9                    & 88.5                    & 73.8                    & 86.3                    & 73.6                    & \textbf{83.2}           & 64.9                    & 95.6                    & 86.9                    & 89.6                                      & 76.5                                      \\
UAVFormer \cite{yi2023uavformer}                                   & \textbf{95.2}           & \textbf{86.9}           & \textbf{90.3}           & \textbf{78.3}           & \textbf{89.6}           & 75.6                    & 88.1                    & 74.8                    & 82.3                    & 65.6                    & \textbf{96.3}           & 84.6                    & \textbf{90.3}                             & 77.6                                      \\
RingMo(U) \cite{sun2022ringmo}                           & 91                      & 80.4                    & 89.4                    & 74.6                    & 85.5                    & 73.2                    & 89.6                    & 78.4                    & 77.8                    & 62.3                    & 91.1                    & 82.9                    & 87.4                                      & 75.3                                      \\
\textcolor{black}{RingMo(M)\cite{sun2022ringmo}} & 
\textcolor{black}{72.18} & \textcolor{black}{59.86} & 
\textcolor{black}{84.77} & \textcolor{black}{71.23} & 
\textcolor{black}{84.62} & \textcolor{black}{71.17} & 
\textcolor{black}{\textbf{95.14}} & \textcolor{black}{\textbf{90.22}} & 
\textcolor{black}{79.66} & \textcolor{black}{67.73} & 
\textcolor{black}{93.53} & \textcolor{black}{87.52} & 
\textcolor{black}{84.98} & \textcolor{black}{74.62} \\

\midrule
RingMo-Aerial(U)                          & 90.4                    & 81.1                    & 89.7                    & 77.1                    & 88.7                    & \textbf{77.4}           & 90.4                    & 80.2                    & 81.3                    & 68.9                    & 95.9                    & 86.3                    & 89.4                                      & 78.5                                      \\
RingMo-Aerial(M)                    & 90.7                    & 80.9                    & 90.2                    & 77.4                    & 88.4                    & 77.1                    & \textbf{90.5}           & \textbf{80.8}           & 82.1                    & \textbf{69.4}           & \textbf{96.3}           & \textbf{87.8}           & 89.7                                      & \textbf{78.9}                             \\
\bottomrule 
\end{tabular}
\end{table*}

\begin{table*}[htbp]
\centering
\setlength{\abovecaptionskip}{1pt}
\setlength{\belowcaptionskip}{1pt}
\caption{Semantic Segmentation: Comp. on Floodnet}
\label{tab:floodnet}
\begin{tabular}{ccccccccccccc}
\toprule
\multicolumn{1}{c}{\multirow{3}{*}{Methods}} & \multicolumn{2}{c}{Building} & \multicolumn{2}{c}{Road} & \multicolumn{1}{c}{\multirow{2}{*}{Water}} & \multicolumn{1}{c}{\multirow{2}{*}{Tree}} & \multicolumn{1}{c}{\multirow{2}{*}{Vehicle}} & \multicolumn{1}{c}{\multirow{2}{*}{Pool}} & \multicolumn{1}{c}{\multirow{2}{*}{Grass}} & \multicolumn{1}{c}{\multirow{3}{*}{OA}} & \multicolumn{1}{c}{\multirow{3}{*}{mIoU}} \\
\multicolumn{1}{c}{} & Flooded & Non-Flooded & Flooded & Non-Flooded & & & & & & & \\

& IoU & IoU & IoU & IoU & IoU & IoU & IoU & IoU & IoU & & \\
\midrule
SegNet \cite{badrinarayanan2017segnet} & 44.72 & 69.85 & 39.70 & 76.44 & 71.17 & 76.88 & 32.74 & 35.87 & 85.84 & 87.98 & 59.25 \\
FCN8s \cite{long2015fully} & 48.93 & 73.84 & 43.11 & 80.22 & 71.71 & 77.33 & 38.19 & 45.25 & 86.85 & 88.78 & 62.83 \\
FCN16s \cite{long2015fully} & 44.97 & 67.54 & 34.14 & 72.80 & 65.39 & 76.07 & 34.28 & 32.27 & 85.66 & 87.23 & 57.01 \\
PSPNet \cite{zhao2017pyramid} & 45.24 & 70.30 & 43.05 & 76.06 & 74.32 & 74.59 & 24.41 & 40.74 & 85.88 & 88.01 & 59.40 \\
DeepLabV3 \cite{chen2017deeplab} & 50.32 & 78.02 & 48.46 & 82.06 & \textbf{76.94} & 79.90 & 47.14 & 48.45 & 88.10 & 90.13 & 66.60 \\
DeepLabV3+ \cite{chen2018encoder} & 47.62 & 74.46 & 43.12 & 79.07 & 68.35 & 76.36 & 35.47 & 36.69 & 85.60 & 87.79 & 60.75 \\
UNet \cite{ronneberger2015u} & 49.95 & 77.58 & 43.90 & 81.90 & 73.46 & 78.88 & 43.30 & 49.30 & 87.54 & 89.56 & 65.09 \\
HRNet \cite{sun2019deep} & 50.04 & \textbf{79.23} & 48.62 & 81.59 & 75.05 & 80.16 & 49.80 & 50.89 & \textbf{88.38} & 90.17 & 67.08 \\
TransUNet \cite{chen2021transunet} & 48.97 & 73.67 & 43.11 & 79.09 & 72.86 & 77.75 & 26.27 & 35.02 & 86.47 & 88.82 & 60.36 \\
RingMo(U) \cite{sun2022ringmo} & 71.95 & 66.80 & 44.22 & 74.24 & 61.97 & 77.97 & 23.11 & 33.01 & 85.22 & 88.46 & 59.83 \\
\textcolor{black}{RingMo(M)\tnote{*}\cite{sun2022ringmo}} & 
\textcolor{black}{50.80} & \textcolor{black}{70.23} & 
\textcolor{black}{39.15} & \textcolor{black}{78.56} & 
\textcolor{black}{68.10} & \textcolor{black}{74.10} & 
\textcolor{black}{47.14} & \textcolor{black}{45.13} & 
\textcolor{black}{84.94} & 
\textcolor{black}{87.50} & \textcolor{black}{62.02} \\
\midrule
RingMo-Aerial(U) & 74.48 & 77.06 & \textbf{55.78} & \textbf{83.00} & 70.35 & 81.41 & 53.96 & 58.08 & 87.04 & 90.40 & 71.24 \\
RingMo-Aerial(M) & \textbf{76.17} & 77.90 & 48.57 & 82.62 & 62.75 & \textbf{81.70} & \textbf{63.16} & \textbf{65.79} & 87.16 & \textbf{90.43} & \textbf{71.76} \\
\bottomrule
\end{tabular}
\end{table*}

\begin{figure*}
\centering
\includegraphics[width=1.0\linewidth]{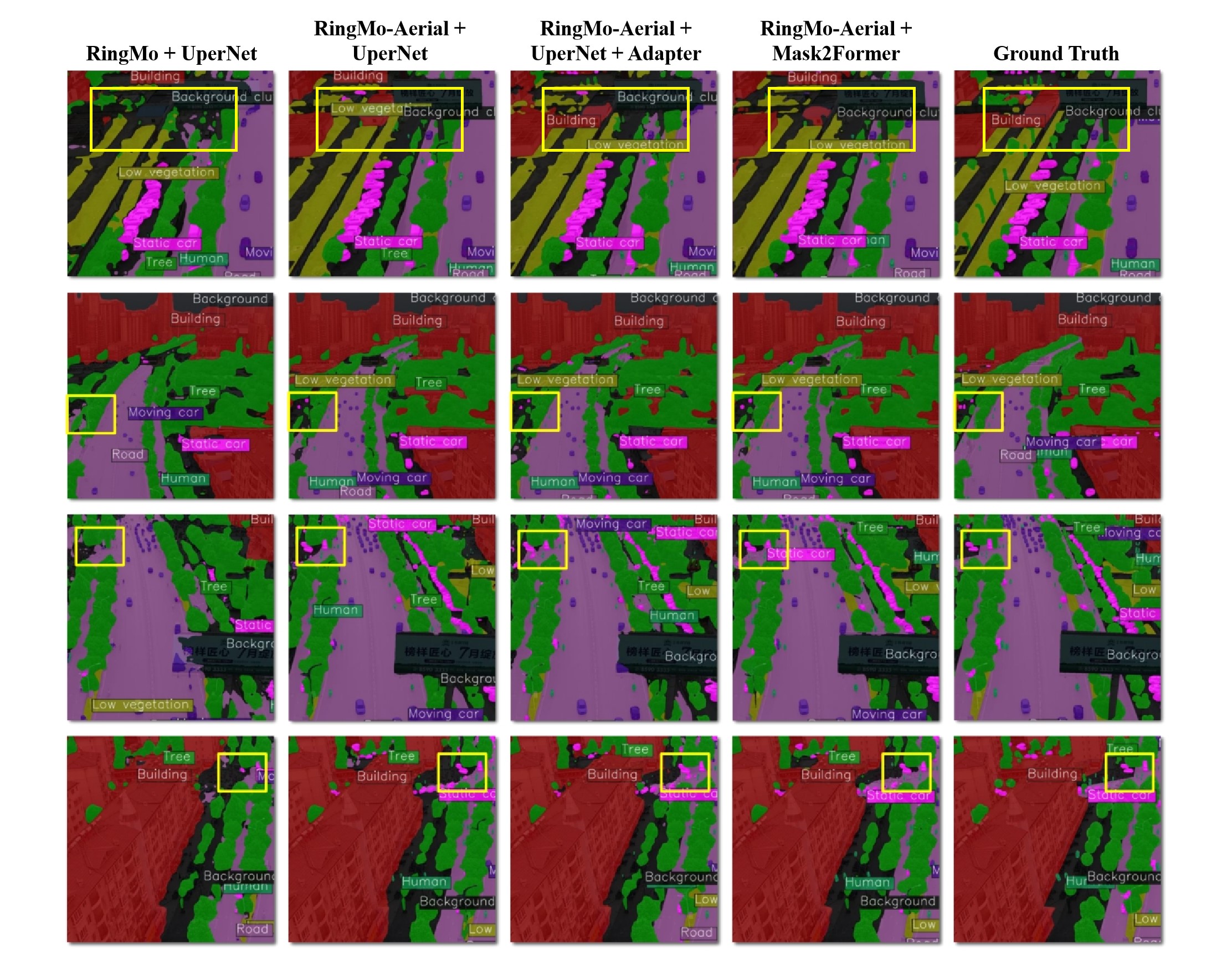}
\caption{Qualitative comparison of semantic segmentation results on the UAVID dataset using different methods. The figure illustrates differences in capturing details such as vegetation, moving vehicles, and buildings, with highlighted regions emphasizing variations in detail and performance.}
\label{fig:seg-vis}
\end{figure*}

\subsection{Datasets and Evaluation Metrics}

For semantic segmentation, the UAVID \cite{lyu2020uavid}, UDD6 \cite{chen2018large} and FloodNet\cite{rahnemoonfar2021floodnet} datasets are employed.

\begin{itemize}
\item \textbf{UAVID.} The UAVID dataset has 42 sequences of 4K resolution in total (20 train, 7 valid, and 15 test), and each sequence contains ten pictures. In the experiment, all images were cropped into 1024×1024 pixel slices with 256 pixels of overlap. A total of 27 sequences in the training and validation sets are used for training, and 15 sequences in the test set are used for testing.

\item \textbf{UDD6.} The training set of the UDD6 dataset includes 106 images, and the validation set includes 35 images. All of the images are 4K resolution. In the experiment, all images were cropped into 1024×1024 pixel slices with 256 pixels of overlap. The training set is used for training, and the validation set is used for testing.

\item \textbf{FloodNet.} The FloodNet dataset contains 500 high-resolution remote sensing images with a resolution of 4K. The dataset is divided into training, validation, and test sets, with the training set containing 300 images, the validation set containing 100 images, and the test set containing 100 images. Images are annotated as flood and non-flood areas, supporting multi-class flood impact assessment. All images are cropped into 1024×1024 pixel slices with a 256-pixel overlap in the experiment, facilitating the training and evaluation of machine learning models.
\end{itemize}

In our experiment of semantic segmentation, the evaluation metrics are mean intersection over union (\textbf{mIoU}), mean accuracy (\textbf{mAcc}), and average accuracy (\textbf{aAcc}).

\subsection{Experiment Settings}

The experiment of the semantic segmentation is based on segmentation \cite{mmseg2020}, UperNet \cite{xiao2018unified}, and Mask2Former \cite{cheng2022masked} are employed as the segmentation framework. The training process spanned a total of 40,000 iterations with a batch size of 8 total (2 per GPU and 4 NVIDIA A40 GPU are used). The AdamW \cite{loshchilov2017decoupled} optimizer is utilized for the optimization, and the base learning rate is set to 0.0001, accompanied by a weight decay of 0.05. In addition, the Poly learning rate strategy is used for learning rate scheduling, and the final learning rate drops to 0.

\subsection{Experiment Results}

Comparative results on the UAVID, UDD6, and FloodNet datasets are presented in Tabs.~\ref{tab:uavid},\ref{tab:udd6}, and \ref{tab:floodnet}, respectively. The combination of RingMo-Aerial with Mask2Former achieves segmentation outcomes that significantly outperform those of other methods.

Furthermore, we present the qualitative results for a selection of methods in different scenarios, which primarily comprise roads, secondary roads, and car parks. As illustrated in Fig.\ref{fig:seg-vis}, RingMo-Aerial consistently produces superior segmentation maps. Among these methods, the RingMo-Aerial approach demonstrates excellent segmentation accuracy and the ability to recognize small targets effectively when combining different architectures. In particular, the RingMo-Aerial + UperNet + ARS-Adapter and RingMo-Aerial + Mask2Former methods demonstrate higher accuracy and clarity in the segmentation of small targets (e.g., stationary and moving vehicles), with enhanced boundary and detail capture. This illustrates the significant advantages of our method in complex urban scenes and diverse object classes.

\section{Object Detection}
\label{appendices: Object Detection}
\subsection{Datasets and Evaluation Metrics}

To conduct a comprehensive comparison of object detection methods, we selected four unimodal benchmark datasets, including the VisDrone-DET\cite{du2019visdrone}, UAVDT\cite{yu2020unmanned} and ShipDataset\cite{zhao2023multiship}, as well as one multimodal benchmark dataset, DroneVehicle\cite{sun2022drone} and our collection of SAR airborne remote sensing data, to demonstrate the performance of the representative methods discussed in this paper. 

\begin{itemize}
\item \textbf{VisDrone-DET.} The Visdrone-DET dataset is the detection task part of the VisDrone dataset. The resolution of the images is about 2k. The RingMo-Aerial experiment crops the original image to 640×640 and uses multi-scale cropping and resizing for enhancement. The training set and validation set are used for training, and the labeled test-dev set is used for testing.
\item \textbf{UAVDT.} The UAVDT dataset is another popular ARS detection dataset, containing 23,258 images for training and 15,069 images for testing with an average resolution of 1080 × 540 pixels. The RingMo-Aerial experiment uses multi-scale cropping and resizing for enhancement.
\item \textbf{ShipDataset.} The ShipDataset, captured at a resolution of $3840\times2160$, encompasses five distinct scenarios in Shanghai, China, characterized by varying shooting angles and lighting conditions. This dataset has $96.97\%$ small ships. The images within this dataset predominantly feature congested ship traffic. The RingMo-Aerial experiment uses multi-scale resizing for enhancement.

\item \textbf{DroneVehicle.} The DroneVehicle dataset is a large-scale dataset specifically designed for drone-based RGB-Infrared cross-modality vehicle detection tasks. It includes 28,439 pairs of RGB-Infrared images, totaling 56,878 images, covering various urban scenarios such as roads, residential areas, and parking lots under different lighting conditions, including day, night, and dark night. The RingMo-Aerial experiment uses multi-scale cropping and resizing for enhancement.

\item \textbf{AerialSARData.} To evaluate the performance of our foundational model across various modalities, we employed a hexacopter drone (1800 Pro) equipped with a miniature Synthetic Aperture Radar (MS-102) mounted on its underside. This setup was utilized to collect a dataset of SAR-based aerial remote sensing for vehicle detection. The dataset comprises a total of 711 images, with the training set consisting of 536 images and the test set comprising 175 images, all at a resolution of 800x800 pixels. The images encompass a diverse range of scenarios, including airports, roads, and residential areas.

\end{itemize}

\begin{table}[t]
\setlength{\abovecaptionskip}{1pt}
\setlength{\belowcaptionskip}{1pt}
\centering
\caption{Object Detection: Comp. on VisDrone-DET}
\label{tab:visdrone}
\begin{threeparttable}
\begin{tabular}{lllll}
\toprule
Methods                     & Backbone              & mAP           & mAP$_{50}$    & mAP$_{75}$    \\
\midrule
Light-RCNN\cite{li2017light}                  & ResNet-50            & 16.5          & 32.8          & 15.1          \\
CornerNet\cite{law2018cornernet}                   & ResNet-50             & 17.4          & 34.1          & 15.8          \\
RetinaNet\cite{lin2017focal}                   & ResNet-50             & 11.8          & 21.4          & 11.6          \\
RetinaNet\cite{lin2017focal}                   & Swin-Base             & 22.6          & 38.1          & 23.5          \\
YOLOv4\cite{bochkovskiy2020yolov4}                      & CSPDarknet-53         & -             & 38.7          &               \\
Cascade   R-CNN\cite{cai2019cascade}             & ResNet-50             & 16.1          & 31.9          & 15.0          \\
Cascade   R-CNN\cite{cai2019cascade}             & Swin-Base             & 28.4          & 45.2          & 30.1          \\
CEASC\cite{lin2024centralised}                       & ResNet-18             & 28.7          & 50.7          & 28.4          \\
SOD-YOLOv7\cite{chen2024small}                  & -                     & -             & 53.2          &               \\
MFFSODNet\cite{jiang2024mffsodnet}                   & -                     & -             & 45.5          &               \\
GLSAN\cite{deng2020global}                       & ResNet-50             & 25.8          & 51.5          & 22.9          \\
QueryDet\cite{yang2022querydet}                    & ResNet-50             & 28.3          & 48.1          & 28.7          \\
VistrongerDet\cite{wan2021vistrongerdet}               & ResNet-50             & 33.9          & 57.3          &               \\
DMNet\cite{li2020dmnet}                       & ResNet-101            & 29.4          & 49.3          & 30.6          \\
HRDNet\cite{liu2021hrdnet}                      & ResNet-101            & 28.3          & 49.2          & 28.1          \\
ClusDet\cite{yang2019clustered}                     & ResNet-101            & 26.7          & 50.4          &               \\
ClusDet\cite{yang2019clustered}                     & ResNeXt-101           & 28.4          & 53.2          &               \\
SDPDet\cite{yin2024sdpdet}                      & ResNet-101            & 34.2          & 57.8          & 34.9          \\
AMRNet\cite{wei2020amrnet}                      & ResNet-101            & 31.7          & 52.6          & 33.0          \\
OGMN\cite{li2023ogmn}                        & ResNeXt-101            & 35.0          & 59.7          & 35.8          \\
DDQ-DETR\cite{zhang2023dense}                    & Swin-Base             & 36.1          & 57.1          & 37.2          \\
\midrule
RingMo-Aerial   (C) \tnote{*}          & RingMo-Aerial          & 31.1          & 51.8          & 32.5          \\
RingMo-Aerial   (A) \tnote{*}          & RingMo-Aerial          & 31.0          & 51.4          & 32.1          \\
RingMo-Aerial   (D) \tnote{*} & RingMo-Aerial & \textbf{38.6} & \textbf{63.3} & \textbf{39.4} \\
\bottomrule 
\end{tabular}
\begin{tablenotes}
\footnotesize
\item[*] RingMo-Aerial (C), (A), (D) represent the RingMo aerial backbone with Cascade RCNN detection algorithm, ARS-Adapter and Cascade RCNN algorithm, and DDQ DETR algorithm. Same below.
\end{tablenotes}
\end{threeparttable}
\end{table}

\begin{table}[t]
\centering
\setlength{\abovecaptionskip}{1pt}
\setlength{\belowcaptionskip}{1pt}
\caption{Object Detection: Comp. on UAVDT}
\label{tab:uavdt}
\begin{tabular}{lllll}
\toprule
Methods                     & Backbone              & mAP           & mAP$_{50}$    & mAP$_{75}$    \\
\midrule
GLSAN\cite{deng2020global}                       & ResNet-50             & 17.0          & 28.1          & 18.8          \\
DMNet\cite{li2020dmnet}                        & ResNet-50             & 14.7          & 24.6          & 16.3          \\
ClusDet\cite{yang2019clustered}                     & ResNet-50             & 13.7          & 26.5          & 12.5          \\
AMRNet\cite{wei2020amrnet}                      & ResNet-50             & 18.2          & 30.4          & 19.8          \\
DSHNet\cite{fu2024dshnet}                      & ResNet-50             & 17.8          & 30.4          & 19.7          \\
EfficientDet\cite{tan2020efficientdet}                & Efficient-B7          & 13.1          & 31.8          & 10.9          \\
YOLOv4\cite{bochkovskiy2020yolov4}                      & CSPDarknet-53         & 13.4          & 30.6          & 11.3          \\
SDPDet\cite{yin2024sdpdet}                      & ResNet-50             & 20.0          & 32.0          & 23.1          \\
QueryDet\cite{yang2022querydet}                    & ResNet-50             & 14.3          & 27.2          & 16.6          \\
Cascade   R-CNN\cite{cai2019cascade}             & Swin-Base             & 19.0          & 31.4          & 19.2          \\
OGMN\cite{li2023ogmn}                        & ResNet-50            & 20.9          & 34.5          & 23.2          \\
OGMN\cite{li2023ogmn}                         & ResNet-101            & 24.2          & 39.9          & 26.8          \\
\midrule
RingMo-Aerial   (C)          & RingMo-Aerial          & 21.4          & 33.1          & 21.9          \\
RingMo-Aerial   (D) & RingMo-Aerial & \textbf{25.2} & \textbf{39.9} & \textbf{27.1} \\
\bottomrule 
\end{tabular}
\end{table}

\begin{table}[t]
\centering
\setlength{\abovecaptionskip}{1pt}
\setlength{\belowcaptionskip}{1pt}
\caption{Object Detection: Comp. on ShipDataset}
\label{tab:ship}
\begin{tabular}{lllll}
\toprule
Methods                     & Backbone              & mAP           & mAP$_{50}$    & mAP$_{s}$    \\
\midrule
RetinaNet\cite{lin2017focal}                    & RetinaNet             &76.0       &98.6         &40.4          \\
UAV-YOLO\cite{shen2022object}                      &DarkNet-53             &86.3       &99.5         &-         \\
TPH-YOLOv5\cite{zhu2021tph}                   &CSPDarknet53+SPP       &89.6       &99.2         &-         \\
YOLOv7\cite{wang2023yolov7}                       &EfficientNet-B0        &88.7       &\textbf{99.6}        &-         \\
NanoDet-Plus\cite{lyunanodet}                &MobileNetV2            &83.1       &96.4        &41.8         \\
YOLOX-L\cite{ge2021yolox}                       &CSPDarkNet-53          &68.5       &93.5        &27.6          \\
Center-Net++\cite{duan2023centernet++}                &ResNet-101             &88.8       &99.0        &70.8          \\
YOLOX-Tiny\cite{ge2021yolox}                    &CSPDarkNet-53          &71.1       &96.4         &30.1          \\
PP-PicoDet-L\cite{yu2021pp}                    &ESNet                  &\textbf{89.7}       &98.9        &67.3        \\
Cascade   R-CNN\cite{cai2019cascade}             & Swin-Base             & 88.3          & 99.0          & 62.6          \\
\midrule
RingMo-Aerial(C)              & RingMo-Aerial          &83.7      &99.1        & \textbf{72.3}          \\

\bottomrule 
\end{tabular}
\end{table}

\begin{table}[t]
\centering
\setlength{\abovecaptionskip}{1pt}
\setlength{\belowcaptionskip}{1pt}
\caption{Object Detection: Comp. on DroneVehicle}
\label{tab:dronevehicle}
\begin{tabular}{lllll}
\toprule
Methods             &Modality        &Backbone                       & mAP$_{50}$\\
\midrule
Faster R-CNN(OBB)\cite{ren2015faster} &RGB &ResNet101 &44.6       \\
RoITransformer\cite{ding2019learning}  &RGB  &ResNet101  &47.9       \\
YOLOv7\cite{wang2023yolov7}  &RGB  &EfficientNet-B0  &68.5       \\
$S^2$ANet\cite{han2021align} &RGB  &RetinaNet  &57.3       \\
Oriented R-CNN\cite{xie2021oriented}  &RGB  &ResNet101  &62.3       \\
Cascade R-CNN\cite{cai2019cascade}  &RGB & Swin-Base &69.5       \\
Faster R-CNN(OBB)\cite{ren2015faster}  &IR  &ResNet101 &44.6       \\
RoITransformer\cite{ding2019learning}  &IR  &ResNet101   &59.2       \\
YOLOv7\cite{wang2023yolov7}  &IR  &EfficientNet-B0  &66.7       \\
$S^2$ANet\cite{han2021align}  &IR  &RetinaNet  &64.8       \\
Oriented R-CNN\cite{xie2021oriented}  &IR  &ResNet101  &65.5       \\
Cascade R-CNN\cite{cai2019cascade}  &IR & Swin-Base &69.8       \\
UA-CMDet\cite{sun2022drone}  &RGB+IR  &EfficientNet-B0  &64.0       \\
Halfway Fusion(OBB)\cite{wagner2016multispectral}  &RGB+IR  &Halfway Fusion  &68.2       \\
CIAN(OBB)\cite{zhang2019cross}  &RGB+IR  &VGG16  &70.2       \\
Dual-YOLO\cite{bao2023dual}  &RGB+IR    &Dual-YOLO          & 71.5          \\
Cascade R-CNN\cite{cai2019cascade}  &RGB+IR & Swin-Base &69.8       \\
\midrule
RingMo-Aerial(C)      &RGB+IR    & RingMo-Aerial          & \textcolor{black}{\textbf{73.5}}          \\
\bottomrule 
\end{tabular}
\end{table}

\begin{table}[t]
\centering
\setlength{\abovecaptionskip}{1pt}
\setlength{\belowcaptionskip}{1pt}
\caption{Object Detection: Comp. on AerialSARData}
\label{tab:sar}
\begin{threeparttable}
\begin{tabular}{lllll}
\toprule
Methods                     & Backbone              & mAP           & mAP$_{50}$        \\
\midrule
RetinaNet\cite{lin2017focal}                    & RetinaNet             &22.9       &64.3                  \\
% YOLOX-S\cite{ge2021yolox}                       &CSPDarkNet-53          &34.5       &85.0                  \\
Cascade   R-CNN\cite{cai2019cascade}             & ResNet101             &34.5      &81.3                    \\

\midrule
RingMo-Aerial(A) \tnote{*}             & RingMo-Aerial          &30.9      &77.6                  \\
% RingMo-Aerial(Y) \tnote{*}             & RingMo-Aerial          &34.0      &81.5                  \\
\textbf{RingMo-Aerial(C)} \tnote{*}             & RingMo-Aerial          &\textbf{34.9}      &\textbf{82.6}                  \\
% RingMo\cite{sun2022ringmo}             & RingMo             &36.2      &84.4                    \\
\bottomrule 
\end{tabular}
\begin{tablenotes}
\footnotesize
\item[*] RingMo-Aerial (A), (C) represent the RingMo-Aerial backbone with ARS-Adapter and Cascade RCNN algorithm.
\end{tablenotes}
\end{threeparttable}
\end{table}

In our object detection experiments, the evaluation metrics employed are mean average precision (\textbf{mAP}), mAP at an IoU threshold of 50 (\textbf{mAP$_{50}$}), and mAP at an IoU threshold of 75 (\textbf{mAP$_{75}$}). For datasets containing smaller objects, we also utilize \textbf{mAP$_{s}$}, which represents the mean average precision for detecting small objects (with pixels smaller than 32×32).

\subsection{Experiment Settings}

The experiment of the object detection is based on mmdetection \cite{mmdetection}, Cascade RCNN \cite{cai2019cascade} and DDQ DETR \cite{zhang2023dense} are employed as the detection framework. The training process spanned a total of 12 epochs for Cascade RCNN, with a batch size of 32 total (8 per GPU and 4 NVIDIA A40 GPU are used). For DDQ DETR, 36 epochs were spanned with a batch size of 16 total (4 per GPU and 4 NVIDIA A40 GPU are used). The AdamW \cite{loshchilov2017decoupled} optimizer is utilized for the optimization, and the base learning rate is set to 0.0001, accompanied by a weight decay of 0.05. In addition, the learning rate is decreased by 10 after the 8th and 11th epochs for Cascade RCNN and after the 27th and 33rd epochs for DDQ DETR.

\begin{figure*}
\centering
\includegraphics[width=0.90\linewidth]{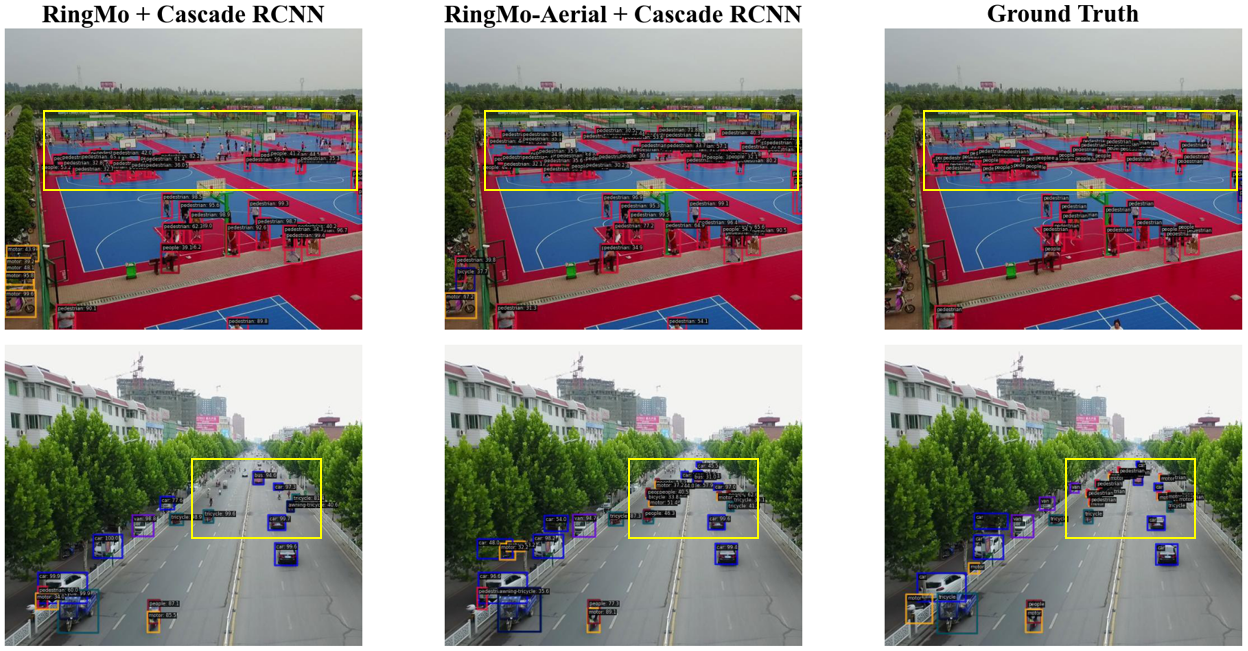}
\caption{RingMo-Aerial combined with Cascade RCNN achieves superior detection accuracy compared to the original RingMo model, particularly in complex environments and small object scenarios. It demonstrates reduced false positives and improved bounding box alignment with Ground Truth, highlighting its effectiveness for aerial imagery detection.}
\label{fig:det-vis}
\end{figure*}

\begin{figure}
\centering
\includegraphics[width=0.9\linewidth]{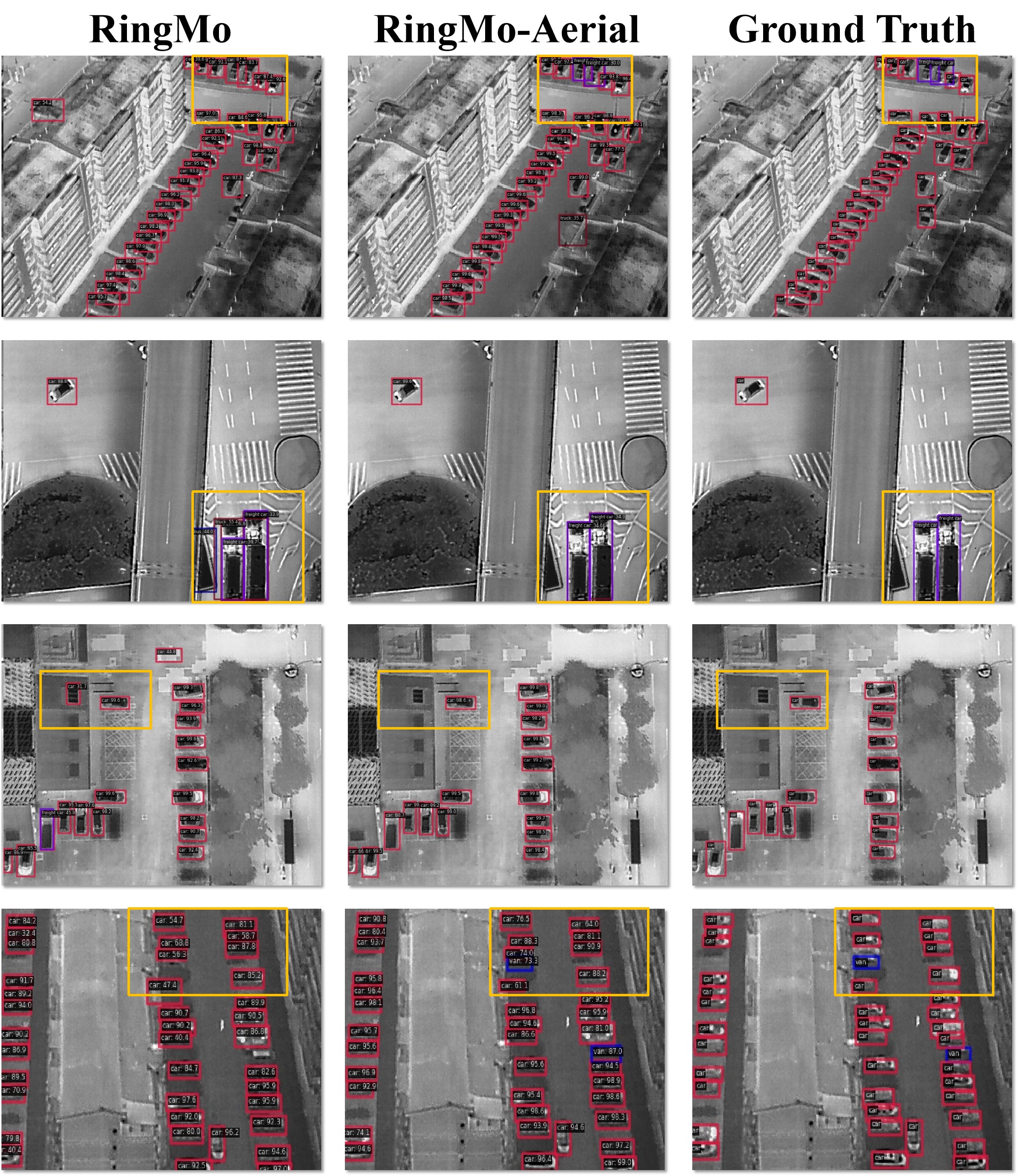}
\caption{Qualitative comparison of object detection results on infrared images from the DroneVehicle-IR dataset using different methods. The figure highlights differences in the detection quality of vehicles, road elements, and pedestrian areas, with emphasized regions showcasing variations in accuracy and detail among the compared methods.}
\label{fig:vis-ir}
\end{figure}

\begin{figure}
\centering
\includegraphics[width=0.9\linewidth]{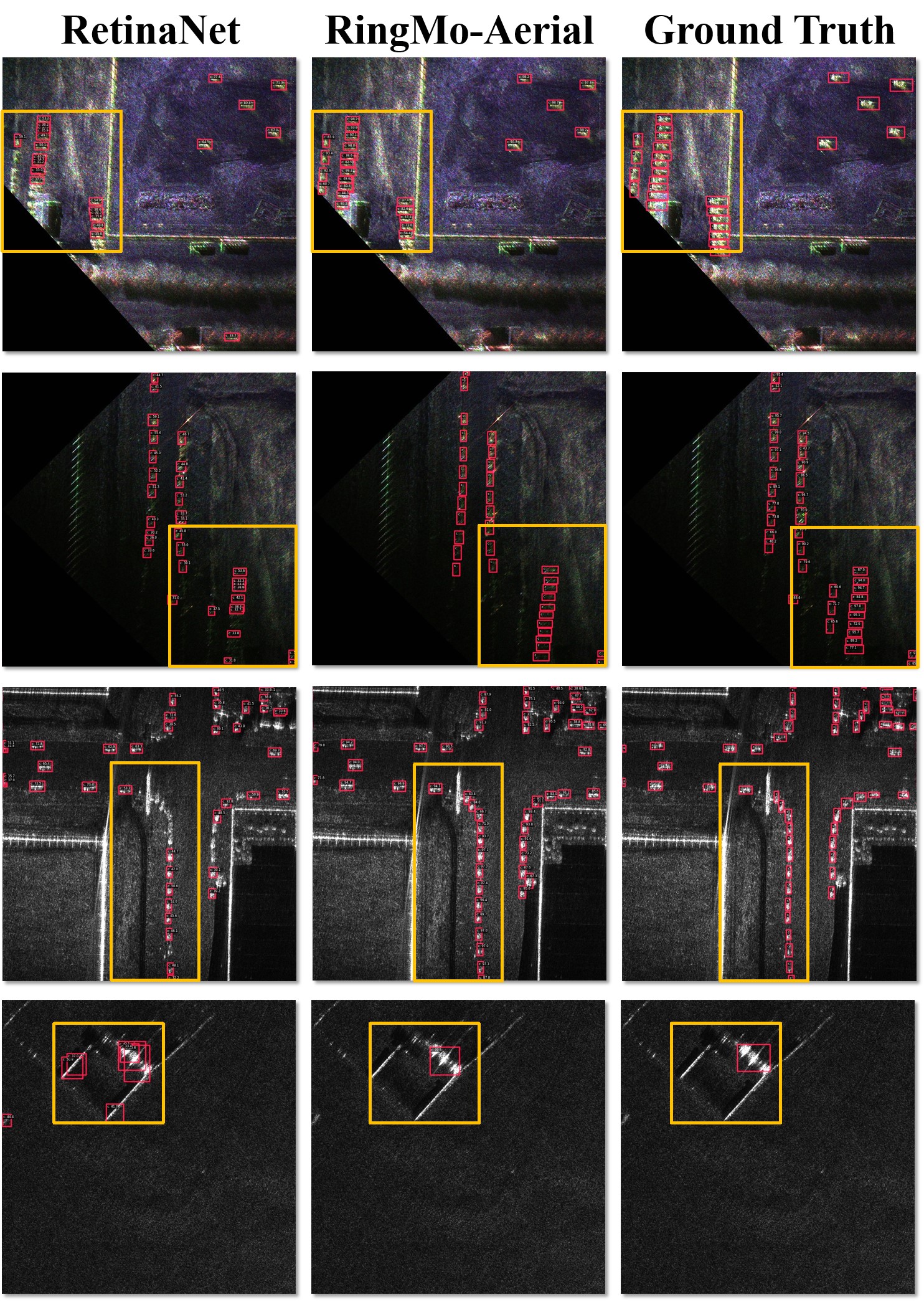}
\caption{Qualitative comparison of object detection results on the AerialSARData dataset using different methods. The figure highlights differences in detection performance for SAR images, particularly focusing on identifying objects of interest with highlighted regions showing variations in accuracy and detection quality among the methods compared.}
\label{fig:vis-sar}
\end{figure}

\subsection{Experiment Results}

An analysis of all tables reveals that the detection algorithms based on the proposed method as a pre-trained model have achieved state-of-the-art (SOTA) performance. On unimodal datasets such as VisDrone-DET (as shown in Tab.\ref{tab:visdrone}) and UAVDT (as shown in Tab.\ref{tab:uavdt}), RingMo-Aerial combined with DDQ DETR (RingMo-Aerial(D))significantly outperforms other methods. The experimental results further demonstrate that RingMo-Aerial, compared to Swin-Base in Cascade RCNN and DDQ DETR, is a more suitable foundational model for UAV-based object detection tasks.

What's more, RingMo-Aerial(C) exhibits the best performance in small object detection (mAP$_{s}$) on the ShipDataset, which primarily targets small vessels (as shown in Tab.\ref{tab:ship}), and PP-PicoDet-L follows. In contrast, YOLOX-L and YOLOX-Tiny show the worst results. Notably, RingMo-Aerial(C) consistently excels in small object detection across all tasks, further validating its superiority in complex scenarios.

For the multimodal dataset, DroneVehicle (as shown in  Tab.\ref{tab:dronevehicle}), RingMo-Aerial (C) achieves the best performance in the RGB+IR modality with an mAP$_{50}$ of 71.6, significantly outperforming other methods. Although methods like Dual-YOLO and CIAN(OBB) also show strong performance in the RGB+IR modality, they still fall short compared to RingMo-Aerial. This indicates that RingMo-Aerial holds significant advantages in handling cross-modality object detection tasks, particularly in UAV image applications. Additionally, the outstanding performance of RingMo-Aerial in multimodal scenarios further validates its effectiveness as a general pre-trained model, particularly in providing higher accuracy and robustness in object detection tasks across diverse and complex environments. 

In the domain of Synthetic Aperture Radar (SAR) datasets (as shown in Tab.\ref{tab:sar}), the RingMo-Aerial backbone network, integrated with various detection algorithms such as Cascade R-CNN, and the proposed ARS-Adapter, has demonstrated a notable enhancement in performance. Mainly, the RingMo-Aerial(C) configuration has achieved an mAP of 36.2\% and a mAP$_{50}$ of 84.4\%, highlighting its superiority in SAR modality. These results substantiate the efficacy of the RingMo-Aerial backbone in processing SAR imagery. Furthermore, the performance of RingMo-Aerial on multimodal datasets reinforces its viability as a universal pre-trained model, especially in enhancing precision and robustness for object detection tasks across diverse and intricate environments. These findings underscore the potential application value of RingMo-Aerial in the field of remote sensing, specifically for aerial object detection tasks utilizing SAR data. 

Additionally, we present the qualitative results of several methods across different scenarios, primarily focusing on roads and sports fields. For optical images, as shown in Fig.\ref{fig:det-vis}, the RingMo-Aerial detection model demonstrates superior performance, particularly in detecting small-scale targets. In the sports field scenario, RingMo-Aerial effectively identifies distant groups of people, even those not annotated in the test set, while RingMo fails to detect these groups. Similarly, RingMo-Aerial accurately detects distant vehicles in road scenarios, whereas RingMo underperforms. RingMo-Aerial also demonstrates outstanding small target recognition capabilities for SAR and IR images. As shown in Fig.\ref{fig:vis-ir} and \ref{fig:vis-sar}, RingMo tends to misclassify densely overlapping vehicles at the end of the road, while RingMo-Aerial maintains a high level of class confidence. The visualized results confirm the model's excellent performance on multimodal data, including optical, SAR, and IR images. RingMo-Aerial's exceptional performance across multiple modalities proves its strong generalization ability and highlights its great potential as the first foundational model for ARS.

\section{Change Detection}
\label{appendices: Change Detection}
\subsection{Datasets and Evaluation Metrics}

For semantic segmentation, the LEVIR \cite{chen2020spatial} and CDD \cite{lebedev2018change} datasets are employed.

\begin{itemize}
\item \textbf{LEVIR.} The LEVIR-CD is a publicly available, extensive remote sensing change detection dataset. It comprises 637 pairs of high-resolution images, each with a dimension of 1024 × 1024 pixels at a 0.5-meter resolution. Adhering to the standard dataset partitioning, we have processed the original images by extracting non-overlapping patches, each measuring 256 × 256 pixels. Consequently, this results in a distribution of 7120, 1024, and 2048 image patches allocated for training, validation, and testing phases.

\item \textbf{CDD.} The CDD dataset is a comprehensive collection of bi-temporal remote sensing imagery, comprising seven pairs of images with a resolution of 4725×2700 pixels and an additional four pairs with dimensions of 1900×1000 pixels. These images have been meticulously synchronized and cropped to yield 16,000 pairs of 256×256 pixel patches. This curated subset is strategically divided into a training set of 10,000 pairs, a validation set of 3,000 pairs, and a test set of the remaining 3,000 pairs. To enhance the realism and applicability of the trained models, the dataset incorporates seasonal variations, thereby providing a more challenging and convincing foundation for network training. 

\end{itemize}

\begin{table}
\centering
\setlength{\abovecaptionskip}{1pt}
\setlength{\belowcaptionskip}{1pt}
\setlength{\abovecaptionskip}{1pt}
\setlength{\belowcaptionskip}{1pt}
\caption{Change Detection: Comp. on LEVIR-CD and CDD}
\label{tab:cd}
\begin{tabular}{lcc|cc}
\hline
\multirow{2}{*}{Methods} & \multicolumn{2}{c|}{LEVIR-CD} & \multicolumn{2}{c}{CDD} \\
\cline{2-5}
 & F1    & IoU   & F1    & IoU    \\
\hline
FC-Siam-Di \cite{daudt2018fully} & 86.31 & 83.31 & 70.61 & 54.57 \\
FC-Siam-Conc \cite{daudt2018fully} & 83.69 & 76.77 & 75.11 & 60.14 \\
IFNet \cite{zhang2020deeply} & 88.13 & 82.93 & 90.30 & 71.91 \\
STANet \cite{chen2020spatial} & 87.30 & 77.40 & 84.12 & 72.22 \\
BIT \cite{chen2021remote} & 89.31 & 89.37 & 88.90 & 80.01 \\
SNUNet \cite{fang2021snunet} & 88.16 & 87.17 & 83.89 & 72.11 \\
ChangeFormer \cite{bandara2022transformer} & 90.40 & 88.80 & 89.83 & 81.53 \\
% RingMo \cite{bandara2022transformer} & 90.40 & 88.80 & - & - \\
RingMo \cite{sun2022ringmo} & 89.53 & 82.38 & 90.90 & 84.09 \\
\midrule
\textbf{RingMo-Aerial} & \textbf{92.71} & \textbf{87.10} & \textbf{94.62} & \textbf{90.09} \\
\hline
\end{tabular}
\end{table}

\subsection{Experiment Settings}

Our model adopts BIT with Swin Transformer as the backbone. The experiments use overall accuracy (OA) and F1 and IoU scores about change detection as the primary evaluation metrics. The model is trained with an AdamW optimizer with a weight decay of 0.01 and beta values equal to (0.9, 0.999). The LR is 2e-3 for the LEVIR-CD dataset and linearly decays to 0 until trained for 200 epochs, with a total batch size 16 for datasets (8 per GPU, using 2 NVIDIA A40 GPUs).

\begin{figure*}
\centering
\includegraphics[width=0.90\linewidth]{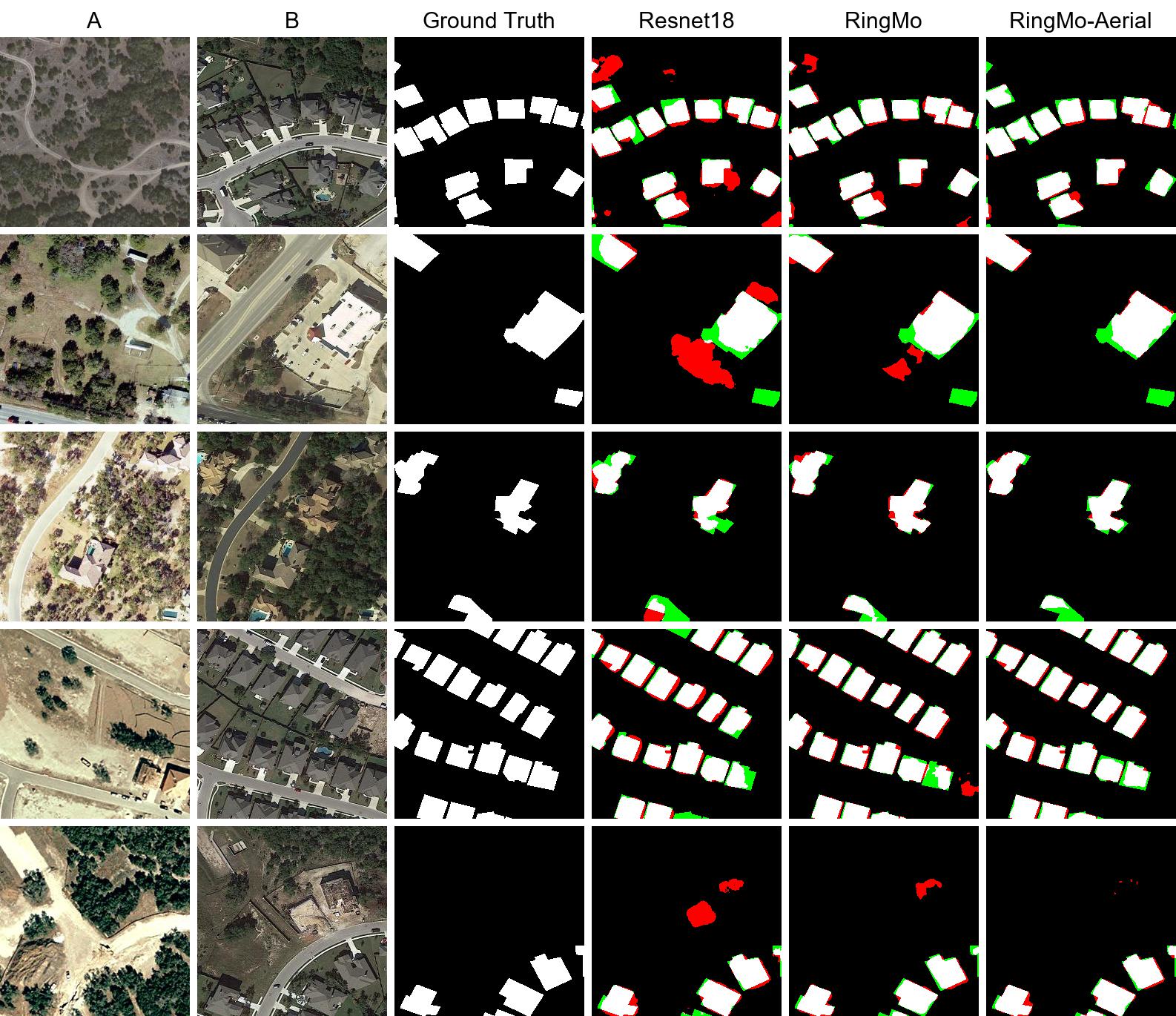}
\caption{Comparison of Detection Visualization Results Using Different Backbone Networks with the BIT Algorithm on the LEVIR Dataset. That shows the results of aerial images, where RingMo-Aerial outperforms ResNet18 and RingMo, with fewer false positives and false negatives. RingMo-Aerial closely matches the Ground Truth, demonstrating the best accuracy in change Detection. The black region represents true negative, the white region represents true positive, the green region represents false negative, the red region
represents false positive.}
\label{fig:cd-vis}
\end{figure*}

\subsection{Experiment Result}

The comparison results are shown in Tab.\ref{tab:cd}. RingMo-Aerial combined with BIT can achieve change detection results that far exceed other methods. 
In a rigorous comparative evaluation against current state-of-the-art technologies, RingMo-Aerial proved its capabilities in the specialized field of remote sensing change detection. When evaluated against LEVIR-CD and CDD datasets, the model demonstrated a clear competitive advantage, highlighting the robustness and effectiveness of the backbone network integrated into our architecture.

The comparison of visual results on the LEVIR dataset reveals significant differences in the performance of models with different backbones in building detection and segmentation tasks, as shown in Fig.\ref{fig:cd-vis}. The RingMo-Aerial model demonstrates superior accuracy, with fewer false positives (red areas) and false negatives (green areas), highlighting its advantages in processing remote sensing images. In contrast, the ResNet18 and Swin-Base models exhibit more false positives and false negatives, particularly in complex building layouts, where RingMo-Aerial shows greater robustness.

RingMo-Aerial's commendable performance not only makes it a key player in the field but also highlights the importance of the impact of foundation model innovation. These elements are critical to pushing the frontiers of remote sensing analysis and setting new benchmarks for future research and applications.

\section{Scene Classification}
\label{appendices: Scene Classification}
\subsection{Datasets and Evaluation Metrics}

\begin{table}[htbp]
\centering
\setlength{\abovecaptionskip}{1pt}
\setlength{\belowcaptionskip}{1pt}
\caption{Scene Classification: Comp. on AID and RESISC-45}
\begin{tabular}{lcc}
\toprule
\textbf{Model} & \textbf{AID (OA)} & \textbf{RESISC-45 (OA)} \\
 & \textbf{(TR=20\%/50\%)} & \textbf{(TR=10\%/20\%)} \\
\midrule
GASSL \cite{ayush2021geography}  & 93.55/95.92 & 90.86/93.06 \\
SeCo \cite{manas2021seasonal}  & 93.47/95.99 & 89.64/92.91 \\
SatMAE \cite{cong2022satmae}  & 95.02/96.94 & 91.72/94.10 \\
RingMo \cite{sun2022ringmo} & 96.90/98.34 & 94.25/95.67 \\
RVSA \cite{wang2022advancing}  & 97.03/98.50 & 93.93/95.69 \\
DINO-MC \cite{wanyan2023dino} & 95.16/97.09 & - \\
TOV \cite{tao2023tov} & 95.16/97.09 & 90.97/93.79 \\
SSL4EO \cite{wang2023ssl4eo} & 91.06/94.74 & 87.60/91.27 \\
CMID \cite{muhtar2023cmid} & 96.11/97.79 & 91.84/94.07 \\
CACo \cite{mall2023change} & 90.88/95.05 & 88.28/91.94 \\
CROMA \cite{fuller2024croma} & 96.44/97.58 & 92.63/95.04 \\
SatLas \cite{bastani2023satlaspretrain} & 94.96/97.38 & 92.16/94.70 \\
GFM \cite{mendieta2023towards} & 95.47/97.09 & - \\
Scale-MAE \cite{reed2023scale} & 96.44/97.58 & 92.63/95.04 \\
SkySense \cite{guo2024skysense} & \textbf{97.68}/\textbf{98.60} & \textbf{94.85}/\textbf{96.32} \\
RingMo \cite{sun2022ringmo} & 96.90/98.34 & 94.25/95.67 \\
\midrule
\textbf{RingMo-Aerial} & 95.81/96.46 & 92.28/95.65 \\
\bottomrule
\end{tabular}
\label{tab:classification}
\end{table}

For scene classification, the AID \cite{xia2017aid} and RESISC-45 \cite{cheng2017remote} datasets are employed.

\begin{itemize}
\item \textbf{AID.} The AID dataset is a large-scale data set used to evaluate the performance of aviation scene classification. It contains images collected from Google Earth and covers 30 scenes.

\item \textbf{NWPU-RESISC45.} The NWPU-RESISC45 dataset is a publicly available Remote Sensing Image Scene Classification (RESISC) benchmark. The dataset contains 31,500 images covering 45 scene classes, with 700 images in each class. 
\end{itemize}

In our experiment of scene classification, the training proportions in the AID dataset were set to TR = 20\% and TR = 50\%, respectively. In the NWPU-RESISC dataset, the proportion of training samples was set to TR = 10\% and TR = 20\%, respectively. The experiments use overall accuracy (OA) as a performance metric, the most commonly used performance metric in scene classification.

\subsection{Experiment Settings}

The experiment of the scene classification is based on mmpretrain \cite{2023mmpretrain},  Cascade RCNN \cite{cai2019cascade} is employed as the classification framework. The training process spanned a total of 300 calendar elements, with a total batch size of 16 for dataset AID (8 per GPU, using 2 NVIDIA A40 GPUs) and 128 for dataset NWPU-RESISC45 (64 per GPU, using 2 NVIDIA A40 GPUs). The AdamW \cite{loshchilov2017decoupled} optimizer is utilized for the optimization, and the base learning rate is set to 0.0005, accompanied by a weight decay of 0.05. In addition, the Poly learning rate strategy is used for learning rate scheduling, and the final learning rate drops to 0.

\subsection{Experiment Results}

The comparison results are shown in Tab.\ref{tab:classification}. Although RingMo-Aerial does not perform as well as specialized satellite models like SkySense in nadir-view classification, it still demonstrates strong performance, reflecting its robust generalization capability.

\begin{table*}[htbp]
\centering
\setlength{\abovecaptionskip}{1pt}
\setlength{\belowcaptionskip}{1pt}
\caption{3D-Reconstruction: Comp. on LuoJia-MVS and WHU}
\label{tabMVSResults}
    \begin{threeparttable}
    \centering
    \begin{tabular}{c|c|ccc|ccc}
    \toprule
    \multirow{2}{*}{Number of Views} & \multirow{2}{*}{Model}  & \multicolumn{3}{c|}{LuoJia-MVS} & \multicolumn{3}{c}{WHU}\\
    ~ & ~ & MAE & $<$0.6m & $<$3-interval & MAE & $<$0.6m & $<$3-interval\\ 
    \midrule
    \multirow{8}{*}{Three views \tnote{*}} & PatchmatchNet\cite{barnes2009patchmatch} & 0.252 & 92.7  & 87.2 & 0.173 & 96.5 & 94.8 \\
    ~ & Fast-MVSNet\cite{yu2020fast}   & 0.194 & 95.7 & 92.0 & 0.184 & 95.5 & 94.1 \\
    ~ & MVSNet\cite{yao2018mvsnet}        & 0.172 & 96.1 & 92.4 & 0.190 & 95.0 & 94.3 \\
    ~ & R-MVSNet\cite{yao2019recurrent}      & 0.177 & 96.0 & 93.5  & 0.183 & 95.3 & 93.5  \\
    ~ & RED-Net\cite{liu2020novel}       & 0.109 & 98.2 & 96.9  & 0.112 & \textbf{98.1} & \textbf{97.9}  \\
    ~ & Cas-MVSNet\cite{gu2020cascade}    & 0.103 & 98.4 & 97.1  & \textbf{0.111} & 97.7 & 97.6  \\  
    ~ & \textbf{RingMo-Aerial}       & \textbf{0.095}  & \textbf{98.7} & \textbf{97.8}  & 0.121 & 97.8 & 97.5  \\
    \midrule
    \multirow{8}{*}{Five views \tnote{*}} & PatchmatchNet\cite{barnes2009patchmatch}  & 0.283 & 90.4 & 84.1  & 0.160 & 96.9 & 95.0 \\
    ~ & Fast-MVSNet\cite{yu2020fast}    & 0.357 & 84.6 & 74.9  & 0.157 & 96.1 & 95.6 \\
    ~ & MVSNet\cite{yao2018mvsnet}         & 0.270 & 91.2 & 81.8  & 0.160 & 95.8 & 95.5\\
    ~ & R-MVSNet\cite{yao2019recurrent}       & 0.259 & 92.3 & 86.7 & 0.173 & 95.4 & 93.8\\
    ~ & RED-Net\cite{liu2020novel}        & 0.156 & 94.9 & 90.5 & 0.104 & \textbf{98.1} & 97.9\\
    ~ & Cas-MVSNet\cite{gu2020cascade}     & 0.141 & 97.9 & 95.4 & 0.095 & 97.8 & 97.8\\
    ~ & \textbf{RingMo-Aerial}        & \textbf{0.122} & \textbf{98.1} & \textbf{96.5} & \textbf{0.091} & 98.0 & \textbf{97.9} \\ 
    \bottomrule
    \end{tabular}
    \begin{tablenotes}
    \footnotesize
    \item[*] `Three Views' and `Five Views' represent the number of views for the network's input is three and five, respectively.
    \end{tablenotes}
    \end{threeparttable}
\end{table*}

\section{3D-Reconstruction}
\label{appendices: 3D-Reconstruction}
\subsection{Datasets and Evaluation Metrics}

For 3D-Reconstruction, the LuoJia-MVS\cite{li2023hierarchical} and WHU \cite{liu2020novel} datasets are employed.

\begin{itemize}

\item \textbf{WHU.} The WHU dataset is a synthetic aerial dataset designed for large-scale multi-view stereo reconstruction tasks. It was generated from a highly accurate 3D surface model and includes a complete set of aerial images as well as cropped sub-image sets specifically for deep learning applications. The images in the dataset cover an area of approximately 6.7 × 2.2 square kilometers in Meitan County, Guizhou Province, China, with a ground resolution of 0.1 meters. The WHU dataset provides crucial data support for the study of large-scale Earth surface and urban reconstruction and can be utilized for training and testing multi-view stereo matching algorithms. 

\item \textbf{LuoJia-MVS.} The LuoJia-MVS dataset comprises 7,972 five-view images, each with a spatial resolution of 10 cm, accompanied by pixel-wise depth information and precise camera parameters. These images were generated from a highly accurate digital surface model (DSM) constructed using thousands of stereo aerial images. 

\end{itemize}

% \textcolor{red}{During the experiment, we used the WHU dataset\cite{liu2020novel} and the LuoJia-MVS dataset\cite{li2023hierarchical} to verify the effectiveness of our proposed method. 
% WHU dataset\cite{liu2020novel} is the first large-scale aerial multi-view stereo dataset for remote sensing. It was collected from Meitan, Guizhou, China, and was obtained by taking large-scale photos with a tilted five-eye camera on a drone platform. The area of the entire dataset is approximately $6.7\times 2.2km^2$, and the image resolution is about 0.1m.
% In addition, the tilt angle of each camera to the central camera is 40°. The WHU dataset gathers high-rise buildings, mountains, forests, and factory scenes, providing a good benchmark for remote sensing multi-view stereo. LuoJia-MVS\cite{li2023hierarchical} dataset is a large-scale multi-view stereo dataset collected from a variety of landform types. It was collected in a part of Baiyun, Guiyang, Guizhou, China. The image size of the entire dataset is $768\times 384$, and the image resolution is about 0.1m. The data consists of five views and pixel-by-pixel depth labels. Land types cover cultivated land, forest land, industrial areas, residential areas, etc.
% }

\textcolor{black}{During the experiment, we use three metrics, mean absolute error (MAE), $<$0.6m, and $<$3-interval, to evaluate the performance of the proposed method.
}

\subsection{Experiment Settings}\label{sec:Experimental Setting}

Our work is coded based on the PyTorch framework. The proposed method is trained and tested on NVIDIA A40, and the batch size is set to 1. During the training process, our method uses the Adam optimizer\cite{kingma2017adam} for parameter optimization and trains for 24 epochs, where the parameter $\beta_1$ is set to 0.9, the parameter $\beta_2$ is set to 0.999, and the initial learning rate is 0.001.
In addition, our method is consistent with the existing multi-stage networks\cite{gu2020cascade}. First, multi-scale features are learned through a multi-scale feature extractor, and the scales are 1/16, 1/4, and 1 of the input size, respectively. Then, deep optimization is performed through three stages of profound hypotheses, and the number of hypotheses is 48, 32, and 8, respectively.
% \textcolor{red}{During the experiment, we use three metrics, mean absolute error (MAE), $<$0.6m, and $<$3-interval, to evaluate the performance of the proposed method.
% }

\subsection{Experiment Results}

Tab.\ref{tabMVSResults} provides a performance comparison of our method and other methods on the LuoJia-MVS and WHU datasets.
Obviously, under the premise of taking five views as input, our method achieves the best estimation performance and the lowest prediction error. On the LuoJia-MVS dataset, our method achieves an MAE error of 0.122m, 98.1\% and 96.5\% of $<$0.6m and $<$3-interval accuracy, which is better than the existing advanced method Cas-MVSNet. It improves MAE, $<$0.6m, $<$3-interval by 13.5\% (0.122 vs. 0.141), 0.2\% (98.1\% vs. 97.9\%) and 1.1\% (96.5\% vs.95.4\%), respectively, which is a significant improvement. In addition, on the WHU dataset, our method surpasses the advanced method Cas-MVSNet and achieves an MAE error of 0.091m, a $<$0.6m accuracy of 98.0\% and a $<$3-interval accuracy of 97.9\%. It can be seen that our proposed method also has superior generalization performance in multi-view stereo reconstruction methods.

\section{Object Tracking}
\label{appendices: Object Tracking}
\subsection{Datasets and Evaluation Metrics}

\begin{table}[t]
\centering
\setlength{\abovecaptionskip}{1pt}
\setlength{\belowcaptionskip}{1pt}
\caption{Object Tracking: Comp. on Visdrone-MOT}
\label{tab:visdrone-mot}
\begin{tabular}{lll}
\toprule
Methods                   & MOTA & IDF1 \\
\midrule
SiamMOT\cite{shuai2021siammot}                & 31.9 & 48.3 \\
ByteTrack\cite{zhang2022bytetrack}                     & 35.7 & 37.0 \\
UAVMOT\cite{liu2022multi}                    & 36.1 & 51.0 \\
OCSORT\cite{maggiolino2023deep}                    & 39.0 & 50.4 \\
MOTR\cite{zeng2022motr}                      & 22.8 & 41.4 \\
TrackFormer\cite{meinhardt2022trackformer}                 & 24.0 & 30.5 \\
FOLT\cite{yao2023folt}                     & 42.1 & 56.9 \\
U2MOT\cite{liu2023uncertainty}                 & 52.3 & \textbf{69.0} \\
RingMo\cite{sun2022ringmo}         & 48.4 & 63.1 \\
\midrule
RingMo-Aerial       & \textbf{52.6} & 67.8 \\
\bottomrule 
\end{tabular}
\end{table}

\begin{table}[t]
\centering
\setlength{\abovecaptionskip}{1pt}
\setlength{\belowcaptionskip}{1pt}
\caption{Object Tracking: Comp. on AIR-MOT}
\label{tab:AIR-MOT}
\begin{tabular}{lll}
\toprule
Methods                   & MOTA & IDF1 \\
\midrule
CKDNet+SMTNet\cite{feng2021cross}                     & 64.2 & 64.9 \\
ByteTrack \cite{zhang2022bytetrack}                    & 56.1 & 73.5 \\
OCSORT\cite{cao2023observation}                    & 35.6 & 53.8 \\
StrongSORT\cite{du2023strongsort}                    & 36.1 & 54.1 \\
CenterTrack\cite{zhou2020tracking}                    & 66.3 & 79.3 \\
TGram\cite{he2022multi}                    & 65.7 & 66.8 \\
RSMOT\cite{xiao2022rsmot}                    & 67.0 & 80.4 \\
CFTrack\cite{kong2023cftracker}                    & \textbf{70.8} & \textbf{82.7} \\
\midrule
% RingMo-Aerial+ByteTrack       & \textbf{75.1} & \textbf{86.6} \\
RingMo-Aerial       &68.6 & 81.3 \\
\bottomrule 
\end{tabular}
\end{table}

\begin{table}[t]
\centering
\setlength{\abovecaptionskip}{1pt}
\setlength{\belowcaptionskip}{1pt}
\caption{Object Tracking: Comp. on AIR-HSAO}
\label{tab:AIR-HSAO}
\begin{tabular}{lll}
\toprule
Methods                   & MOTA & IDF1 \\
\midrule
TraDeS\cite{wu2021track}                    & 59.6 & 75.1 \\
CenterTrack\cite{zhou2020tracking}                    & 60.3 & 74.5 \\
DSFNet\cite{xiao2021dsfnet}                    & 60.2 & 75.8 \\
FairMOT \cite{zhang2021fairmot}                   & 54.8 & 71.3 \\
CTracker\cite{peng2020chained}                    & 51.1 & 65.1 \\
ByteTrack\cite{zhang2022bytetrack}               & 60.3 & 74.5 \\
MGTrack\cite{ren2024motion}                 & \textbf{63.5} &\textbf{77.4} \\
\midrule
RingMo-Aerial      & 60.7 & 74.4 \\
\bottomrule 
\end{tabular}
\end{table}

\begin{figure*}
\centering
\includegraphics[width=0.95\linewidth]{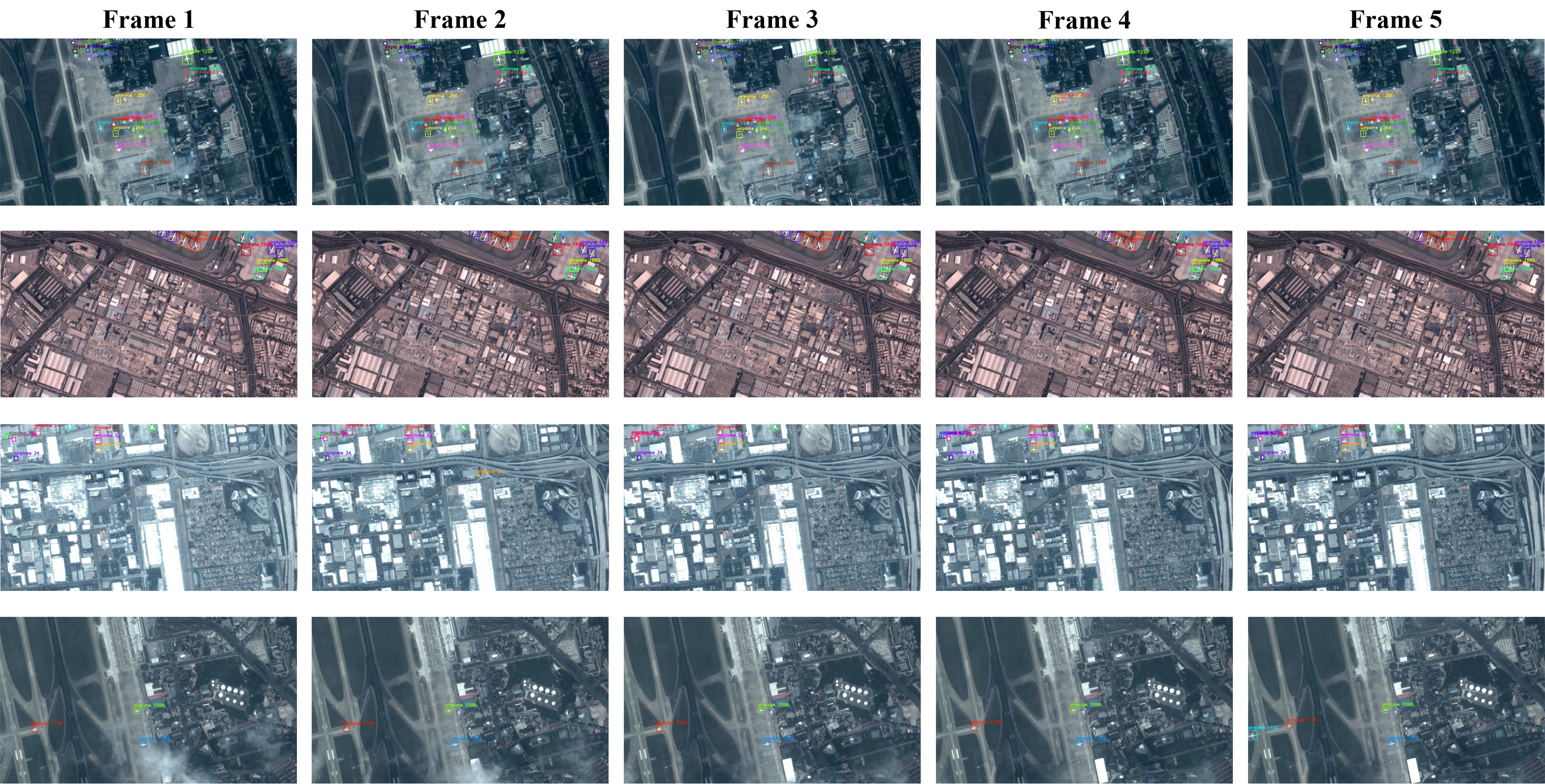}
\caption{Qualitative results on the target tracking dataset AIR-MOT using RingMo-Aerial. Four video sequences are selected, where each target is indicated with different colors. The category and target ID are displayed above the bounding boxes, demonstrating RingMo-Aerial's ability to effectively distinguish and track individual targets.}
\label{fig:mot-vis}
\end{figure*}

For Object Tracking, the following aerial dataset is employed. 

\begin{itemize}
\item \textbf{VisDrone-MOT.} The VisDrone-MOT\cite{chen2021visdrone}  dataset consists of 96 sequences, including five classes such as pedestrian, car, van, bus, and truck. The VisDrone-MOT train and val set are used for training, and the test set is used for the test.
\end{itemize}
In addition to the aerial dataset, we conducted experiments on two satellite remote sensing datasets.
\begin{itemize}
    \item \textbf{AIR-MOT.} The AIR-MOT dataset\cite{he2022multi}, derived from the JiLin-1 satellite, is designed for multi-class multiple object tracking (MOT). It includes two categories: airplanes and ships. The dataset comprises 152 videos, with frame rates ranging from 5 to 10 frames per second, a spatial resolution of 1–1.2 meters, and a pixel resolution of 1920 × 1080. The target categories in this dataset move relatively slowly.

\item \textbf{AIR-HSAO.} The AIR-HSAO dataset\cite{ren2024motion} is designed for tracking high-speed aerial objects derived from the JiLin-1 satellite. It contains 199 videos, though fewer instances are present due to the rapid movement of these objects. The dataset spans 20 diverse regions, including urban areas, harbors, and airports, offering complex backgrounds that challenge object detection and tracking.
\end{itemize}

In our experiment of object tracking, Multiple Object Tracking Accuracy (\textbf{MOTA}) and Identification Flatness (\textbf{IDF1}) are used as evaluation metrics.

\subsection{Experiment Settings}

In the proposed experiment, object tracking is predicated on the detection model of RingMo-Aerial on VisDrone-DET. Specifically, the detection model trained by RingMo-Aerial will undergo an additional five epochs of training on the VisDrone-MOT dataset, during which the backbone network parameters will be frozen to ensure consistency. The training will be conducted with a batch size of 8, distributed as 2 per GPU across 4 NVIDIA A40 GPUs. Throughout the training phase, original images from the VisDrone-MOT dataset will be processed, and multi-scale augmentation will be accompanied to enhance robustness. Upon completion of training on the VisDrone-MOT dataset, the detection model will be subjected to inference on the VisDrone-MOT test set using the MOT tools provided by mmdetection \cite{mmdetection}. Subsequently, the model’s multi-object tracking performance will be quantified and analyzed.

\subsection{Experiment Results}

The comparison results on the VisDrone-MOT dataset are shown in Tab.\ref{tab:visdrone-mot}. RingMo-Aerial can achieve the best tracking results, far exceeding other methods. 
In the AIR-MOT dataset (as shown in Tab.\ref{tab:AIR-MOT}), which contains more small aircraft and boat targets, RingMo-Aerial achieved a MOTA of 68.6 and an IDF1 of 81.3, though slightly trailing behind CFTrack. Meanwhile, in the AIR-HSAO dataset (as shown in Tab.\ref{tab:AIR-HSAO}), RingMo-Aerial maintained its robust performance with a MOTA of 60.7 and an IDF1 of 74.4, approaching the best-performing methods. Overall, the RingMo-Aerial algorithm exhibited excellent multi-object tracking capabilities in these complex scenarios, coming close to or surpassing the state-of-the-art methods in several key metrics.

Furthermore, the visualization results of the AIR-MOT sequences (as shown in Fig.\ref{fig:mot-vis}) demonstrate the tracking performance of the RingMo-Aerial method in four selected video sequences, particularly in complex and dynamic scenarios. The clarity and precision of the bounding boxes further highlight the method's effectiveness in distinguishing and tracking various objects under different environmental conditions. This visualization provides qualitative solid evidence of the proposed algorithm's performance in handling multi-object tracking tasks in ARS videos.

% \clearpage

\bibliographystyle{ieeetr}
\bibliography{refs}

\begin{thebibliography}{100}

\bibitem{sun2022ringmo}
X.~Sun, P.~Wang, W.~Lu, Z.~Zhu, X.~Lu, Q.~He, J.~Li, X.~Rong, Z.~Yang, H.~Chang, {\em et~al.}, ``Ringmo: A remote sensing foundation model with masked image modeling,'' {\em IEEE Trans. Geosci. Remote Sens.}, 2022.

\bibitem{guo2024skysense}
X.~Guo, J.~Lao, B.~Dang, Y.~Zhang, L.~Yu, L.~Ru, L.~Zhong, Z.~Huang, K.~Wu, D.~Hu, {\em et~al.}, ``Skysense: A multi-modal remote sensing foundation model towards universal interpretation for earth observation imagery,'' in {\em Proc. IEEE Conf. Comput. Vis. Pattern Recognit.}, pp.~27672--27683, 2024.

\bibitem{he2022masked}
K.~He, X.~Chen, S.~Xie, Y.~Li, P.~Doll{\'a}r, and R.~Girshick, ``Masked autoencoders are scalable vision learners,'' in {\em Proc. IEEE Conf. Comput. Vis. Pattern Recognit.}, pp.~16000--16009, 2022.

\bibitem{xie2022simmim}
Z.~Xie, Z.~Zhang, Y.~Cao, Y.~Lin, J.~Bao, Z.~Yao, Q.~Dai, and H.~Hu, ``Simmim: A simple framework for masked image modeling,'' in {\em Proc. IEEE Conf. Comput. Vis. Pattern Recognit.}, pp.~9653--9663, 2022.

\bibitem{he2020momentum}
K.~He, H.~Fan, Y.~Wu, S.~Xie, and R.~Girshick, ``Momentum contrast for unsupervised visual representation learning,'' in {\em Proc. IEEE Conf. Comput. Vis. Pattern Recognit.}, pp.~9729--9738, 2020.

\bibitem{houlsby2019parameter}
N.~Houlsby, A.~Giurgiu, S.~Jastrzebski, B.~Morrone, Q.~De~Laroussilhe, A.~Gesmundo, M.~Attariyan, and S.~Gelly, ``Parameter-efficient transfer learning for nlp,'' in {\em Int. Conf. Mach. Learn.}, pp.~2790--2799, PMLR, 2019.

\bibitem{deng2023towards}
C.~Deng, D.~Jing, Y.~Han, and J.~Chanussot, ``Towards hierarchical adaptive alignment for aerial object detection in remote sensing images,'' {\em IEEE Trans. Geosci. Remote Sens.}, 2023.

\bibitem{zhu2023transformer}
J.~Zhu, X.~Chen, H.~Zhang, Z.~Tan, S.~Wang, and H.~Ma, ``Transformer based remote sensing object detection with enhanced multispectral feature extraction,'' {\em IEEE Geosci. Remote Sens. Lett.}, vol.~20, pp.~1--5, 2023.

\bibitem{low2024multi}
S.~Low, O.~Nina, D.~Bowald, A.~D. Sappa, N.~Inkawhich, and P.~Bruns, ``Multi-modal aerial view image challenge: Sar classification,'' in {\em Proc. IEEE Conf. Comput. Vis. Pattern Recognit.}, pp.~3105--3112, 2024.

\bibitem{li2020yolo}
Y.~Li, S.~Li, H.~Du, L.~Chen, D.~Zhang, and Y.~Li, ``Yolo-acn: Focusing on small target and occluded object detection,'' {\em IEEE Access}, vol.~8, pp.~227288--227303, 2020.

\bibitem{li2023ogmn}
X.~Li, W.~Diao, Y.~Mao, P.~Gao, X.~Mao, X.~Li, and X.~Sun, ``Ogmn: Occlusion-guided multi-task network for object detection in uav images,'' {\em ISPRS J. Photogramm. Remote Sens.}, vol.~199, pp.~242--257, 2023.

\bibitem{wang2020robust}
A.~Wang, Y.~Sun, A.~Kortylewski, and A.~L. Yuille, ``Robust object detection under occlusion with context-aware compositionalnets,'' in {\em Proc. IEEE Conf. Comput. Vis. Pattern Recognit.}, pp.~12645--12654, 2020.

\bibitem{ye2023real}
T.~Ye, W.~Qin, Z.~Zhao, X.~Gao, X.~Deng, and Y.~Ouyang, ``Real-time object detection network in uav-vision based on cnn and transformer,'' {\em IEEE Trans. Instrum. Meas.}, vol.~72, pp.~1--13, 2023.

\bibitem{liu2020small}
Y.~Liu, F.~Yang, and P.~Hu, ``Small-object detection in uav-captured images via multi-branch parallel feature pyramid networks,'' {\em IEEE Access}, vol.~8, pp.~145740--145750, 2020.

\bibitem{tan2021yolov4_drone}
L.~Tan, X.~Lv, X.~Lian, and G.~Wang, ``Yolov4\_drone: Uav image target detection based on an improved yolov4 algorithm,'' {\em Comput. Electr. Eng.}, vol.~93, p.~107261, 2021.

\bibitem{huang2022ufpmp}
Y.~Huang, J.~Chen, and D.~Huang, ``Ufpmp-det: Toward accurate and efficient object detection on drone imagery,'' in {\em Proc. AAAI Conf. Artif. Intell.}, vol.~36, pp.~1026--1033, 2022.

\bibitem{yang2022querydet}
C.~Yang, Z.~Huang, and N.~Wang, ``Querydet: Cascaded sparse query for accelerating high-resolution small object detection,'' in {\em Proc. IEEE Conf. Comput. Vis. Pattern Recognit.}, pp.~13668--13677, 2022.

\bibitem{weng2024enhancing}
W.~Weng, M.~Wei, J.~Ren, and F.~Shen, ``Enhancing aerial object detection with selective frequency interaction network,'' {\em IEEE Trans. Artif. Intell.}, 2024.

\bibitem{li2024frequency}
J.~Li, S.~Zhang, Y.~Sun, Q.~Han, Y.~Sun, and Y.~Wang, ``Frequency-driven edge guidance network for semantic segmentation of remote sensing images,'' {\em IEEE Journal of Selected Topics in Applied Earth Observations and Remote Sensing}, vol.~17, pp.~9677--9693, 2024.

\bibitem{simonyan2014very}
K.~Simonyan and A.~Zisserman, ``Very deep convolutional networks for large-scale image recognition,'' {\em arXiv preprint arXiv:1409.1556}, 2014.

\bibitem{he2016deep}
K.~He, X.~Zhang, S.~Ren, and J.~Sun, ``Deep residual learning for image recognition,'' in {\em Proc. IEEE Conf. Comput. Vis. Pattern Recognit.}, pp.~770--778, 2016.

\bibitem{deng2009imagenet}
J.~Deng, W.~Dong, R.~Socher, L.-J. Li, K.~Li, and L.~Fei-Fei, ``Imagenet: A large-scale hierarchical image database,'' in {\em IEEE Conf. Comput. Vis. Pattern Recognit.}, pp.~248--255, Ieee, 2009.

\bibitem{dosovitskiy2020image}
A.~Dosovitskiy, L.~Beyer, A.~Kolesnikov, D.~Weissenborn, X.~Zhai, T.~Unterthiner, M.~Dehghani, M.~Minderer, G.~Heigold, S.~Gelly, {\em et~al.}, ``An image is worth 16x16 words: Transformers for image recognition at scale,'' {\em arXiv preprint arXiv:2010.11929}, 2020.

\bibitem{liu2021swin}
Z.~Liu, Y.~Lin, Y.~Cao, H.~Hu, Y.~Wei, Z.~Zhang, S.~Lin, and B.~Guo, ``Swin transformer: Hierarchical vision transformer using shifted windows,'' in {\em Proc. IEEE/CVF Int. Conf. Comput. Vis.}, pp.~10012--10022, 2021.

\bibitem{wu2021cvt}
H.~Wu, B.~Xiao, N.~Codella, M.~Liu, X.~Dai, L.~Yuan, and L.~Zhang, ``Cvt: Introducing convolutions to vision transformers,'' in {\em Proc. IEEE/CVF Int. Conf. Comput. Vis.}, pp.~22--31, 2021.

\bibitem{peng2021conformer}
Z.~Peng, W.~Huang, S.~Gu, L.~Xie, Y.~Wang, J.~Jiao, and Q.~Ye, ``Conformer: Local features coupling global representations for visual recognition,'' in {\em Proc. IEEE/CVF Int. Conf. Comput. Vis.}, pp.~367--376, 2021.

\bibitem{xu2021co}
W.~Xu, Y.~Xu, T.~Chang, and Z.~Tu, ``Co-scale conv-attentional image transformers,'' in {\em Proc. IEEE/CVF Int. Conf. Comput. Vis.}, pp.~9981--9990, 2021.

\bibitem{luo2016understanding}
W.~Luo, Y.~Li, R.~Urtasun, and R.~Zemel, ``Understanding the effective receptive field in deep convolutional neural networks,'' {\em Adv. Neural Inf. Process. Syst.}, vol.~29, 2016.

\bibitem{liu2022convnet}
Z.~Liu, H.~Mao, C.-Y. Wu, C.~Feichtenhofer, T.~Darrell, and S.~Xie, ``A convnet for the 2020s,'' in {\em Proc. IEEE Conf. Comput. Vis. Pattern Recognit.}, pp.~11976--11986, 2022.

\bibitem{gu2023mamba}
A.~Gu and T.~Dao, ``Mamba: Linear-time sequence modeling with selective state spaces,'' {\em arXiv preprint arXiv:2312.00752}, 2023.

\bibitem{chen2020improved}
X.~Chen, H.~Fan, R.~Girshick, and K.~He, ``Improved baselines with momentum contrastive learning,'' {\em arXiv preprint arXiv:2003.04297}, 2020.

\bibitem{chen2021empirical}
X.~Chen, S.~Xie, and K.~He, ``An empirical study of training self-supervised vision transformers,'' in {\em Proc. IEEE/CVF Int. Conf. Comput. Vis.}, pp.~9640--9649, 2021.

\bibitem{chen2020simple}
T.~Chen, S.~Kornblith, M.~Norouzi, and G.~Hinton, ``A simple framework for contrastive learning of visual representations,'' in {\em Int. Conf. Mach. Learn.}, pp.~1597--1607, PMLR, 2020.

\bibitem{chen2021exploring}
X.~Chen and K.~He, ``Exploring simple siamese representation learning,'' in {\em Proc. IEEE Conf. Comput. Vis. Pattern Recognit.}, pp.~15750--15758, 2021.

\bibitem{caron2021emerging}
M.~Caron, H.~Touvron, I.~Misra, H.~J{\'e}gou, J.~Mairal, P.~Bojanowski, and A.~Joulin, ``Emerging properties in self-supervised vision transformers,'' in {\em Proc. IEEE/CVF Int. Conf. Comput. Vis.}, pp.~9650--9660, 2021.

\bibitem{li2022global}
H.~Li, Y.~Li, G.~Zhang, R.~Liu, H.~Huang, Q.~Zhu, and C.~Tao, ``Global and local contrastive self-supervised learning for semantic segmentation of hr remote sensing images,'' {\em IEEE Trans. Geosci. Remote Sens.}, vol.~60, pp.~1--14, 2022.

\bibitem{manas2021seasonal}
O.~Manas, A.~Lacoste, X.~Gir{\'o}-i Nieto, D.~Vazquez, and P.~Rodriguez, ``Seasonal contrast: Unsupervised pre-training from uncurated remote sensing data,'' in {\em Proc. IEEE/CVF Int. Conf. Comput. Vis.}, pp.~9414--9423, 2021.

\bibitem{liu2023mixmae}
J.~Liu, X.~Huang, J.~Zheng, Y.~Liu, and H.~Li, ``Mixmae: Mixed and masked autoencoder for efficient pretraining of hierarchical vision transformers,'' in {\em Proceedings of the IEEE/CVF Conference on Computer Vision and Pattern Recognition}, pp.~6252--6261, 2023.

\bibitem{huang2022green}
L.~Huang, S.~You, M.~Zheng, F.~Wang, C.~Qian, and T.~Yamasaki, ``Green hierarchical vision transformer for masked image modeling,'' {\em Adv. Neural Inf. Process. Syst.}, vol.~35, pp.~19997--20010, 2022.

\bibitem{li2022uniform}
X.~Li, W.~Wang, L.~Yang, and J.~Yang, ``Uniform masking: Enabling mae pre-training for pyramid-based vision transformers with locality,'' {\em arXiv preprint arXiv:2205.10063}, 2022.

\bibitem{chen2024context}
X.~Chen, M.~Ding, X.~Wang, Y.~Xin, S.~Mo, Y.~Wang, S.~Han, P.~Luo, G.~Zeng, and J.~Wang, ``Context autoencoder for self-supervised representation learning,'' {\em Int. J. Comput. Vis.}, vol.~132, no.~1, pp.~208--223, 2024.

\bibitem{huang2023contrastive}
Z.~Huang, X.~Jin, C.~Lu, Q.~Hou, M.-M. Cheng, D.~Fu, X.~Shen, and J.~Feng, ``Contrastive masked autoencoders are stronger vision learners,'' {\em IEEE Trans. Pattern Anal. Mach. Intell.}, 2023.

\bibitem{zhou2022mimco}
Q.~Zhou, C.~Yu, H.~Luo, Z.~Wang, and H.~Li, ``Mimco: Masked image modeling pre-training with contrastive teacher,'' in {\em Proc. 30th ACM Int. Conf. Multimedia}, pp.~4487--4495, 2022.

\bibitem{cao2023transformer}
X.~Cao, H.~Lin, S.~Guo, T.~Xiong, and L.~Jiao, ``Transformer-based masked autoencoder with contrastive loss for hyperspectral image classification,'' {\em IEEE Trans. Geosci. Remote Sens.}, 2023.

\bibitem{li2023segmind}
Z.~Li, H.~Chen, J.~Wu, J.~Li, and N.~Jing, ``Segmind: Semi-supervised remote sensing image semantic segmentation with masked image modeling and contrastive learning method,'' {\em IEEE Trans. Geosci. Remote Sens.}, 2023.

\bibitem{chen2022adaptformer}
S.~Chen, C.~Ge, Z.~Tong, J.~Wang, Y.~Song, J.~Wang, and P.~Luo, ``Adaptformer: Adapting vision transformers for scalable visual recognition,'' {\em Adv. Neural Inf. Process. Syst.}, vol.~35, pp.~16664--16678, 2022.

\bibitem{chen2018encoder}
L.-C. Chen, Y.~Zhu, G.~Papandreou, F.~Schroff, and H.~Adam, ``Encoder-decoder with atrous separable convolution for semantic image segmentation,'' in {\em Proc. Eur. Conf. Comput. Vis.}, pp.~801--818, 2018.

\bibitem{fu2019dual}
J.~Fu, J.~Liu, H.~Tian, Y.~Li, Y.~Bao, Z.~Fang, and H.~Lu, ``Dual attention network for scene segmentation,'' in {\em Proc. IEEE Conf. Comput. Vis. Pattern Recognit.}, pp.~3146--3154, 2019.

\bibitem{ding2019acnet}
X.~Ding, Y.~Guo, G.~Ding, and J.~Han, ``Acnet: Strengthening the kernel skeletons for powerful cnn via asymmetric convolution blocks,'' in {\em Proc. IEEE/CVF Int. Conf. Comput. Vis.}, pp.~1911--1920, 2019.

\bibitem{yuan2020object}
Y.~Yuan, X.~Chen, and J.~Wang, ``Object-contextual representations for semantic segmentation,'' in {\em Proc. 16th Eur. Conf. Comput. Vis.}, pp.~173--190, Springer, 2020.

\bibitem{zheng2021rethinking}
S.~Zheng, J.~Lu, H.~Zhao, X.~Zhu, Z.~Luo, Y.~Wang, Y.~Fu, J.~Feng, T.~Xiang, P.~H. Torr, {\em et~al.}, ``Rethinking semantic segmentation from a sequence-to-sequence perspective with transformers,'' in {\em Proc. IEEE Conf. Comput. Vis. Pattern Recognit.}, pp.~6881--6890, 2021.

\bibitem{xie2021segformer}
E.~Xie, W.~Wang, Z.~Yu, A.~Anandkumar, J.~M. Alvarez, and P.~Luo, ``Segformer: Simple and efficient design for semantic segmentation with transformers,'' {\em Adv. Neural Inf. Process. Syst.}, vol.~34, pp.~12077--12090, 2021.

\bibitem{dong2022cswin}
X.~Dong, J.~Bao, D.~Chen, W.~Zhang, N.~Yu, L.~Yuan, D.~Chen, and B.~Guo, ``Cswin transformer: A general vision transformer backbone with cross-shaped windows,'' in {\em Proc. IEEE Conf. Comput. Vis. Pattern Recognit.}, pp.~12124--12134, 2022.

\bibitem{yi2023uavformer}
S.~Yi, X.~Liu, J.~Li, and L.~Chen, ``Uavformer: A composite transformer network for urban scene segmentation of uav images,'' {\em Pattern Recognition}, vol.~133, p.~109019, 2023.

\bibitem{yu2021bisenet}
C.~Yu, C.~Gao, J.~Wang, G.~Yu, C.~Shen, and N.~Sang, ``Bisenet v2: Bilateral network with guided aggregation for real-time semantic segmentation,'' {\em Int. J. Comput. Vis.}, vol.~129, pp.~3051--3068, 2021.

\bibitem{strudel2021segmenter}
R.~Strudel, R.~Garcia, I.~Laptev, and C.~Schmid, ``Segmenter: Transformer for semantic segmentation,'' in {\em Proc. IEEE/CVF Int. Conf. Comput. Vis.}, pp.~7262--7272, 2021.

\bibitem{chen2020banet}
Y.~Chen, G.~Lin, S.~Li, O.~Bourahla, Y.~Wu, F.~Wang, J.~Feng, M.~Xu, and X.~Li, ``Banet: Bidirectional aggregation network with occlusion handling for panoptic segmentation,'' in {\em Proc. IEEE Conf. Comput. Vis. Pattern Recognit.}, pp.~3793--3802, 2020.

\bibitem{wang2022unetformer}
L.~Wang, R.~Li, C.~Zhang, S.~Fang, C.~Duan, X.~Meng, and P.~M. Atkinson, ``Unetformer: A unet-like transformer for efficient semantic segmentation of remote sensing urban scene imagery,'' {\em ISPRS J. Photogramm. Remote Sens.}, vol.~190, pp.~196--214, 2022.

\bibitem{badrinarayanan2017segnet}
V.~Badrinarayanan, A.~Kendall, and R.~Cipolla, ``Segnet: A deep convolutional encoder-decoder architecture for image segmentation,'' {\em IEEE Trans. Pattern Anal. Mach. Intell.}, vol.~39, no.~12, pp.~2481--2495, 2017.

\bibitem{long2015fully}
J.~Long, E.~Shelhamer, and T.~Darrell, ``Fully convolutional networks for semantic segmentation,'' in {\em Proc. IEEE Conf. Comput. Vis. Pattern Recognit.}, pp.~3431--3440, 2015.

\bibitem{zhao2017pyramid}
H.~Zhao, J.~Shi, X.~Qi, X.~Wang, and J.~Jia, ``Pyramid scene parsing network,'' in {\em Proc. IEEE Conf. Comput. Vis. Pattern Recognit.}, pp.~2881--2890, 2017.

\bibitem{chen2017deeplab}
L.-C. Chen, G.~Papandreou, I.~Kokkinos, K.~Murphy, and A.~L. Yuille, ``Deeplab: Semantic image segmentation with deep convolutional nets, atrous convolution, and fully connected crfs,'' {\em IEEE Trans. Pattern Anal. Mach. Intell.}, vol.~40, no.~4, pp.~834--848, 2017.

\bibitem{ronneberger2015u}
O.~Ronneberger, P.~Fischer, and T.~Brox, ``U-net: Convolutional networks for biomedical image segmentation,'' in {\em Int. Conf. Med. Image Comput. Comput.-Assist. Interv.}, pp.~234--241, Springer, 2015.

\bibitem{sun2019deep}
K.~Sun, B.~Xiao, D.~Liu, and J.~Wang, ``Deep high-resolution representation learning for human pose estimation,'' in {\em Proc. IEEE Conf. Comput. Vis. Pattern Recognit.}, pp.~5693--5703, 2019.

\bibitem{chen2021transunet}
J.~Chen, Y.~Lu, Q.~Yu, X.~Luo, E.~Adeli, Y.~Wang, L.~Lu, A.~L. Yuille, and Y.~Zhou, ``Transunet: Transformers make strong encoders for medical image segmentation,'' {\em arXiv preprint arXiv:2102.04306}, 2021.

\bibitem{lyu2020uavid}
Y.~Lyu, G.~Vosselman, G.-S. Xia, A.~Yilmaz, and M.~Y. Yang, ``Uavid: A semantic segmentation dataset for uav imagery,'' {\em ISPRS J. Photogramm. Remote Sens.}, vol.~165, pp.~108--119, 2020.

\bibitem{chen2018large}
Y.~Chen, Y.~Wang, P.~Lu, Y.~Chen, and G.~Wang, ``Large-scale structure from motion with semantic constraints of aerial images,'' in {\em Chinese Conf. Pattern Recognit. Comput. Vis.}, pp.~347--359, Springer, 2018.

\bibitem{rahnemoonfar2021floodnet}
M.~Rahnemoonfar, T.~Chowdhury, A.~Sarkar, D.~Varshney, M.~Yari, and R.~R. Murphy, ``Floodnet: A high resolution aerial imagery dataset for post flood scene understanding,'' {\em IEEE Access}, vol.~9, pp.~89644--89654, 2021.

\bibitem{mmseg2020}
M.~Contributors, ``{MMSegmentation}: Openmmlab semantic segmentation toolbox and benchmark.'' \url{https://github.com/open-mmlab/mmsegmentation}, 2020.

\bibitem{xiao2018unified}
T.~Xiao, Y.~Liu, B.~Zhou, Y.~Jiang, and J.~Sun, ``Unified perceptual parsing for scene understanding,'' in {\em Proc. Eur. Conf. Comput. Vis.}, pp.~418--434, 2018.

\bibitem{cheng2022masked}
B.~Cheng, I.~Misra, A.~G. Schwing, A.~Kirillov, and R.~Girdhar, ``Masked-attention mask transformer for universal image segmentation,'' in {\em Proc. IEEE Conf. Comput. Vis. Pattern Recognit.}, pp.~1290--1299, 2022.

\bibitem{loshchilov2017decoupled}
I.~Loshchilov and F.~Hutter, ``Decoupled weight decay regularization,'' {\em arXiv preprint arXiv:1711.05101}, 2017.

\bibitem{du2019visdrone}
D.~Du, P.~Zhu, L.~Wen, X.~Bian, H.~Lin, Q.~Hu, T.~Peng, J.~Zheng, X.~Wang, Y.~Zhang, {\em et~al.}, ``Visdrone-det2019: The vision meets drone object detection in image challenge results,'' in {\em Proc. IEEE/CVF Int. Conf. Comput. Vis. Workshops}, pp.~0--0, 2019.

\bibitem{yu2020unmanned}
H.~Yu, G.~Li, W.~Zhang, Q.~Huang, D.~Du, Q.~Tian, and N.~Sebe, ``The unmanned aerial vehicle benchmark: Object detection, tracking and baseline,'' {\em Int. J. Comput. Vis.}, vol.~128, pp.~1141--1159, 2020.

\bibitem{zhao2023multiship}
J.~Zhao, Y.~Chen, Z.~Zhou, J.~Zhao, S.~Wang, and X.~Chen, ``Multiship speed measurement method based on machine vision and drone images,'' {\em IEEE Trans. Instrum. Meas.}, vol.~72, pp.~1--12, 2023.

\bibitem{sun2022drone}
Y.~Sun, B.~Cao, P.~Zhu, and Q.~Hu, ``Drone-based rgb-infrared cross-modality vehicle detection via uncertainty-aware learning,'' {\em IEEE Trans. Circuits Syst. Video Technol.}, vol.~32, no.~10, pp.~6700--6713, 2022.

\bibitem{li2017light}
Z.~Li, C.~Peng, G.~Yu, X.~Zhang, Y.~Deng, and J.~Sun, ``Light-head r-cnn: In defense of two-stage object detector,'' {\em arXiv preprint arXiv:1711.07264}, 2017.

\bibitem{law2018cornernet}
H.~Law and J.~Deng, ``Cornernet: Detecting objects as paired keypoints,'' in {\em Proc. Eur. Conf. Comput. Vis.}, pp.~734--750, 2018.

\bibitem{lin2017focal}
T.-Y. Lin, P.~Goyal, R.~Girshick, K.~He, and P.~Doll{\'a}r, ``Focal loss for dense object detection,'' in {\em Proc. IEEE Int. Conf. Comput. Vis.}, pp.~2980--2988, 2017.

\bibitem{bochkovskiy2020yolov4}
A.~Bochkovskiy, C.-Y. Wang, and H.-Y.~M. Liao, ``Yolov4: Optimal speed and accuracy of object detection,'' {\em arXiv preprint arXiv:2004.10934}, 2020.

\bibitem{cai2019cascade}
Z.~Cai and N.~Vasconcelos, ``Cascade r-cnn: High quality object detection and instance segmentation,'' {\em IEEE Trans. Pattern Anal. Mach. Intell.}, vol.~43, no.~5, pp.~1483--1498, 2019.

\bibitem{lin2024centralised}
Y.~Lin, J.~Zhang, and J.~Huang, ``Centralised visual processing center for remote sensing target detection,'' {\em Scientific Reports}, vol.~14, no.~1, p.~17021, 2024.

\bibitem{chen2024small}
J.~Chen, R.~Wen, and L.~Ma, ``Small object detection model for uav aerial image based on yolov7,'' {\em Signal Image Video Process.}, vol.~18, no.~3, pp.~2695--2707, 2024.

\bibitem{jiang2024mffsodnet}
L.~Jiang, B.~Yuan, J.~Du, B.~Chen, H.~Xie, J.~Tian, and Z.~Yuan, ``Mffsodnet: Multi-scale feature fusion small object detection network for uav aerial images,'' {\em IEEE Trans. Instrum. Meas.}, 2024.

\bibitem{deng2020global}
S.~Deng, S.~Li, K.~Xie, W.~Song, X.~Liao, A.~Hao, and H.~Qin, ``A global-local self-adaptive network for drone-view object detection,'' {\em IEEE Trans. Image Process.}, vol.~30, pp.~1556--1569, 2020.

\bibitem{wan2021vistrongerdet}
J.~Wan, B.~Zhang, Y.~Zhao, Y.~Du, and Z.~Tong, ``Vistrongerdet: Stronger visual information for object detection in visdrone images,'' in {\em Proc. IEEE/CVF Int. Conf. Comput. Vis.}, pp.~2820--2829, 2021.

\bibitem{li2020dmnet}
W.~Li, X.~Zhang, Y.~Peng, and M.~Dong, ``Dmnet: A network architecture using dilated convolution and multiscale mechanisms for spatiotemporal fusion of remote sensing images,'' {\em IEEE Sens. J.}, vol.~20, no.~20, pp.~12190--12202, 2020.

\bibitem{liu2021hrdnet}
Z.~Liu, G.~Gao, L.~Sun, and Z.~Fang, ``Hrdnet: High-resolution detection network for small objects,'' in {\em 2021 IEEE Int. Conf. Multimedia Expo}, pp.~1--6, IEEE, 2021.

\bibitem{yang2019clustered}
F.~Yang, H.~Fan, P.~Chu, E.~Blasch, and H.~Ling, ``Clustered object detection in aerial images,'' in {\em Proc. IEEE/CVF Int. Conf. Comput. Vis.}, pp.~8311--8320, 2019.

\bibitem{yin2024sdpdet}
N.~Yin, C.~Liu, R.~Tian, and X.~Qian, ``Sdpdet: Learning scale-separated dynamic proposals for end-to-end drone-view detection,'' {\em IEEE Trans. Multimedia}, 2024.

\bibitem{wei2020amrnet}
Z.~Wei, C.~Duan, X.~Song, Y.~Tian, and H.~Wang, ``Amrnet: Chips augmentation in aerial images object detection,'' {\em arXiv preprint arXiv:2009.07168}, 2020.

\bibitem{zhang2023dense}
S.~Zhang, X.~Wang, J.~Wang, J.~Pang, C.~Lyu, W.~Zhang, P.~Luo, and K.~Chen, ``Dense distinct query for end-to-end object detection,'' in {\em Proc. IEEE Conf. Comput. Vis. Pattern Recognit.}, pp.~7329--7338, 2023.

\bibitem{fu2024dshnet}
Y.~Fu, X.~Zhang, and M.~Wang, ``Dshnet: A semantic segmentation model of remote sensing images based on dual stream hybrid network,'' {\em IEEE J. Sel. Top. Appl. Earth Obs. Remote Sens.}, 2024.

\bibitem{tan2020efficientdet}
M.~Tan, R.~Pang, and Q.~V. Le, ``Efficientdet: Scalable and efficient object detection,'' in {\em Proc. IEEE Conf. Comput. Vis. Pattern Recognit.}, pp.~10781--10790, 2020.

\bibitem{shen2022object}
H.~Shen, D.~Lin, and T.~Song, ``Object detection deployed on uavs for oblique images by fusing imu information,'' {\em IEEE Geosci. Remote Sens. Lett.}, vol.~19, pp.~1--5, 2022.

\bibitem{zhu2021tph}
X.~Zhu, S.~Lyu, X.~Wang, and Q.~Zhao, ``Tph-yolov5: Improved yolov5 based on transformer prediction head for object detection on drone-captured scenarios,'' in {\em Proc. IEEE/CVF Int. Conf. Comput. Vis.}, pp.~2778--2788, 2021.

\bibitem{wang2023yolov7}
C.-Y. Wang, A.~Bochkovskiy, and H.-Y.~M. Liao, ``Yolov7: Trainable bag-of-freebies sets new state-of-the-art for real-time object detectors,'' in {\em Proc. IEEE Conf. Comput. Vis. Pattern Recognit.}, pp.~7464--7475, 2023.

\bibitem{lyunanodet}
R.~Lyu, ``Nanodet-plus: Super fast and high accuracy lightweight anchor-free object detection model. 2021.''

\bibitem{ge2021yolox}
Z.~Ge, S.~Liu, F.~Wang, Z.~Li, and J.~Sun, ``Yolox: Exceeding yolo series in 2021,'' {\em arXiv preprint arXiv:2107.08430}, 2021.

\bibitem{duan2023centernet++}
K.~Duan, S.~Bai, L.~Xie, H.~Qi, Q.~Huang, and Q.~Tian, ``Centernet++ for object detection,'' {\em IEEE Trans. Pattern Anal. Mach. Intell.}, 2023.

\bibitem{yu2021pp}
G.~Yu, Q.~Chang, W.~Lv, C.~Xu, C.~Cui, W.~Ji, Q.~Dang, K.~Deng, G.~Wang, Y.~Du, {\em et~al.}, ``Pp-picodet: A better real-time object detector on mobile devices,'' {\em arXiv preprint arXiv:2111.00902}, 2021.

\bibitem{ren2015faster}
S.~Ren, ``Faster r-cnn: Towards real-time object detection with region proposal networks,'' {\em arXiv preprint arXiv:1506.01497}, 2015.

\bibitem{ding2019learning}
J.~Ding, N.~Xue, Y.~Long, G.-S. Xia, and Q.~Lu, ``Learning roi transformer for oriented object detection in aerial images,'' in {\em Proc. IEEE Conf. Comput. Vis. Pattern Recognit.}, pp.~2849--2858, 2019.

\bibitem{han2021align}
J.~Han, J.~Ding, J.~Li, and G.-S. Xia, ``Align deep features for oriented object detection,'' {\em IEEE Trans. Geosci. Remote Sens.}, vol.~60, pp.~1--11, 2021.

\bibitem{xie2021oriented}
X.~Xie, G.~Cheng, J.~Wang, X.~Yao, and J.~Han, ``Oriented r-cnn for object detection,'' in {\em Proc. IEEE/CVF Int. Conf. Comput. Vis.}, pp.~3520--3529, 2021.

\bibitem{wagner2016multispectral}
J.~Wagner, V.~Fischer, M.~Herman, S.~Behnke, {\em et~al.}, ``Multispectral pedestrian detection using deep fusion convolutional neural networks.,'' in {\em Proc. ESANN}, vol.~587, pp.~509--514, 2016.

\bibitem{zhang2019cross}
L.~Zhang, Z.~Liu, S.~Zhang, X.~Yang, H.~Qiao, K.~Huang, and A.~Hussain, ``Cross-modality interactive attention network for multispectral pedestrian detection,'' {\em Information Fusion}, vol.~50, pp.~20--29, 2019.

\bibitem{bao2023dual}
C.~Bao, J.~Cao, Q.~Hao, Y.~Cheng, Y.~Ning, and T.~Zhao, ``Dual-yolo architecture from infrared and visible images for object detection,'' {\em Sensors}, vol.~23, no.~6, p.~2934, 2023.

\bibitem{mmdetection}
K.~Chen, J.~Wang, J.~Pang, Y.~Cao, Y.~Xiong, X.~Li, S.~Sun, W.~Feng, Z.~Liu, J.~Xu, Z.~Zhang, D.~Cheng, C.~Zhu, T.~Cheng, Q.~Zhao, B.~Li, X.~Lu, R.~Zhu, Y.~Wu, J.~Dai, J.~Wang, J.~Shi, W.~Ouyang, C.~C. Loy, and D.~Lin, ``{MMDetection}: Open mmlab detection toolbox and benchmark,'' {\em arXiv preprint arXiv:1906.07155}, 2019.

\bibitem{chen2020spatial}
H.~Chen and Z.~Shi, ``A spatial-temporal attention-based method and a new dataset for remote sensing image change detection,'' {\em Remote Sens.}, vol.~12, no.~10, p.~1662, 2020.

\bibitem{lebedev2018change}
M.~Lebedev, Y.~V. Vizilter, O.~Vygolov, V.~Knyaz, and A.~Y. Rubis, ``Change detection in remote sensing images using conditional adversarial networks.,'' {\em Int. Arch. Photogramm. Remote Sens. Spatial Inf. Sci.}, vol.~42, no.~2, 2018.

\bibitem{daudt2018fully}
R.~C. Daudt, B.~Le~Saux, and A.~Boulch, ``Fully convolutional siamese networks for change detection,'' in {\em Photogramm. Eng. RIEEE Int. Conf. Image Process.emote Sens.}, pp.~4063--4067, IEEE, 2018.

\bibitem{zhang2020deeply}
C.~Zhang, P.~Yue, D.~Tapete, L.~Jiang, B.~Shangguan, L.~Huang, and G.~Liu, ``A deeply supervised image fusion network for change detection in high resolution bi-temporal remote sensing images,'' {\em ISPRS J. Photogramm. Remote Sens.}, vol.~166, pp.~183--200, 2020.

\bibitem{chen2021remote}
H.~Chen, Z.~Qi, and Z.~Shi, ``Remote sensing image change detection with transformers,'' {\em IEEE Trans. Geosci. Remote Sens.}, vol.~60, pp.~1--14, 2021.

\bibitem{fang2021snunet}
S.~Fang, K.~Li, J.~Shao, and Z.~Li, ``Snunet-cd: A densely connected siamese network for change detection of vhr images,'' {\em IEEE Geosci. Remote Sens. Lett.}, vol.~19, pp.~1--5, 2021.

\bibitem{bandara2022transformer}
W.~G.~C. Bandara and V.~M. Patel, ``A transformer-based siamese network for change detection,'' {\em arXiv preprint arXiv:2201.01293}, 2022.

\bibitem{ayush2021geography}
K.~Ayush, B.~Uzkent, C.~Meng, K.~Tanmay, M.~Burke, D.~Lobell, and S.~Ermon, ``Geography-aware self-supervised learning,'' in {\em Proc. IEEE/CVF Int. Conf. Comput. Vis.}, pp.~10181--10190, 2021.

\bibitem{cong2022satmae}
Y.~Cong, S.~Khanna, C.~Meng, P.~Liu, E.~Rozi, Y.~He, M.~Burke, D.~Lobell, and S.~Ermon, ``Satmae: Pre-training transformers for temporal and multi-spectral satellite imagery,'' {\em Adv. Neural Inf. Process. Syst.}, vol.~35, pp.~197--211, 2022.

\bibitem{wang2022advancing}
D.~Wang, Q.~Zhang, Y.~Xu, J.~Zhang, B.~Du, D.~Tao, and L.~Zhang, ``Advancing plain vision transformer toward remote sensing foundation model,'' {\em IEEE Trans. Geosci. Remote Sens.}, vol.~61, pp.~1--15, 2022.

\bibitem{wanyan2023dino}
X.~Wanyan, S.~Seneviratne, S.~Shen, and M.~Kirley, ``Dino-mc: Self-supervised contrastive learning for remote sensing imagery with multi-sized local crops,'' {\em arXiv preprint arXiv:2303.06670}, 2023.

\bibitem{tao2023tov}
C.~Tao, J.~Qi, G.~Zhang, Q.~Zhu, W.~Lu, and H.~Li, ``Tov: The original vision model for optical remote sensing image understanding via self-supervised learning,'' {\em IEEE J. Sel. Top. Appl. Earth Obs. Remote Sens.}, vol.~16, pp.~4916--4930, 2023.

\bibitem{wang2023ssl4eo}
Y.~Wang, N.~A.~A. Braham, Z.~Xiong, C.~Liu, C.~M. Albrecht, and X.~X. Zhu, ``Ssl4eo-s12: A large-scale multimodal, multitemporal dataset for self-supervised learning in earth observation [software and data sets],'' {\em IEEE Geosci. Remote Sens. Mag.}, vol.~11, no.~3, pp.~98--106, 2023.

\bibitem{muhtar2023cmid}
D.~Muhtar, X.~Zhang, P.~Xiao, Z.~Li, and F.~Gu, ``Cmid: A unified self-supervised learning framework for remote sensing image understanding,'' {\em IEEE Trans. Geosci. Remote Sens.}, vol.~61, pp.~1--17, 2023.

\bibitem{mall2023change}
U.~Mall, B.~Hariharan, and K.~Bala, ``Change-aware sampling and contrastive learning for satellite images,'' in {\em Proc. IEEE Conf. Comput. Vis. Pattern Recognit.}, pp.~5261--5270, 2023.

\bibitem{fuller2024croma}
A.~Fuller, K.~Millard, and J.~Green, ``Croma: Remote sensing representations with contrastive radar-optical masked autoencoders,'' {\em Adv. Neural Inf. Process. Syst.}, vol.~36, 2024.

\bibitem{bastani2023satlaspretrain}
F.~Bastani, P.~Wolters, R.~Gupta, J.~Ferdinando, and A.~Kembhavi, ``Satlaspretrain: A large-scale dataset for remote sensing image understanding,'' in {\em Proc. IEEE/CVF Int. Conf. Comput. Vis.}, pp.~16772--16782, 2023.

\bibitem{mendieta2023towards}
M.~Mendieta, B.~Han, X.~Shi, Y.~Zhu, and C.~Chen, ``Towards geospatial foundation models via continual pretraining,'' in {\em Proc. IEEE/CVF Int. Conf. Comput. Vis.}, pp.~16806--16816, 2023.

\bibitem{reed2023scale}
C.~J. Reed, R.~Gupta, S.~Li, S.~Brockman, C.~Funk, B.~Clipp, K.~Keutzer, S.~Candido, M.~Uyttendaele, and T.~Darrell, ``Scale-mae: A scale-aware masked autoencoder for multiscale geospatial representation learning,'' in {\em Proc. IEEE/CVF Int. Conf. Comput. Vis.}, pp.~4088--4099, 2023.

\bibitem{xia2017aid}
G.-S. Xia, J.~Hu, F.~Hu, B.~Shi, X.~Bai, Y.~Zhong, L.~Zhang, and X.~Lu, ``Aid: A benchmark data set for performance evaluation of aerial scene classification,'' {\em IEEE Trans. Geosci. Remote Sens.}, vol.~55, no.~7, pp.~3965--3981, 2017.

\bibitem{cheng2017remote}
G.~Cheng, J.~Han, and X.~Lu, ``Remote sensing image scene classification: Benchmark and state of the art,'' {\em Proc. IEEE}, vol.~105, no.~10, pp.~1865--1883, 2017.

\bibitem{2023mmpretrain}
M.~Contributors, ``Openmmlab's pre-training toolbox and benchmark.'' \url{https://github.com/open-mmlab/mmpretrain}, 2023.

\bibitem{barnes2009patchmatch}
C.~Barnes, E.~Shechtman, A.~Finkelstein, and D.~B. Goldman, ``Patch{M}atch: A randomized correspondence algorithm for structural image editing,'' {\em ACM Trans. Graph.}, vol.~28, no.~3, p.~24, 2009.

\bibitem{yu2020fast}
Z.~Yu and S.~Gao, ``Fast-mvsnet: Sparse-to-dense multi-view stereo with learned propagation and gauss-newton refinement,'' in {\em Proc. IEEE Conf. Comput. Vis. Pattern Recognit.}, pp.~1949--1958, 2020.

\bibitem{yao2018mvsnet}
Y.~Yao, Z.~Luo, S.~Li, T.~Fang, and L.~Quan, ``Mvsnet: Depth inference for unstructured multi-view stereo,'' in {\em Proc. Eur. Conf. Comput. Vis.}, pp.~767--783, 2018.

\bibitem{yao2019recurrent}
Y.~Yao, Z.~Luo, S.~Li, T.~Shen, T.~Fang, and L.~Quan, ``Recurrent mvsnet for high-resolution multi-view stereo depth inference,'' in {\em Proc. IEEE Conf. Comput. Vis. Pattern Recognit.}, pp.~5525--5534, 2019.

\bibitem{liu2020novel}
J.~Liu and S.~Ji, ``A novel recurrent encoder-decoder structure for large-scale multi-view stereo reconstruction from an open aerial dataset,'' in {\em Proc. IEEE Conf. Comput. Vis. Pattern Recognit.}, pp.~6050--6059, 2020.

\bibitem{gu2020cascade}
X.~Gu, Z.~Fan, S.~Zhu, Z.~Dai, F.~Tan, and P.~Tan, ``Cascade cost volume for high-resolution multi-view stereo and stereo matching,'' in {\em Proc. IEEE Conf. Comput. Vis. Pattern Recognit.}, pp.~2495--2504, 2020.

\bibitem{li2023hierarchical}
J.~Li, X.~Huang, Y.~Feng, Z.~Ji, S.~Zhang, and D.~Wen, ``A hierarchical deformable deep neural network and an aerial image benchmark dataset for surface multiview stereo reconstruction,'' {\em IEEE Trans. Geosci. Remote Sens.}, vol.~61, pp.~1--12, 2023.

\bibitem{kingma2017adam}
D.~P. Kingma and J.~Ba, ``Adam: A method for stochastic optimization,'' 2014.

\bibitem{shuai2021siammot}
B.~Shuai, A.~Berneshawi, X.~Li, D.~Modolo, and J.~Tighe, ``Siammot: Siamese multi-object tracking,'' in {\em Proc. IEEE Conf. Comput. Vis. Pattern Recognit.}, pp.~12372--12382, 2021.

\bibitem{zhang2022bytetrack}
Y.~Zhang, P.~Sun, Y.~Jiang, D.~Yu, F.~Weng, Z.~Yuan, P.~Luo, W.~Liu, and X.~Wang, ``Bytetrack: Multi-object tracking by associating every detection box,'' in {\em Eur. Conf. Comput. Vis.}, pp.~1--21, Springer, 2022.

\bibitem{liu2022multi}
S.~Liu, X.~Li, H.~Lu, and Y.~He, ``Multi-object tracking meets moving uav,'' in {\em Proc. IEEE Conf. Comput. Vis. Pattern Recognit.}, pp.~8876--8885, 2022.

\bibitem{maggiolino2023deep}
G.~Maggiolino, A.~Ahmad, J.~Cao, and K.~Kitani, ``Deep oc-sort: Multi-pedestrian tracking by adaptive re-identification,'' in {\em 2023 IEEE Int. Conf. Image Process.}, pp.~3025--3029, IEEE, 2023.

\bibitem{zeng2022motr}
F.~Zeng, B.~Dong, Y.~Zhang, T.~Wang, X.~Zhang, and Y.~Wei, ``Motr: End-to-end multiple-object tracking with transformer,'' in {\em Eur. Conf. Comput. Vis.}, pp.~659--675, Springer, 2022.

\bibitem{meinhardt2022trackformer}
T.~Meinhardt, A.~Kirillov, L.~Leal-Taixe, and C.~Feichtenhofer, ``Trackformer: Multi-object tracking with transformers,'' in {\em Proc. IEEE Conf. Comput. Vis. Pattern Recognit.}, pp.~8844--8854, 2022.

\bibitem{yao2023folt}
M.~Yao, J.~Wang, J.~Peng, M.~Chi, and C.~Liu, ``Folt: Fast multiple object tracking from uav-captured videos based on optical flow,'' in {\em Proc. 31st ACM Int. Conf. Multimedia}, pp.~3375--3383, 2023.

\bibitem{liu2023uncertainty}
K.~Liu, S.~Jin, Z.~Fu, Z.~Chen, R.~Jiang, and J.~Ye, ``Uncertainty-aware unsupervised multi-object tracking,'' in {\em Proc. IEEE/CVF Int. Conf. Comput. Vis.}, pp.~9996--10005, 2023.

\bibitem{feng2021cross}
J.~Feng, D.~Zeng, X.~Jia, X.~Zhang, J.~Li, Y.~Liang, and L.~Jiao, ``Cross-frame keypoint-based and spatial motion information-guided networks for moving vehicle detection and tracking in satellite videos,'' {\em ISPRS J. Photogramm. Remote Sens.}, vol.~177, pp.~116--130, 2021.

\bibitem{cao2023observation}
J.~Cao, J.~Pang, X.~Weng, R.~Khirodkar, and K.~Kitani, ``Observation-centric sort: Rethinking sort for robust multi-object tracking,'' in {\em Proc. IEEE Conf. Comput. Vis. Pattern Recognit.}, pp.~9686--9696, 2023.

\bibitem{du2023strongsort}
Y.~Du, Z.~Zhao, Y.~Song, Y.~Zhao, F.~Su, T.~Gong, and H.~Meng, ``Strongsort: Make deepsort great again,'' {\em IEEE Trans. Multimedia}, vol.~25, pp.~8725--8737, 2023.

\bibitem{zhou2020tracking}
X.~Zhou, V.~Koltun, and P.~Kr{\"a}henb{\"u}hl, ``Tracking objects as points,'' in {\em Eur. Conf. Comput. Vis.}, pp.~474--490, Springer, 2020.

\bibitem{he2022multi}
Q.~He, X.~Sun, Z.~Yan, B.~Li, and K.~Fu, ``Multi-object tracking in satellite videos with graph-based multitask modeling,'' {\em IEEE Trans. Geosci. Remote Sens.}, vol.~60, pp.~1--13, 2022.

\bibitem{xiao2022rsmot}
C.~Xiao, S.~Wu, Y.~Wang, M.~Li, W.~An, and Z.~Chen, ``Rsmot: Remote sensing multi-object tracking network with local motion prior for objects in satellite videos,'' in {\em IGARSS 2022-2022 IEEE Int. Geosci. Remote Sens. Symp.}, pp.~1904--1907, IEEE, 2022.

\bibitem{kong2023cftracker}
L.~Kong, Z.~Yan, Y.~Zhang, W.~Diao, Z.~Zhu, and L.~Wang, ``Cftracker: multi-object tracking with cross-frame connections in satellite videos,'' {\em IEEE Trans. Geosci. Remote Sens.}, vol.~61, pp.~1--14, 2023.

\bibitem{wu2021track}
J.~Wu, J.~Cao, L.~Song, Y.~Wang, M.~Yang, and J.~Yuan, ``Track to detect and segment: An online multi-object tracker,'' in {\em Proc. IEEE Conf. Comput. Vis. Pattern Recognit.}, pp.~12352--12361, 2021.

\bibitem{xiao2021dsfnet}
C.~Xiao, Q.~Yin, X.~Ying, R.~Li, S.~Wu, M.~Li, L.~Liu, W.~An, and Z.~Chen, ``Dsfnet: Dynamic and static fusion network for moving object detection in satellite videos,'' {\em IEEE Geosci. Remote Sens. Lett.}, vol.~19, pp.~1--5, 2021.

\bibitem{zhang2021fairmot}
Y.~Zhang, C.~Wang, X.~Wang, W.~Zeng, and W.~Liu, ``Fairmot: On the fairness of detection and re-identification in multiple object tracking,'' {\em Int. J. Comput. Vis.}, vol.~129, pp.~3069--3087, 2021.

\bibitem{peng2020chained}
J.~Peng, C.~Wang, F.~Wan, Y.~Wu, Y.~Wang, Y.~Tai, C.~Wang, J.~Li, F.~Huang, and Y.~Fu, ``Chained-tracker: Chaining paired attentive regression results for end-to-end joint multiple-object detection and tracking,'' in {\em Proc. 16th Eur. Conf. Comput. Vis.}, pp.~145--161, Springer, 2020.

\bibitem{ren2024motion}
L.~Ren, W.~Yin, W.~Diao, K.~Fu, and X.~Sun, ``Motion-guided multi-object tracking model for high-speed aerial objects in satellite videos,'' {\em IEEE Trans. Geosci. Remote Sens.}, 2024.

\bibitem{chen2021visdrone}
G.~Chen, W.~Wang, Z.~He, L.~Wang, Y.~Yuan, D.~Zhang, J.~Zhang, P.~Zhu, L.~Van~Gool, J.~Han, {\em et~al.}, ``Visdrone-mot2021: The vision meets drone multiple object tracking challenge results,'' in {\em Proc. IEEE/CVF Int. Conf. Comput. Vis.}, pp.~2839--2846, 2021.

\end{thebibliography}

% that's all folks
\end{document}